\documentclass[doublespaced]{ut-thesis}

\usepackage{times}
\usepackage{latexsym}
\usepackage{graphicx}
\usepackage{amsfonts}
\usepackage{amsmath}
\usepackage{amssymb}
\usepackage{natbib}
\usepackage{url}
\usepackage{algorithm2e}
\usepackage{color}
\usepackage{multirow}
\usepackage{subcaption}
\usepackage{pifont}
\usepackage{array}
\usepackage{caption}
\usepackage{subcaption}
\usepackage{linguex}

\degree{Doctor of Philosophy}
\department{Computer Science}
\gradyear{2022}
\author{Bai Li}
\title{Integrating Linguistic Theory and Neural Language Models}

\newcommand{\vect}[1]{\boldsymbol{#1}}
\newcommand{\bfemph}[1]{\textbf{\textit{#1}}}

\begin{document}

\begin{preliminary}

\maketitle


\begin{abstract}
Transformer-based language models have recently achieved remarkable results in many natural language tasks. However, performance on leaderboards is generally achieved by leveraging massive amounts of training data and computation, and rarely by encoding explicit linguistic knowledge into neural models. This has led many to question the relevance of linguistics for modern natural language processing. In this dissertation, I present several case studies to illustrate how theoretical linguistics and deep neural language models are still relevant to each other. First, language models are useful to linguists by providing an automatic and objective tool to measure semantic distance, which is difficult to do using traditional methods. On the other hand, linguistic theory contributes to language modelling research by providing frameworks and sources of data to probe our language models for specific aspects of language understanding.

This thesis contributes three studies that explore different aspects of the syntax-semantics interface in language models. In the first part of my thesis, I apply language models to the problem of word class flexibility, a long-debated issue in theoretical linguistics. Using mBERT as a source of semantic distance measurements across many languages, I present evidence in favour of analyzing word class flexibility as a directional process. In the second part of my thesis, I propose a method to measure surprisal at intermediate layers of language models, using Gaussian models for density estimation. My experiments show that sentences containing morphosyntactic anomalies trigger surprisals earlier in language models than semantic and commonsense anomalies. Finally, in the third part of my thesis, I adapt several psycholinguistic studies to show that language models contain knowledge of argument structure constructions (a proposed analysis of verb valency from construction grammar theory). In summary, my thesis develops new connections between natural language processing, linguistic theory, and psycholinguistics to provide fresh perspectives for the interpretation of language models.
\end{abstract}





\begin{acknowledgements}

I am grateful for the funding provided by the University of Toronto and the Vector Institute, allowing me to pursue graduate research in natural language processing.

Thanks to my supervisor, Prof. Frank Rudzicz for being my guide and mentor throughout my journey in graduate school, from when I first joined your lab in 2017 as a master's student until the completion of my PhD. Thank you for teaching me how to do research, then eventually encouraging me to set my own direction, yet always being available when I needed your support.

Thanks to Prof. Robert Frank for serving as the external examiner for my final thesis defence, Prof. Suzanne Stevenson for serving on the defence committee, and Prof. Steven Waslander for serving as the meeting chair.

I would like to thank my supervisory committee, Prof. Yang Xu and Prof. Guillaume Thomas, for being outstanding collaborators. You have taught me an immense amount about linguistics and cognitive science, and this interdisciplinary research would not have been possible without your expertise. Thanks to your unrelenting critical feedback on my experimental design and paper writing, so that I may submit the paper confidently and without fearing Reviewer \#2.

Thanks to the graduate students in SPOClab and the Computational Linguistics group, including Arnold, Demetres, Francois, Hillary, Ian, Jixuan, John, KP, Raeid, Serena, Stephane, Yoona, Yuchen, and Zining. I learned a lot from collaborating with you on various coursework and research projects, and engaging in presentations and paper reading group discussions.

Thanks to my coworkers at Snaptravel (now Snapcommerce) and Ada Support, where I worked part-time as a machine learning engineer for a large part of my PhD. There, I have had the opportunity to experience many machine learning projects in the real world, beyond my narrow research domain.

I am thankful to have wonderful friends including Andrei, Kevin D, Kevin P, Leon, Michael, and Niranjan. It was fun to engage in heated debates with you about philosophy, world politics, economics, business ideas, and all sorts of other topics when I'm procrastinating on research.

Thanks to my printer, dutifully producing approximately 15,000 pages of papers and textbooks while hardly ever jamming.

Thanks to my parents in Calgary and sister in Toronto for encouraging me when times were uncertain and feeding me when I am at home. After three years of asking me {\em ``when will you be finishing your PhD?''}, I finally have a straight answer.

Finally, I am thankful for my wife Elaine, who has remained by my side to share my every success and failure, and offer me support when I needed it the most. I am lucky to be with you.

\end{acknowledgements}

\tableofcontents

\listoftables

\listoffigures


\end{preliminary}


\chapter{Introduction}

Recent advances in Transformer-based neural networks such as GPT \citep{gpt}, BERT \citep{bert}, and RoBERTa \citep{roberta} have achieved state-of-the-art performances in many natural language processing tasks (NLP), such as machine translation, text classification, parsing, and question answering. A major challenge with these language models (LMs) is understanding how they function, since they are composed of millions of neurons whose meaning is not easily interpretable by humans. At the same time, LMs are built from complex mechanisms that manipulate large matrices of real numbers, with no basis in linguistic theories that humans have developed to understand language. This naturally raises the question: {\em how can neural language models be integrated with linguistic theory}? In this thesis, I demonstrate how linguistic theories can serve as a foundation upon which LMs can be evaluated for language understanding; on the other hand, LMs can also provide evidence to support linguistic theories.

\section{Motivation: why linguistic probing?}

At first glance, it is not immediately obvious why linguistic theory is necessary or desirable for understanding language models. Progress in NLP is usually measured by a set of standard benchmarks, such as SQuAD 2 \citep{squad2} for question answering or SuperGLUE \citep{superglue} for various types of language understanding. Using these benchmarks, researchers can compare their models against previous ones, and public leaderboards rank all currently available models by their relative performance (often with a human baseline for comparison). Practitioners can use these benchmarks to decide which model to apply for their own use case (possibly along with other considerations such as model size and efficiency, which in any case can also be measured in benchmarks). One may wonder: if benchmarks already serve the needs of researchers and practitioners, why do we need to involve linguistic theory?

Benchmarks suffer from several drawbacks in practice. They become ``saturated'', where models quickly surpass human baseline on the benchmark, even though they do not outperform humans in general: this leads to a loss of trust in the benchmark's validity and hinders further progress in the field \citep{bowman-dahl}. Both SQuAD 2 and SuperGLUE had models that exceeded human performance within months of their release, even though they were designed to be difficult tasks, and neither question answering nor text classification are generally considered solved. This can happen when models learn to exploit biases in the benchmark dataset that enable it to predict the correct answer in a way that is unintended (and incorrect): for example, predicting that two sentences are contradictions if a negation word is present \citep{gururangan-artifact}. Eliminating all such biases is difficult because they are annotated by crowdworkers on naturally occurring data, and both tend to contain biases. Moreover, in most language understanding tasks, we lack a precise understanding of how the correct answer is related to the inputs of a task instance, so annotations must rely on the possibly differing interpretations of human annotators.

Targeted linguistic evaluation offers a remedy to these limitations of benchmarking. It draws on decades of research aimed at describing language in as precise detail as possible: which sentences are grammatical and ungrammatical, and how the meaning of a sentence is related to its surface structure. This body of knowledge gives the researcher the ability to control for biases and eliminate shortcut heuristics, revealing deficiencies in our models' understanding of many language phenomena, such as negation \citep{ettinger-psycholinguistic}, phrasal composition \citep{yu-ettinger}, and long-range dependencies \citep{van-schijndel-etal-2019-quantity}.

Probing tasks can be derived from linguistic theory via templates: generating sentences of a given structure according to a theoretical description of some language feature. An example of this is BLiMP \citep{blimp}, a benchmark of 67 sets of template-generated sentences testing linguistic phenomena such as agreement and movement. A second and more direct approach is taking data from psycholinguistic publications to use as probing -- these sentences are written by linguists for human experimental stimuli and are carefully controlled for possible biases. I use both techniques extensively to obtain probing data in Chapters 5 and 6 of this thesis.

Although linguistic probing offers certain advantages over standard NLP benchmarks, they are not meant to be a replacement for benchmarking -- the two serve different needs of the research community. The crucial difference is that probing aims to deepen our understanding of existing widely used language models, whereas developing new models to achieve a high performance on the probing task is of lesser importance. Therefore, linguistic probing can be viewed as a form of post-hoc interpretability, offering a global view of the model's capabilities by identifying areas of weakness from the perspective of linguistic theory \citep{posthoc-survey}.

\section{Bridging NLP and linguistics}

\begin{figure}[t]
    \centering
    \includegraphics[width=0.8\linewidth]{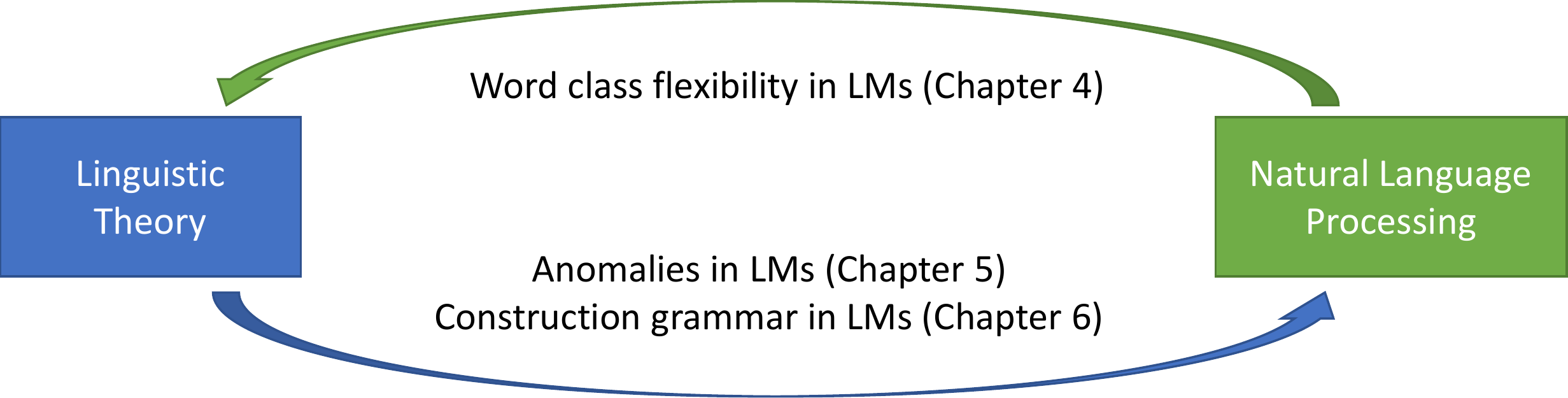}
    \caption{The main contributions of this thesis. Chapter 4 on word class flexibility uses LMs as evidence for a debate in linguistic theory; Chapter 5 on linguistic anomalies and Chapter 6 on construction grammar apply linguistic frameworks and data toward LM probing.}
    \label{fig:summary-flowchart}
\end{figure}

There has been relatively little contact between natural language processing and theoretical linguistics since the deep learning revolution. While the two fields share some similarities -- both involve data in human languages -- their primary goals are different. NLP aims to develop computational systems to solve language tasks as accurately as possible, whereas linguistics aims to describe properties of languages and how humans process them. Given these divergent goals, it is not obvious how advances in either field should be relevant to the other. Some researchers have attempted to incorporate linguistic and structural knowledge into deep neural models, but these methods have not shown substantial performance improvements over models without linguistic knowledge \citep{lappin2021}. The likely reason for this failure is that language models are able to learn implicit structural properties of language through the usual training procedure \citep{hewitt-syntax, miaschi2020linguistic}, so providing explicit knowledge is redundant.

Instead of improving model performance directly, the more promising avenue for linguistics to contribute to NLP has been linguistic probing. Previous work has probed LMs for knowledge of many linguistic phenomena, which I will survey in Chapter 3. In this dissertation, I expand this body of work by studying two linguistic phenomena which have not been covered in earlier work: how LMs represent different types of linguistic anomalies (Chapter 5), and how they understand argument structure constructions (Chapter 6).

In the opposite direction, it has been even more difficult for deep learning to contribute to linguistic theory. Neural network probing experiments do not provide information about how humans process language, and as a result, this body of work has been rarely cited in linguistic publications \citep{baroni-proper-role}. My research offers an avenue of contribution in this direction: using deep models as a tool to measure semantic distance between occurrences of a word in different contexts (Chapter 4). Semantic distance is a metric of importance for theories of word class flexibility, but is difficult to measure using traditional methods. My three projects in this thesis serve to bridge the gap between NLP and linguistics and provide examples of interdisciplinary collaboration for both research communities.

\section{The syntax-semantics interface in language models}

A traditional dichotomy in the study of language is between syntax and semantics. Syntax studies well-formed phrases and their internal structures, whereas semantics studies how meaning is derived from syntactic structures. Many linguistic phenomena involve the interplay between syntax and semantics, often with complex effects that pose challenging problems for linguistic theories. My thesis focuses on three phenomena on the interface between syntax and semantics, and explores these phenomena with novel experimental methods involving language models.

In Chapter 4, I study the phenomenon of word class flexibility, where certain ``flexible'' words may be used as different parts of speech: for example, in English, the words {\em work} or {\em sleep} may be used either as a noun or a verb. Whether all languages possess a distinction between nouns and verbs is a controversial question in linguistic typology, due to competing definitions of word classes. Word classes are defined by a combination of morphosyntactic distribution (e.g., only nouns may follow a determiner in English), and semantic criteria (e.g., nouns are associated with objects and verbs with actions; \citet{vanLierRijkhoff2013}). However, linguistic theories disagree over the analysis of flexible words: are they lexemes with underspecified word class, or cases of conversion, whereby a lexeme undergoes a derivational process into a different word class? My work presents evidence supporting the conversion theory, using measures of semantic variability.

Next, in Chapter 5, I explore sentences containing syntactic, semantic, and commonsense violations. Various linguistic theories have proposed differences between syntactic and semantic violations. In generative syntax, \citet{chomsky1957} proposed that ungrammaticality (e.g., {\em ``furiously sleep ideas green colorless''}) should be distinguished from semantic anomalies (e.g., {\em ``colorless green ideas sleep furiously''}): the latter being a meaningless but grammatically well-formed sentence. In psycholinguistics, studies found that semantic violations triggered the N400 event-related potential (ERP) in the brain whereas morphosyntactic violations triggered a different P600 ERP, although this dichotomy was abandoned after further evidence \citep{psycholinguistics-electrified}. Inspired by this psycholinguistic work, I probe language models for whether they exhibit any differences in internal processing in response to varying types of anomalies.

Finally, in Chapter 6, I adopt a construction grammar framework to probe language models. Construction grammar proposes that all linguistic knowledge is encoded in constructions, which map between form and meaning, and there is no separation between lexical and grammatical knowledge. Argument structure constructions (ASCs) theorize that certain syntactic patterns are associated with semantic meaning independently of the main verb \citep{goldberg1995}. For example, the ditransitive construction (e.g., {\em ``Bob cut Joe the bread''}) is associated with the transfer of the indirect object to the direct object recipient, no matter which verb is used. In contrast, lexicalist theories assume that the main verb is responsible for assigning semantic roles to each of its syntactic arguments. In psycholinguistics, sentence sorting and priming studies supported the theory of ASCs in humans; I adapt several of these studies to show evidence for ASCs in language models as well.

\section{Structure of thesis}

The rest of my thesis is structured as follows.

\begin{itemize}
    \item Chapter 2 gives a survey of modern neural language models, including static word embeddings based on the distributional hypothesis, RNN and LSTM sequence models, and Transformer-based LMs such as BERT and GPT. I also discuss the benefits and tradeoffs of some common diagnostic classifier schemes used for probing LMs.
    \item Chapter 3 surveys the recent literature connecting linguistic theories to LMs, including behavioural probes of syntactic structure and probes targeting the internal representations of LMs. Next, I discuss neural network probing methods adapted from psycholinguistics, and applications of LMs to linguistic theory.
    \item Chapter 4 presents my cross-lingual study on word class flexibility: my method leverages contextual embeddings from LMs as a source of automated semantic distance judgments, and I find evidence supporting the theoretical view of word class flexibility as a directional process.
    \item Chapter 5 presents my work on probing for linguistic anomalies within intermediate layers of LMs. Inspired by neurolinguistic event-related potential (ERP) studies, I propose an anomaly detection method based on Gaussian models and find that different internal activation patterns are triggered in response to different types of linguistic anomalies.
    \item Chapter 6 presents my work on probing LMs for construction grammar: specifically, a family of constructions known as argument structure constructions. I adapt several human psycholinguistic studies to be suitable for LMs, and show that LMs exhibit knowledge of argument structure constructions similarly to humans.
    \item Chapter 7 concludes my thesis by summarizing my contributions and highlighting some areas for future improvement.
\end{itemize}

\section{Relationship to published work}

Several chapters of this thesis have previously appeared in peer-reviewed publications:

\begin{itemize}
    \item {\bf Chapter 4.} Bai Li, Guillaume Thomas, Yang Xu, and Frank Rudzicz. ``Word class flexibility: A deep contextualized approach''. {\em Proceedings of the 2020 Conference on Empirical Methods in Natural Language Processing (EMNLP 2020)}.
    
    \item {\bf Chapter 5.} Bai Li, Zining Zhu, Guillaume Thomas, Yang Xu, and Frank Rudzicz. ``How is BERT surprised? Layerwise detection of linguistic anomalies''. {\em Proceedings of the Joint Conference of the 59th Annual Meeting of the Association for Computational Linguistics and the 11th International Joint Conference on Natural Language Processing (ACL-IJCNLP 2021)}.
    
    \item {\bf Chapter 6.} Bai Li, Zining Zhu, Guillaume Thomas, Frank Rudzicz, and Yang Xu. ``Neural reality of argument structure constructions''. {\em Proceedings of the 60th Annual Meeting of the Association for Computational Linguistics (ACL 2022)}.
\end{itemize}

Additionally, the following peer-reviewed publications are not included in this thesis but were published during my doctorate:

\begin{itemize}
    \item Bai Li, Nanyi Jiang, Joey Sham, Henry Shi, and Hussein Fazal. ``Real-world Conversational AI for Hotel Bookings''. {\em IEEE Annual Conference on Artificial Intelligence for Industries (AI4I 2019)}.

    \item Bai Li, Jing Yi Xie, and Frank Rudzicz. ``Representation Learning for Discovering Phonemic Tone Contours''. {\em 17th SIGMORPHON Workshop on Computational Research in Phonetics, Phonology, and Morphology (SIGMORPHON at ACL 2020)}.

    \item Bai Li and Frank Rudzicz. ``TorontoCL at CMCL 2021 Shared Task: RoBERTa with Multi-Stage Fine-Tuning for Eye-Tracking Prediction''. {\em Proceedings of the Workshop on Cognitive Modeling and Computational Linguistics (CMCL at NAACL 2021)}.
    
    \item Zining Zhu, Jixuan Wang, Bai Li, and Frank Rudzicz. ``On the data requirements of probing''. {\em Findings of the Association for Computational Linguistics: ACL 2022}.
\end{itemize}
\chapter{Modern neural language models}

\section{Introduction}

Modern natural language processing has achieved much of its success by leveraging a relatively small number of foundational models that serve as building blocks for many diverse applications. These models are trained on vast amounts of unstructured language data, allowing them to learn general-purpose representations of human language and transfer this knowledge towards downstream tasks. The architectures of these models have evolved over the years with research progress: in Section \ref{sec:word-vec}, I begin with static word vector models based on distributional semantics, which are typically used in the input layer in sequential RNN and LSTM neural networks (Section \ref{sec:rnn-lstm}). Next, in Section \ref{sec:transformers}, I cover contextualized language models such as BERT, which use a two-step procedure in which the model is pre-trained on large amounts of text, then fine-tuned to perform specific tasks. Finally, in Section \ref{sec:probing-classifiers}, I describe some methods to probe the internal representations of large language models and the difficulties and shortcomings of these probing methodologies.

\section{Distributional semantics and word embeddings} \label{sec:word-vec}

\begin{figure}
    \centering
    \includegraphics[width=0.4\linewidth]{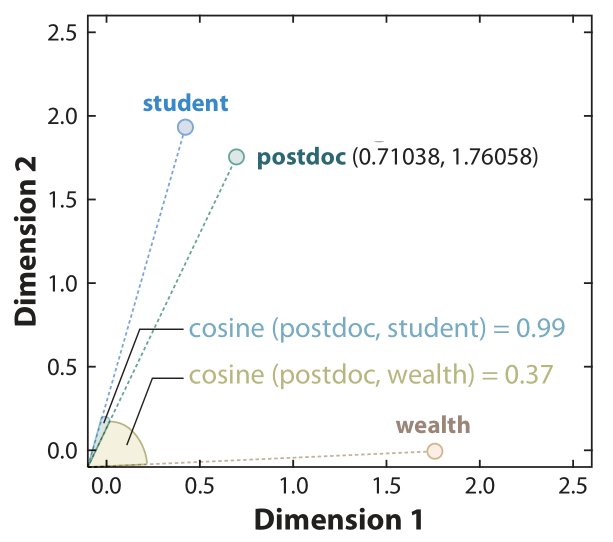}
    \caption{Visualizing distributional semantic models. Each individual dimension is semantically meaningless, but similar words are closer together in the vector space, as measured by Euclidean or cosine distance. Figure adapted from \citet{distributional-semantics-survey}.}
    \label{fig:distributional-semantics}
\end{figure}

Word embeddings (or vectors) are one of the most basic building blocks of natural language processing, encoding the meaning of a word into a fixed-dimensional vector of real numbers. These vectors are learned using large corpus data, and individual dimensions of these vectors do not contain any meaning; the meaning of words is captured by their geometric relationship to other word vectors. Words that are related in meaning are close together in vector space (measured by Euclidean or cosine distance), while words that are unrelated are farther apart (Figure \ref{fig:distributional-semantics}). Sometimes, word vectors exhibit vector arithmetic properties: for instance, the vector difference between {\em Canada} and {\em Ottawa} is close to the vector difference of {\em China} and {\em Beijing} because both word pairs exhibit the country-capital relationship.

The theoretical basis for word embeddings is from distributional semantics. \citet{harris-distributional} proposed the {\em distributional hypothesis}: words which occur in similar contexts have similar meaning, and conversely, a word's meaning can be defined by its contextual distribution. For example, given the sentence {\em ``The \_\_\_ stayed up all night to finish her paper''}, two likely completions for the blank are {\em student} and {\em postdoc}, so these two words have similar meaning according to distributional semantics.

There are many word embedding models based on the distributional hypothesis; one of the most popular and first neural approaches was word2vec \citep{word2vec}. Word2vec used a shallow neural network to either predict a word given its context (the continuous bag-of-words model), or predict the context of a word (the skip-gram model). In the continuous bag-of-words variant, the neural network is given a one-hot encoding of words in a context window surrounding it, and uses a classification layer to predict the middle word. After training, the word embeddings are derived from the hidden layer of this neural network.

A popular alternative method to learn word vectors was GLoVE \citep{glove}. Rather than predicting word identities from context, GLoVE learns vectors that reconstruct a word-to-word co-occurrence matrix, where each entry in the matrix is the count of how many times that pair of words co-occurred in the same context window in a corpus. Practically, word2vec and GLoVE produce word vectors that behave similarly, and neither one consistently outperforms the other.

Both word2vec and GLoVE have a weakness in that they treat each word as an atomic unit, so the internal structure of words is not utilized, and they cannot handle out-of-vocabulary words. Fasttext \citep{fasttext} proposed to enrich the word2vec skip-gram method with sub-word information: in addition to learning a vector for each word, Fasttext also learns vectors for character n-grams so that an out-of-vocabulary word can be represented by the sum of its character n-gram vectors.

Word embeddings have some weaknesses that limit their utilities in some situations. First, since they always generate a static vector for each word, they cannot capture homographs which have the same form but multiple senses. For example, the noun and verb senses of the word {\em bear} are completely unrelated, but both are assigned the same word embedding. The result is a less-than-ideal representation for both senses, since the optimization procedure produces an embedding that lies somewhere in between the two senses in an attempt to capture both simultaneously. In the next section, I discuss contextual embeddings which generate better representations of polysemous words. Another weakness is that words that have similar vectors do not always have similar meanings: one can only conclude that they occur in similar distributional contexts. For example, the sentence {\em ``The test was very \_\_\_''}, both {\em easy} and {\em difficult} are reasonable completions, despite having opposite meanings. Thus, it is generally difficult to distinguish synonyms from antonyms and hypernyms using vector space distance \citep{lenci-survey}.

\section{Sequential models and contextual embeddings} \label{sec:rnn-lstm}

Word embeddings provide representations for individual words, but they are unable to represent word order in longer texts such as sentences or paragraphs. In order to work with text, sequential models such as recurrent neural networks (RNNs; \citet{rnn}) are needed. An RNN is a type of neural network that reads sequential input one token at a time; the input may consist of one-hot encodings of words or lower-dimensional static word embeddings. The RNN keeps track of a hidden state representing the input read thus far and updates it after reading each token. The final state may be used as a representation for the entire sequence.

\begin{figure}
    \centering
    \includegraphics[width=0.55\linewidth]{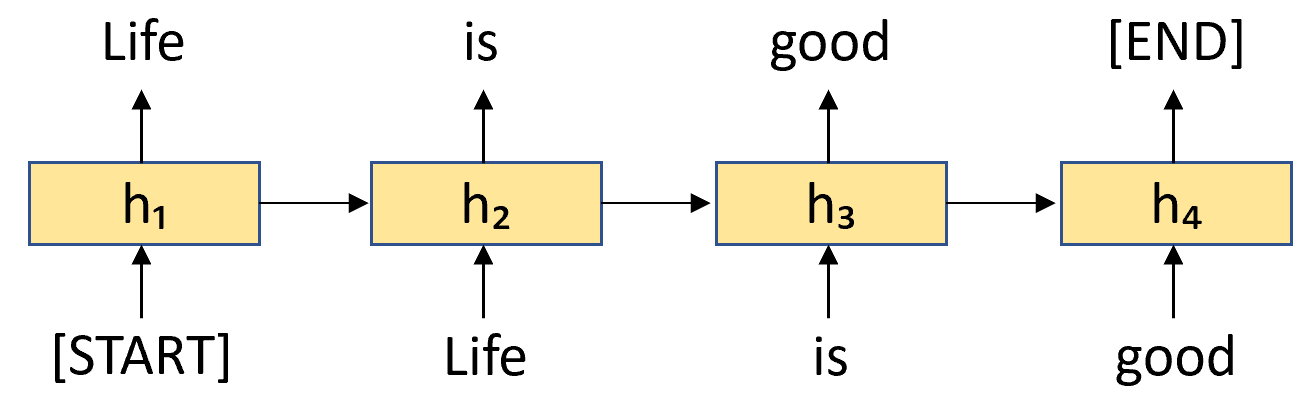}
    \caption{A recurrent neural network (RNN) for language modelling. At each step, the RNN predicts the next word in the sequence, given a hidden state that is derived from the previous words. The inputs may be one-hot encodings of the words, or static word embeddings such as word2vec or GLoVE.}
    \label{fig:rnn-lm}
\end{figure}

A popular variant of the RNN is long short-term memory networks (LSTMs; \citet{lstm}). Plain RNNs have difficulty processing long-range dependencies in text since information in the hidden layer must be retained across many steps. In contrast, LSTMs use a memory cell in addition to the hidden state and a system of input, output, and forget gates to modify the memory cell, allowing the model to more easily retain long-term information. Sometimes, it is useful to read input in both directions: a bidirectional LSTM (bi-LSTM) consists of a forward and backward LSTM whose outputs are concatenated together.

RNNs and LSTMs are used in several common setups for supervised prediction. For sequence classification, such as predicting the topic of a sentence, the last hidden representation may be fed into a discrete classification layer. When the task requires per-token classification, such as part-of-speech tagging, the hidden representation at each token can be fed into a classification layer. When the output is a sequence, such as machine translation, the last hidden representation may be fed into a decoder RNN or LSTM, which produces a sequential output.

A common unsupervised use case for sequential models is language modelling. In language modelling, the model is given a sequence of words and predicts which word is likely to come next; after training on a corpus of unlabelled text, the model can then be used to generate similar sequences. Language models also assign a probability to any given sequence: the {\em perplexity} is defined as the negative log likelihood of this probability:
$$p(S) = -\sum_{i=0}^{|S|} \log p(w_i | w_1, \cdots, w_{i-1}),$$
where each $w_i$ is a word in the sentence $S$. A related concept, {\em surprisal}, is the negative log likelihood of a single token given previous context: $- \log p(w_i | w_1, \cdots, w_{i-1})$. Perplexity is used as an unsupervised evaluation metric measuring how well a language model fits a corpus; it can also be used to rank sentences for relative acceptability according to the language model. A fair comparison using perplexity requires that the models must have the same vocabulary and the sentences must be the same length.

When trained for language modelling, neural models learn hidden representations that capture general properties of language that are useful for many tasks; we call these {\em language models} (LMs). This approach benefits from leveraging large amounts of unlabelled data, which is available in much larger quantities compared to labelled data required for supervised learning. Once trained, transfer learning can be applied to fine-tune the LM to specific tasks. Universal Language Model Fine-tuning (ULMFiT; \citet{ulmfit}) was the earliest model to use the transfer learning paradigm that is now dominant in natural language processing. ULMFiT first trained an LSTM model on a large corpus (the pre-training step), followed by a small amount of additional training on task-specific data (the fine-tuning step); the model used a slanted triangular learning rate schedule and gradual unfreezing of layers to prevent forgetting pre-trained information during the fine-tuning stage.

Another transfer method using unsupervised learning is Embeddings from Language Models (ELMo). ELMo trained a multilayer bidirectional LSTM on a corpus, then uses the hidden representation of the forward and backward LSTMs concatenated together as contextual embeddings. That is, rather than mapping each word to a static word vector, ELMo generates a sequence of contextual vectors for a sequence of tokens; these contextual vectors can then be substituted for any model that assumes a sequence of word vectors as input. By incorporating information from surrounding words, contextual vectors are able to generate different representations for different senses of a word like {\em bear}, overcoming the word sense ambiguity problem in static word vectors.

\section{Transformer-based language models} \label{sec:transformers}

\begin{figure}
    \centering
    \includegraphics[width=1.0\linewidth]{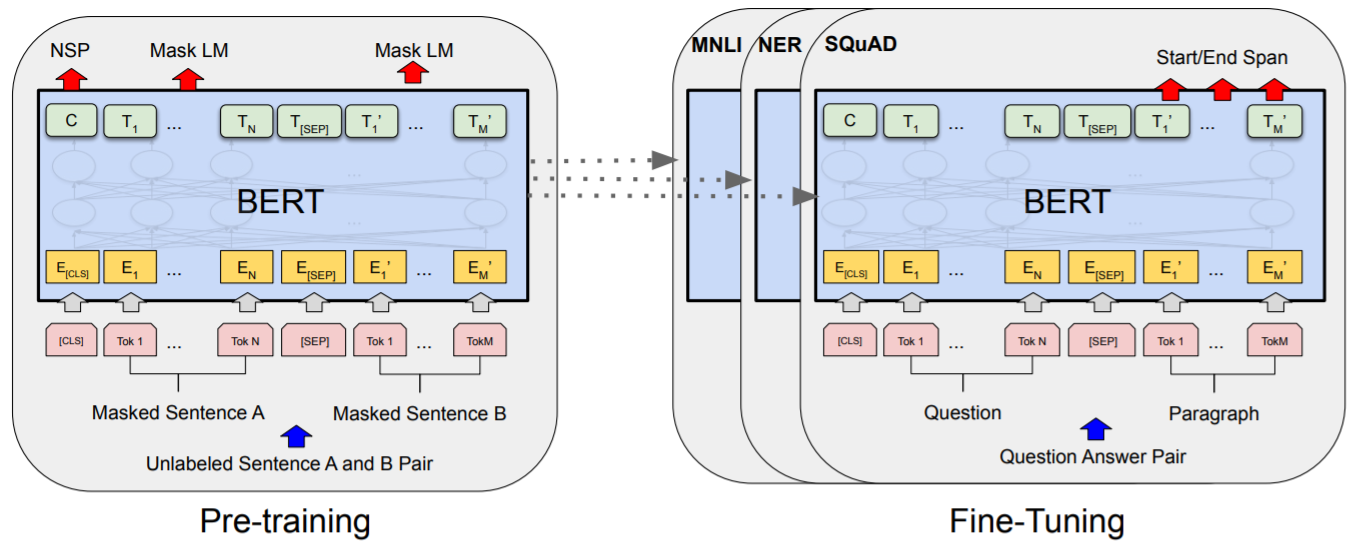}
    \caption{The BERT model \citep{bert}, consisting of 12 layers of Transformer modules \citep{transformer}. The model is first pre-trained on masked language modelling and next sentence prediction, then fine-tuned for downstream tasks. BERT may take either a single sentence as input, or two sentences separated by a \texttt{[SEP]} token.}
    \label{fig:bert}
\end{figure}

The next major architectural innovation after LSTMs was the Transformer module \citep{transformer}. Contrary to sequential models which process tokens one at a time, Transformers use the {\em self-attention} mechanism to process the entire sequence simultaneously; each token is associated with a positional encoding to retain word order information. The advantage of the Transformer architecture over RNNs and LSTMs is that it avoids problems with long-range dependencies where information needs to be carried forward across many steps: with self-attention, the maximum path length is constant (equal to the number of layers in the model), and does not depend on the distance between the input words. The Transformer module was originally employed for machine translation, using a stack of 6 encoder and 6 decoder Transformer layers; it subsequently replaced recurrent layers in many neural models.

Transformers were soon applied to unsupervised language modelling. OpenAI proposed the Generative Pre-trained Transformer (GPT; \citet{gpt}): similar to ULMFiT, GPT was pre-trained on the language modelling objective and then fine-tuned to perform specific tasks, but used Transformers instead of LSTMs. Bidirectional Encoder Representations from Transformers (BERT; \citet{bert}) incorporated bidirectional context into language modelling: whereas GPT predicted the next word from previous context, BERT proposed the {\em masked language modelling} (MLM) objective, where tokens were randomly replaced with \texttt{[MASK]} tokens which the model was trained to predict. BERT's pre-training procedure consisted of MLM as well as next-sentence prediction (NSP), where the model predicts whether two sentences are consecutive in the original text or two randomly chosen sentences. After pre-training, BERT is then fine-tuned to perform classification, sequence tagging, or sentence pair classification by adding a linear layer on top of the last layer and training for a small number of steps on task-specific data (Figure \ref{fig:bert}). When released, BERT immediately broke new records on many natural language benchmarks, leading to many efforts to improve upon it, study its internals, and apply it to downstream tasks.

BERT can be used to generate contextual embeddings and for language modelling. At each layer, BERT generates hidden representations at each token that can be used as contextual representations: typically, the last or second-to-last layer is used. While BERT is not naturally suitable for language modelling because it assumes bidirectional context is available, a substitute for perplexity was proposed by \citet{salazar-mlm-scoring}, where each token is replaced one at a time with the \texttt{[MASK]} token, BERT computes the log-likelihood score for each masked token, and the sum of log-likelihood scores (called the {\em pseudo-perplexity}) may be used for evaluation or to compare sentences for relative acceptability. However, pseudo-perplexity cannot be directly compared to normal perplexity scores from forward language models like LSTMs and GPT. Both forward and bidirectional models have their merits: bidirectional models generate better embeddings as they have access to context in both directions, but forward models are useful for conditionally generating text given an initial prompt.

After the release of BERT, many models have improved its performance by modifying its architecture, data, and training procedure. RoBERTa \citep{roberta} used the same architecture as BERT, but obtained superior results by removing the NSP task and training on a larger dataset. XLNet \citep{xlnet} proposed {\em permutation language modelling}, where a random permutation of the sentence is generated and the model learns to predict the next word in the permutation, given previous words and their positional encodings; this avoids the pretrain-finetune discrepancy in BERT where the \texttt{[MASK]} token is seen only in pre-training and not during fine-tuning. ALBERT \citep{albert} proposed a factorized embedding representation so that models with larger hidden layers can be trained using the same amount of memory, and replaced the NSP task with sentence-order prediction, where the model predicts the order of two sentences that were originally consecutive. ELECTRA \citep{electra} proposed the {\em replaced token detection} pre-training task to improve efficiency over MLM: instead of predicting masked tokens, the model is given corrupted text and predicts which tokens were original and which ones were replacements, using a smaller network to generate replacement tokens.

Transformer language models have been trained for other languages as well, often in a massively multilingual setting so that a single model is able to process text in many languages. Multilingual BERT (mBERT) was released by the authors of BERT, using the same architecture and trained on Wikipedia text. XLM \citep{xlm} added a {\em translation language modelling} task to mBERT, where the model predicts a masked token from context and a translation of the sentence into another language: this allows parallel corpora to be leveraged for pre-training. XLM-RoBERTa (XLM-R; \citep{xlmr}) trained XLM on a larger dataset and obtained results competitive with or surpassing the best monolingual models in each language.

\section{Probing classifiers and their shortcomings} \label{sec:probing-classifiers}

The success of BERT on natural language tasks quickly led many researchers to investigate what information is contained within BERT, and how the information is spread across its 12 layers. This task is nontrivial, as Transformer models internally consist of millions of neurons that are not easily interpretable. A popular approach is by using {\em probing classifiers} (or {\em probes}): a classifier that takes in BERT embeddings as input and is trained on some target task; the performance on this task is taken as a measure of how much information about the task is contained in the embeddings. The purpose of this probe is not to perform well on the task (since fine-tuning BERT would usually result in better performance), but to measure the extent that the embeddings contain information relevant to the task, without any fine-tuning.

\begin{figure}
    \centering
    \includegraphics[width=0.9\linewidth]{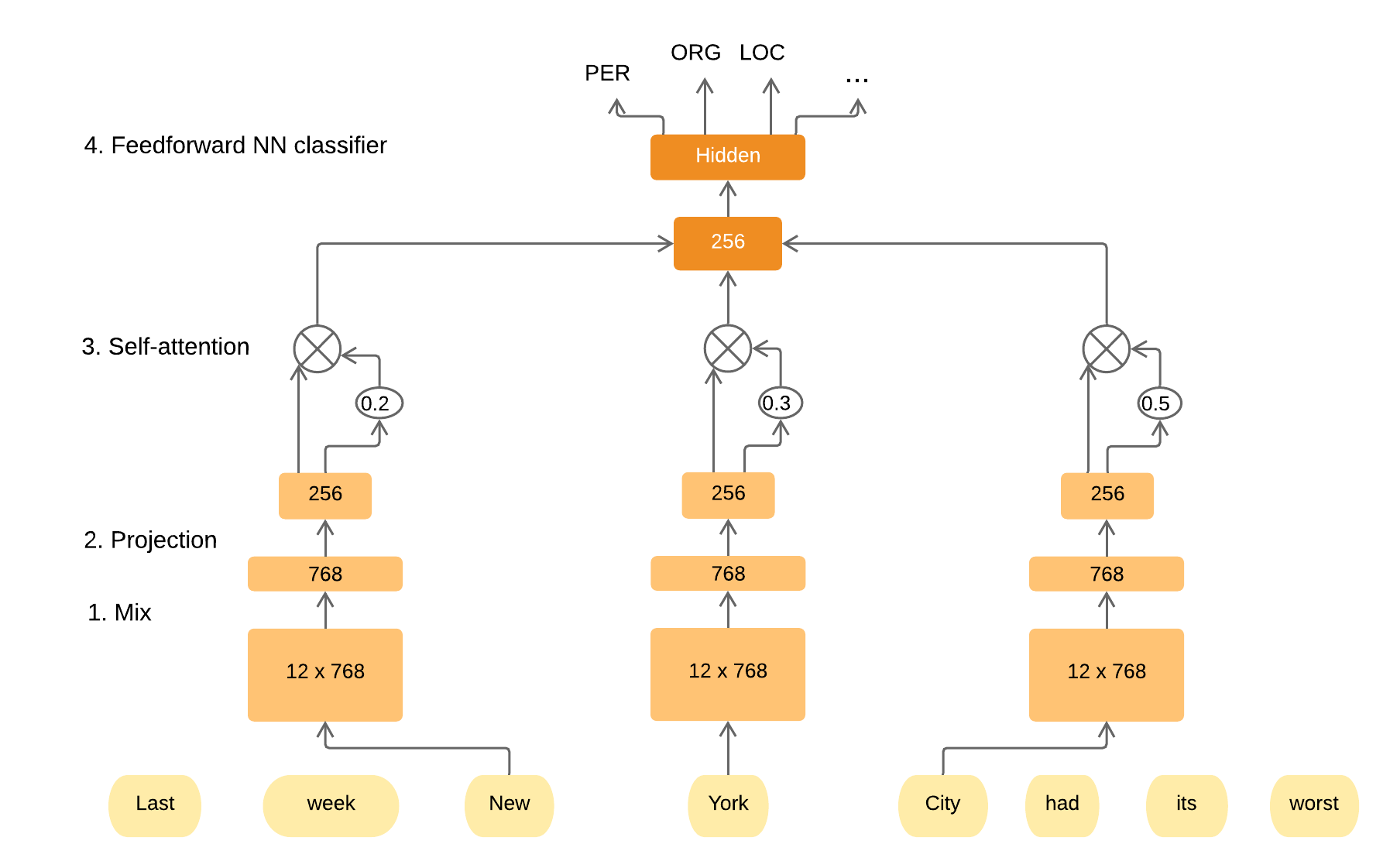}
    \caption{Illustration of the edge probing method proposed by \citet{edge-probing}. Here, the probe is given a sentence {\em ``Last week New York City had its worst ...''} and a span {\em ``New York City''} and the probing task is to classify the type of named entity represented by the span ({\em Location}).}
    \label{fig:edge-probe}
\end{figure}

\citet{edge-probing} introduced the {\em edge probing} method to determine how much more information is contained in contextual embeddings compared to static baselines. The edge probing setup assumes that the input consists of a sentence and up to two spans of consecutive tokens within the sentence, and the output consists of a single label. This formulation is applicable to part-of-speech tagging, dependency arc labelling, and coreference resolution, among others.

The probing model (Figure \ref{fig:edge-probe}) first uses BERT to generate contextual vectors for each token. Then, the {\em mix} step learns a task-specific linear combination of the layers; the projection and self-attention pooling produce a fixed-length span representation, and finally, a feedforward layer outputs the classification label. The probe weights are initialized randomly and trained using gradient descent while the BERT weights are kept frozen. Using this method, \citet{edge-probing} found the biggest advantage of contextual over static embeddings occurred in syntactic tasks. In a follow-up work, \citet{bert-rediscovers} inspected the layerwise weights learned in the {\em mix} step and found that semantic tasks learned higher weights on the upper layers of BERT compared to the syntactic tasks, suggesting that the upper layers contained more semantic information. I will discuss the linguistic implications of this experiment and other layerwise probing work in Section \ref{sec:representational-probes}.

One criticism of probing classifiers (especially complex ones such as edge probing) is that high probe performance could either mean that the representation is rich in information, or that the probe itself is powerful and learning the task. \citet{probes-control-tasks} proposed to use {\em control tasks}: randomized versions of the probing task constructed so that high performance is only possible if the probe itself learns the task. They defined {\em selectivity} as the difference between the real and control task performance; a good probe should have high selectivity; In their experiments, the simple linear probes had the highest selectivity. Alternatively, \citet{voita-mdl} proposed a {\em minimum description length} (MDL) probe based on information theory that simultaneously quantifies the probing accuracy and probe model accuracy. Their MDL probe obtained similar results as \citet{probes-control-tasks} but avoided some instability difficulties of the standard probe training procedure. Suffice it to say that there is still no agreement on the best method to probe the quality of LM representations.

Many papers have been published in the last few years that probe various aspects of how BERT and other Transformer models work. This subfield, commonly known as {\em BERTology}, explores questions such as what knowledge is contained in which neurons, information present at each layer, the role of architectural choices and pre-training regimes, etc. A detailed survey of BERTology is beyond the scope of this thesis; \citet{bertology} is a recent and comprehensive survey of the field. In the next chapter, I will provide a narrower survey of BERTology work that involves linguistic theory, such as constructing test suites of syntactic phenomena and probing based on psycholinguistics.
\chapter{Probing for linguistic knowledge in LMs}

\section{Introduction}

In this chapter, I survey the recent literature on probing language models for various aspects of linguistic knowledge. Many papers have been recently published that try to align neural networks with linguistic theory in some way; although each paper's specific methodology is different, probing methods can generally be categorized into two types: {\em behavioural probes} and {\em representational probes}. I will discuss behavioural probes in Section \ref{sec:behavioral-probes} and representational probes in Section \ref{sec:representational-probes}. To maintain a reasonable scope, this chapter will mostly cover probing work involving transformer models (i.e., BERT and later models).

Behavioural probes apply a black-box approach to probing: they assume little about the internals of the model, and carefully construct input sentences so that the way that the model responds to the input reveals information about its internal capabilities and biases. This approach is relatively robust and likely to remain relevant in spite of future advances in language modelling research because it makes no assumptions about model architecture. In contrast, representational probes assume full access to the model's internals (for example, its trained parameters, its contextual vectors, and attention weights when fed an input sentence). Because neural representations consist of millions of real numbers, sometimes complex machinery is required to make these results interpretable for humans, and one challenge is choosing which tools are most appropriate. Unlike in human psycholinguistics, we can easily access every state in a neural model and at any point during its processing, making representational probing a powerful and flexible methodology.

Next, in Section \ref{sec:psycholinguistic-probing}, I will survey LM probing research based on psycholinguistics. The field of psycholinguistics has developed many methods for indirectly exposing language processing facilities in the human brain: common experimental paradigms include lexical decision tests, priming, and EEG studies. This vast literature provides a rich starting point for investigating similar linguistic phenomena in language models. Finally, in Section \ref{sec:lm-linguistic-theory}, I discuss some recent efforts using language models to provide evidence for linguistic theories: while neural networks have not yet made a significant impact in theoretical linguistics, we are beginning to see initial progress in this direction as well.

\section{Behavioural probes for syntax} \label{sec:behavioral-probes}

\subsection{Agreement}

Much of the initial work in behavioural probing investigated agreement between the subject and the verb. In English, the subject and verb of a sentence must agree on number and person, for example:

\ex. \a.The {\bf boy} {\bf likes} music. \label{ex:boy-likes}
     \b.*The {\bf car} {\bf are} yellow. \label{ex:common-unacceptable}
     \c.Those {\bf narwhals} {\bf come} from Tuktoyaktuk. \label{ex:rare-acceptable}
     
There are several ways this can be operationalized in a behavioural probe. The first possibility is treating it as a binary classification task of classifying whether a sentence is acceptable or unacceptable. This can be done in a supervised setting (i.e., training on one set of acceptable and unacceptable sentences and evaluating on a different set), or an unsupervised setting (i.e., picking a threshold such that sentences with language model probability above the threshold are considered acceptable). However, the binary classification setup has the drawback that it does not control for the length and contents of the sentence, so that an acceptable sentence \ref{ex:rare-acceptable} containing rare words may have a lower probability than an unacceptable sentence \ref{ex:common-unacceptable} containing common words \citep{lau2017}.

A different approach that resolves this issue by framing the task as a forced choice between two minimal pairs: sentences that only differ on the critical region of interest, and are identical in all other aspects, for example:

\ex. \a.The car {\bf (is/*are)} yellow.
     \b.Those narwhals {\bf (*comes/come)} from Tuktoyaktuk.

This presents a natural setup for language models supporting masked word prediction such as BERT and RoBERTa: one can feed into the model the sentence {\em The boy [MASK] music}, obtaining a probability distribution for the masked token, and consider the model correct if the model's probability for {\em likes} is higher than for {\em like}. Otherwise, for forward sequential models like LSTMs and GPT, the input stimulus is usually modified so that only the prefix before the masked token is required to predict the correct completion. For example, \ref{ex:boy-likes} would be truncated to {\em The boy ...}; the model reads this input and generates a prediction for the subsequent token, and is judged as correct if the probability for {\em likes} is higher than for {\em like}. Construction of input stimuli for forward sequential models is therefore more restrictive than for masked language models, since the critical token must be the final token in the sentence.

\citet{linzen2016} tested LSTM models on number agreement between subject and verb, using natural sentences from Wikipedia. In addition to simple sentences, they examined more complex sentences containing {\em attractors} -- nouns between the subject and verb with opposite number from the subject. Such sentences are relatively uncommon but may occur if the subject is modified by a relative clause or prepositional phrase, for example, {\em The car behind those trees (is/*are) yellow.} These are difficult cases because the simple heuristic of agreeing with the most recent noun does not work; humans are also known to sometimes produce the incorrect inflection in the presence of attractors \citep{bock-miller}. \citet{linzen2016} found that the supervised models had higher error rates on sentences with attractors (but still better than random), while the unsupervised models were unable to predict agreement better than random when attractors are present.

\citet{gulordava-colorless} investigated the capabilities of LSTMs on long-distance agreement in nonsensical sentences, where all lexical items are replaced with random nonce words that match in part-of-speech and morphological features. The purpose of this experiment was to remove the possibility that the models rely on lexical and semantic cues to predict agreement. The authors generated nonce sentences by perturbing UD treebank sentences in 4 languages: English, Italian, Hebrew, and Russian; they found that in all languages, the LSTM performed worse on the nonce than the original sentences, but better than random. They concluded that the model could not have achieved its results solely through pattern-matching on surface structure, and instead it must have learned some form of deeper linguistic structure.

\subsection{Other syntactic phenomena}

Agreement is a relatively simple linguistic phenomenon, well suited for an initial case study of behavioural probing, and after its initial success, researchers soon expanded behavioural probing to other phenomena. Standard syntax textbooks (e.g., \citet{carnie-syntax}) describe a wide assortment of phenomena that may be adapted into benchmark tests for language models.

One of the most direct approaches was the Corpus of Linguistic Acceptability (CoLA; \citet{cola}). CoLA collected 10,657 sentences from various linguistic publications including textbooks and dissertations. Each sentence was labelled in the original publications as either acceptable or unacceptable. Models were allowed a train-test split, where the training set and test set were drawn from different linguistic publications; evaluation was by Matthews correlation coefficient with the ground truth. CoLa has since been incorporated into the GLUE text classification benchmark \citep{glue}.

Although CoLA contains a wide variety of phenomena, which is desirable for a general-purpose benchmark, its heterogeneity of source material is a hindrance for analysis of specific phenomena. The authors annotated the corpus with the presence or absence of 15 broad classes of phenomena and 63 fine-grained phenomena, but this still resulted in a loosely-related set of sentences within each phenomenon; they instead found template-based generation to be better suited for phenomenon-specific analysis.

\citet{marvin-linzen} introduced the technique of generating syntactic stimuli using templates: they used a recursive context-free grammar to generate random sentences of different linguistic structures. The advantage of this method is that it allows precise control over the structure of each sentence and how many of each type to generate (since certain structures rarely appear in natural corpora). Furthermore, template generation avoids lexical or any other potential confounds, similar to \citet{gulordava-colorless}. The authors generated agreement samples containing various types of complements and relative clauses, and additionally examine LSTM performance on reflexive anaphora and negative polarity items (NPIs).

\begin{table*}[t]\centering
\begin{tabular}{lp{0.36\linewidth}p{0.36\linewidth}}
\hline
Phenomenon & Acceptable Example & Unacceptable Example\\
\hline
Anaphor agreement & \emph{Many girls insulted \bfemph{themselves}.} & \emph{Many girls insulted \bfemph{herself}.}\\
Argument structure & \emph{Rose wasn't \bfemph{disturbing} Mark.} & \emph{Rose wasn't \bfemph{boasting} Mark.}\\
Binding & \emph{Carlos said that Lori helped \bfemph{him}.} & \emph{Carlos said that Lori helped  \bfemph{himself}.} \\
Control/raising & \emph{There was \bfemph{bound} to be a fish escaping.} & \emph{There was \bfemph{unable} to be a fish escaping.}\\
Determiner-noun agr. & \emph{Rachelle had bought that \bfemph{chair}.} & \emph{Rachelle had bought that \bfemph{chairs}.}\\
Ellipsis & \emph{Anne's doctor cleans one \bfemph{important} book and Stacey cleans a few.} & \emph{Anne's doctor cleans one book and Stacey cleans a few \bfemph{important}.}\\
Filler-gap & \emph{Brett knew \bfemph{what} many waiters find.} & \emph{Brett knew \bfemph{that} many waiters find.}\\
Irregular forms & \emph{Aaron \bfemph{broke} the unicycle.} & \emph{Aaron \bfemph{broken} the unicycle.}\\
Island effects & \emph{Which \bfemph{bikes} is John fixing?} & \emph{Which is John fixing \bfemph{bikes}?}\\
NPI licensing & \emph{The truck has \bfemph{clearly} tipped over.} & \emph{The truck has \bfemph{ever} tipped over.}\\
Quantifiers & \emph{No boy knew \bfemph{fewer than} six guys.} & \emph{No boy knew \bfemph{at most} six guys.}\\
Subject-verb agr. & \emph{These casseroles \bfemph{disgust} Kayla.} & \emph{These casseroles \bfemph{disgusts} Kayla.}\\
\hline
\end{tabular}
\caption{Example acceptable and unacceptable sentences for the 12 types of linguistic phenomena in BLiMP \citep{blimp}.}\label{tab:blimp-table}
\end{table*}

The Benchmark of Linguistic Minimal Pairs (BLiMP; \citet{blimp}) extended the template generation method to 12 different syntactic phenomena and 67 different paradigms (i.e., sub-phenomena). These phenomena include agreement, argument structure, filler-gaps, and NPI licensing (Table \ref{tab:blimp-table} gives an example for each of the phenomena). They generated 1000 sentences for each paradigm using templates and a lexicon containing about 3,000 items, annotated with various grammatical features to ensure the validity of the generated sentences. They obtained human forced choice judgments from MTurk for quality assurance and to establish a human baseline, and evaluated several forward sequential models (n-gram, LSTM, Transformer-XL, and GPT-2), in an unsupervised setting.

\citet{hu-syntax-assessment} constructed a similar syntactic test suite of 34 linguistic phenomena, aiming to cover an introductory syntax textbook \citep{carnie-syntax}. However, rather than using binary minimal pairs as in BLiMP, \citet{hu-syntax-assessment} defined success criteria differently depending on the task, where the model is considered correct if its perplexity on the sentences simultaneously satisfies several pre-defined inequalities. This design allows more control over lexical confounds and task instances involving more than two sentences (for example, a 2x2 design is used for subject-verb agreement). They manipulated model architectures and the amount of training data, and found that the models' performance on their test suite is not always correlated with perplexity (a common evaluation metric for LM performance).

\subsection{Gradience of acceptability}

Most studies on neural linguistic acceptability so far have adopted a binary scale for acceptability: each sentence is assumed to be either acceptable or not; the model's response is either correct or incorrect. The final reported metric is either accuracy or Matthews correlation coefficient (MCC), both of which require a discrete threshold and do not allow a gradient response. \citet{niu-penn} criticized the use of accuracy and MCC for linguistic acceptability benchmarks because they ignore the magnitude of LM probabilities, thus obscuring the differences between the models that are meant to be compared.

Linguistic theories disagree about whether grammars should assign a binary of gradient value to sentence well-formedness. Traditional generative grammar theories \citep{chomsky1957, chomsky1965} assume sentence grammaticality is a binary property; any disagreement of native speakers is due to {\em performance} factors such as processing difficulty, while only {\em competence} (the stable grammatical knowledge that speakers possess) is relevant to linguistic theory. \citet{lau2017} provided empirical evidence of acceptability as a gradient phenomenon: they obtained a set of corrupted sentences by feeding English sentences through a round-trip machine translation procedure into another language and back. Then they obtained acceptability judgments on these sentences from MTurk and found that the distribution of ratings more closely resembled ratings of a continuous variable (body weight) than a discrete variable (integer parity). Then, they evaluated several language models on this dataset by computing the correlation between LM probabilities and human acceptability judgments, and found that several of the models correlated nearly as much with humans as human correlations with each other.

\citet{wilcox21} investigated the magnitude of surprisals when an LM encounters an ungrammatical section in a sentence. They constructed a test suite of syntactic minimal pairs similar to BLiMP, and collected human reaction data using the {\em Interpolated Maze} paradigm. In this maze task, human participants were presented with a sentence one word at a time, and had to choose between the correct word and a distractor at each step; the response time captures the processing difficulty and is expected to be higher for ungrammatical sections. They compared human response times to surprisals for several LMs, and found that the models consistently under-predicted the magnitude of human processing difficulty. Therefore, they concluded that LMs are less sensitive to syntactic violations than humans.

\section{Representational probes of LM embeddings} \label{sec:representational-probes}

\subsection{Layerwise probing}

Transformer language models have a relatively homogeneous architecture, generating a fixed-dimension vector representation at each layer for each input token. Depending on the use case, these vectors can be probed directly, or they can be collapsed into sentence vectors before probing by taking an average of the vectors for each token. The layerwise architecture makes it straightforward to probe for what linguistic information is contained in the embeddings at each layer.

\citet{bert-rediscovers} applied the edge probing method (Section \ref{sec:probing-classifiers}) on BERT embeddings on a variety of tasks, and found that the probe learned a higher weighting for the upper layers when the task was more semantic (e.g., semantic proto-roles and relation classification), while the middle layers were preferred for syntactic tasks (e.g., part-of-speech tagging and dependency arc labelling). This suggested that the upper layers derived complex semantic representations from simpler syntactic representations in the lower layers, similar to the stages in a traditional NLP pipeline. \citet{jawahar-probe} applied the SentEval toolkit \citep{senteval} to BERT embeddings, with a similar result: the lower layers were the best at capturing surface features, the middle layers contained the most syntactic information, and the upper layers contained the most semantics.

\citet{kelly-sentence-probe} probed BERT as well as some static word vector models for extractability of syntactic construction from a sentence embedding. The task was to determine which of two constructions were used in the input sentence, for example, a ditransitive dative or a prepositional dative. They measured spatial separability in the embedding space as well as probing classifier accuracy, and found that the middle layers of BERT were the most sensitive to the type of syntactic construction present in the sentence.

\citet{yu-ettinger} assessed several transformer LMs for their ability to represent compositional phrases. In their setup, phrasal representations were probed for paraphrase similarity: whether two short phrases had similar meaning or not. In order to control for lexical overlap, they experimented with AB-BA pairs, where the model must determine the similarity of a two-word phrase and its reversal (e.g., {\em law school} has low similarity with {\em school law}, whereas {\em adult female} has a similar meaning as {\em female adult}). Despite trying several different models and methods for generating phrase representations, they obtained poor results on the paraphrase similarity task when lexical overlap was controlled, indicating that the LMs have strong sensitivity to word content but not to nuanced composition.

\citet{miaschi2020linguistic} created a suite of probes for a wide range of linguistic features ranging from surface to lexical and syntactic, and probed layers of BERT for the ability to predict these linguistic features. They found that most features had the best performance in one of the middle layers; performance decreased in the upper layers, and dropped drastically in the last layer. They also experimented with fine-tuning: all linguistic features performed worse after the model was fine-tuned, especially the upper layers, agreeing with earlier results by \citet{liu-probing} that the upper layers become more specialized towards the specific task when fine-tuned.

Overall, the various layerwise probing experiments reach a similar conclusion about how linguistic information is propagated through the layers of transformer LMs. The initial layer is a stream of non-contextual token embeddings, so only word-level and surface features are available. The lower, middle, and upper layers gradually extract morphosyntactic and semantic features that serve as a good general-purpose representation of language and are useful to many different downstream tasks. Finally, the last layer aggregates information from the previous layers into a representation optimized for the specific task (i.e., masked token prediction in the case of pretrained LMs).

\subsection{Structural probes for syntax}

\citet{hewitt-syntax} proposed a {\em structural probe} for discovering syntax trees contained in contextual embeddings. Their probe measured the extent to which the depth of each token within a dependency parse tree can be uncovered via a linear transformation of their embeddings. This is sufficient to demonstrate the existence of dependency trees in the embeddings because a minimum spanning tree algorithm can be applied to extract the best (undirected) dependency tree.

Specifically, their structural probe finds a linear transformation matrix $\vect{B}$ such that the linear distance of $\vect{B}$ applied to embeddings $\vect{w_1}$ and $\vect{w_2}$ approximates their distance in the parse tree:
$$d_{\vect{B}}(\vect{w_1}, \vect{w_2}) = || \vect{B} \vect{w_1} - \vect{B} \vect{w_2}||_2.$$
The linear transformation $\vect{B}$ is learned via gradient descent by minimizing the total difference to the ground truth tree distance across all pairs of tokens $(\vect{w_1}, \vect{w_2})$ from the same sentence in a corpus. The evaluation metrics used were undirected unlabelled attachment score (UUAS) and the Spearman correlation of tree distances compared to the ground truth. They found that dependency trees could be recovered from BERT and ELMo embeddings, but not from the baselines. The middle layers of BERT performed the best for extracting dependency trees, and the performance increased as the rank of $\vect{B}$ approached 64 but increasing the rank beyond that did not further increase the performance. The authors thus concluded that syntactic information was encoded in a fairly low-rank subspace in BERT's embeddings.

\citet{chi-hewitt} extended the structural probe to a multilingual setting, using mBERT and data from Universal Dependencies (UD; \citet{universal-dependencies}) in 11 languages. They found that the same linear probe was able to reconstruct dependency tree distances in all languages, demonstrating that all languages share a common subspace for syntax in mBERT. Next, they applied a t-SNE visualization on vector representations of dependency arcs, and found that similar dependencies across languages were clustered together. This is surprising given that neither mBERT nor the structural probe had access to labelled dependency information during training.

\citet{white-nonlinear} extended the structural probing method to support nonlinear probes, recasting the probe as kernalized metric learning. This enabled the use of common nonlinear kernels such as the radial basis function (RBF) kernel, which yield nonlinear probes without increasing the probe complexity. They noted that the mathematical structure of the RBF kernel resembled BERT's self-attention, and hypothesized that this resemblance explained why the RBF kernel outperformed the linear one.

Although the dependency formalism (used by Universal Dependencies) is a popular syntactic framework in computational linguistics, it is only one among many proposed formalisms for syntax. Depending on which formalism is used in probing, we may draw different conclusions about LMs' syntactic capabilities. \citet{ud-vs-sud} compared the structural probe on UD and an alternative syntactic framework, Surface-Syntactic Universal Dependencies (SUD; \citep{sud}); they found that BERT and ELMo performed better with UD than SUD annotations in most languages. Similarly, in semantic role labelling, \citet{kuznetsov-framing} found that the results of probing differed depending on which formalism was used to annotate the data (PropBank, VerbNet, FrameNet, or Semantic Proto-Roles). Any linguistic probing work must commit to a particular formalism, thus probing research depends substantially on the underlying linguistic theory and the availability of data annotated in these frameworks.

\subsection{Probes involving LM training}

The most common approach to linguistic LM probing has so far been on pretrained models (examining either their outputs or internal representations). From a practical perspective, these experiments have the advantage that they can be performed without large amounts of data or computational resources. Nonetheless, some recent work have explored using pretraining or fine-tuning as a tool to gain insights about the linguistic properties of LMs. By incorporating LM training, these methods reveal which properties are learnable by a family of models or architectures, whereas static probing methods examine which capabilities have been learned by specific models such as BERT and RoBERTa.

Language models are trained on extremely large amounts of data, compared to what an average human is exposed to during language learning. For example, RoBERTa is trained on 30B tokens, while children are exposed to no more than 3-11M words of input a year \citep{30m-word-gap}, or somewhere on the order of 100M words by the time they reach puberty. \citet{zhang-billions} tested the performance of MiniBERTa models (\citet{mini-bertas}; variants of RoBERTa trained on 1M to 30B words) on a variety of benchmarks, and found that only about 10-100M words are sufficient to learn most syntactic features, but commonsense knowledge and most downstream tasks require much more data. A similar result was reported by \citet{probing-across-time}, who probed RoBERTa throughout its pretraining process, finding that linguistic knowledge was quickly acquired while commonsense and reasoning was only acquired much later. Thus, it appears that linguistic knowledge is relatively easy for LMs to acquire.

On the other hand, some studies have argued that LMs are still less efficient than humans at acquiring syntax. \citet{babyberta} trained BabyBERTa, a smaller version of RoBERTa using child-directed corpora as training data, and evaluated its syntactic capabilities on an adapted version of BLiMP where the vocabulary was replaced with appropriate child-level words. BabyBERTa had poor performance on the syntactic tests relative to RoBERTa, despite receiving a similar amount of input as a 6-year-old child. \Citet{van-schijndel-etal-2019-quantity} showed that most syntactic tasks can be learned from limited data, but some more complex tasks fall short of human performance, and even training on very large datasets does not improve performance to human levels.

Another experimental paradigm involving training is probing LMs for {\em inductive biases}. Human languages tend to have syntactic rules that operate on a structural level rather than the surface level (for example, subject-auxiliary inversion moves the structurally highest auxiliary, not the linearly last auxiliary), and \citet{chomsky1981} hypothesized that humans have an innate structural bias that helps them learn language from limited input.

\citet{warstadt-structural-bias} proposed a {\em poverty of stimulus} design for probing, where they fine-tuned BERT to classify grammaticality using sentences designed so that it is ambiguous whether a surface or structural rule is required. During test time, they used a different set of sentences to determine whether BERT has learned a surface or structural rule; they found that BERT prefers to learn structural over surface rules when both are equally probable. In a subsequent study, \citet{mini-bertas} pretrained MiniBERTa models on between 1M to 30B tokens to investigate the inductive biases of models of varying data size. They found that the smaller models preferred surface generalizations while larger models preferred structural generalizations, which may explain why larger LMs are so successful at downstream tasks but only after crossing a certain data threshold.

\section{Adapting psycholinguistics to LMs} \label{sec:psycholinguistic-probing}

Psycholinguistics is a field that shares many commonalities with LM probing, but seeks to understand human language processing rather than neural models. Both fields contend with the challenges of indirectly deducing the mechanisms of an entity whose internals are inaccessible or difficult to interpret; thus, psycholinguistics, an older field, provides a rich source of methods and data that can be applied to LM probing. Several differences make LM probing a generally easier endeavour than psycholinguistics: first, each neuron in an LM can be inspected in precise detail whereas human brains can only be imaged with relatively coarse-grained techniques such as EEG and fMRI; second, LMs can be cheaply run on large amounts of data with fully deterministic results. Still, methods from psycholinguistics sometimes require substantial modification to be suitable for LMs.

\subsection{The N400 response and surprisal}

A popular method of measuring human responses to language is using electroencephalography (EEG) to detect electrical activity via electrodes placed at the scalp. Event-related potentials (ERPs) are brain responses to specific stimuli derived from EEG signals, and are good indicators for tracking automatic responses during language processing. A well-known ERP is the N400 response, characterized by a negative potential roughly 400ms after a stimulus, and is generally associated with the stimulus being semantically anomalous with respect to the preceding context. The N400 response is not specific to language (for example, it has been observed using images or environmental sounds as stimuli), nor is it produced by all linguistic anomalies (for example, morphosyntactic violations do not trigger the N400); precisely what conditions trigger the N400 is still an open question \citep{kutas-federmeier}. Early psycholinguistic studies proposed that semantic anomalies produce the N400 response while syntactic anomalies produce the P600, a different type of ERP, but this dichotomy was challenged in later studies \citep{psycholinguistics-electrified}. Presently, the N400 is not known to be aligned with any single component in a linguistic theory.

The N400 is correlated with cloze probability, so it is often used as an approximate estimate of probability in humans. \citet{frank-erp-surprisal} found that the N400 was correlated with surprisals in a variety of RNN models. \citet{michaelov-n400} gathered a large set of psycholinguistic stimuli from different papers about the N400 response, and ran them on RNN models, comparing the LM surprisals with human responses. They confirmed that LM surprisals generally predicted N400 responses, but in some cases such as in morphosyntactic or event structure violations, the LM surprisals were more sensitive than the N400 in humans; thus, the N400 cannot be explained by surprisal alone.

\citet{ettinger-psycholinguistic} probed BERT using stimuli from three psycholinguistic studies involving commonsense knowledge, semantic role reversals, and negation. These studies were selected because they were cases where cloze probabilities were low, yet the N400 response did not trigger: in other words, they represented hard cases where automatic processing mechanisms could not reveal the extent of the surprisal and deliberate judgment is required. Ettinger found that BERT performed reasonably well on the commonsense reasoning task, was less sensitive than humans to role reversal anomalies, and failed completely at understanding negation (by filling in words of the matching category for prompts like {\em ``A robin is not a [MASK]``}).

\subsection{Priming in LMs}

Priming is another popular experimental paradigm in psycholinguistics, a phenomenon where exposure to prior stimuli (e.g., lexical items or structures) affects responses to a later stimuli \citep{pickering-priming}. In structural priming, when people are exposed to a particular syntactic structure, they are more likely to produce the same structure later \citep{bock90}, and are able to process sentences of the same structure more quickly. This effect is useful for understanding our cognitive mechanisms for language, since priming is evidence that two sentences share a common internal representation.

Structural and lexical priming have been explored for LM probing, but the priming methodology does not naturally carry over to LMs, which do not have any concept of temporality. \citet{misra-priming} simulated lexical priming by prepending the prime word (either by itself or situated in a carrier sentence) before the target sentence in which BERT predicts a masked word. They found that BERT was facilitated by a relevant prime word only when the target sentence was unconstrained, but when the target sentence was constrained, the prime word instead acted as a distractor and lowered the LM's probability for the correct word. A limitation of this method is that concatenating a prime and a target sentence creates an unnatural combination that does not occur in natural text, possibly leading to unpredictable out-of-distribution effects.

\citet{prasad19} formulated structural priming as model fine-tuning: to measure a priming effect, they proposed to fine-tune a model on a set of sentences with one syntactic structure, then measure the surprisal on another set of target sentences. A priming effect is considered to exist if the surprisal for the target sentences is lower after fine-tuning than the original LM; they found evidence of priming in LSTM models for various types of relative clauses.

\section{Using LMs as evidence for linguistic theory} \label{sec:lm-linguistic-theory}

During the past few years since the rise of transformer models, there have been an abundance of papers probing the linguistic capabilities of these models. Probing work owes a great deal to the linguistic theories that it is based upon; however, the contribution has so far mostly been unidirectional -- neural network probing papers have had negligible impact on theoretical linguistics \citep{baroni-proper-role}. Experiments involving LMs have given us an increasingly detailed view of how they process language, but these experiments cannot offer insights about how {\em humans} process language. Even when LMs exhibit similar linguistic responses to stimuli as human subjects, it is unclear what exactly are the implications for human linguistics, since the neural architectures share few similarities with the human brain. Yet recently, a small number of papers have proposed ways in which neural networks may contribute to linguistic theory.

\Citet{van-schijndel-single-stage} considered two theories explaining the processing difficulty of garden path sentences. In the traditional two-stage theory, readers maintain a partial parse while reading a sentence and are forced to reanalyze the parse in a garden path situation, causing a processing delay. In the more recent single-stage theory, readers maintain several parses at the same time, and there is a delay at each word from the processing required to integrate the word into all available parses. The authors used LSTMs to measure the surprisal of each word based on the preceding context, and found that the LSTM surprisals consistently under-predicted the magnitude of processing delays in garden path sentences. By using LSTM surprisals as a measure of predictability, this experiment provided empirical evidence against the single-stage theory, which predicts a linear relationship between reading time and predictability.

\citet{wilcox-island} studied the learnability of island constraints, where subtle constraints sometimes prevent movement of a wh-phrase. Linguistic nativism theories have argued that innate knowledge of universal grammar is necessary for children to learn island constraints from limited data, while opponents have denied this claim. The authors tested LMs on sensitivity to a variety of island constraints, finding that LMs are generally successful at this task. This constituted evidence against nativism theories, since LMs are general domain learners that cannot possibly possess innate knowledge of human grammar, yet were able to learn island constraints from data.

The next chapter presents my work on applying LMs toward theories of word class flexibility, contributing a case study that demonstrates the utility of LMs for linguistic theory.

\newcommand \NumFlexibleLanguages{25 }
\newcommand \NumLanguages{37 }

\chapter{Word class flexibility in LMs}

{\em The contents of this chapter are based on my previous publication \citep{word-class-flexibility}.}

\section{Introduction}

In this chapter, we present a computational methodology to quantify semantic regularities in word class flexibility using contextual word embeddings. Word class flexibility refers to the phenomenon whereby a single word form is used across different grammatical categories, and is considered one of the challenging topics in linguistic typology \citep{EvansLevinson2009}. For instance, the word \textit{buru} in Mundari can be used as a noun to denote `mountain', or as a verb to denote `to heap up' \citep{evans-osada}.

There is an extensive literature on how languages vary in word class flexibility, either directly \citep{Hengeveld1992, VogelComrie2000, vanLierRijkhoff2013} or through related notions such as word class conversion (with zero-derivation) \citep{Vonen1994, Don2003, BauerValera2005a, Manova2011, StekauerEtAl2012}. However, existing studies tend to rely on analyses of small sets of lexical items that may not be representative of word class flexibility in the broad lexicon. Critically lacking are systematic analyses of word class flexibility across many languages, and existing typological studies have only focused on qualitative comparisons of word class systems. 

We take to our knowledge the first step towards computational quantification of word class flexibility in \NumLanguages languages, taken from the Universal Dependencies project \citep{universal-dependencies}. We focus on lexical items that can be used both as nouns and as verbs, i.e., noun-verb flexibility. This choice is motivated by the fact that the distinction between nouns and verbs is the most stable in word class systems across languages: if a language makes any distinction between word classes at all, it will likely be a distinction between nouns and verbs \citep{Hengeveld1992, Evans2000, Croft2003}. However, our understanding of cross-linguistic regularity in noun-verb flexibility is impoverished.

We operationalize word class flexibility as a property of lemmas. We define a lemma as flexible if some of its occurrences are tagged as nouns and others as verbs. Flexible lemmas are sorted into noun dominant lemmas, which occur more frequently as nouns, and verb dominant lemmas that occur more frequently as verbs. Our methodology builds on contextualized word embedding models (e.g., ELMo \citep{elmo} and BERT \citep{bert}) to quantify semantic shift between grammatical classes of a lemma, within a single language. This methodology can also help quantify metrics of flexibility in the lexicon across  languages.

We use our methodology to address one of the most fundamental questions in the study of word class flexibility: should this phenomenon be analyzed as a directional word-formation process similar to derivation, or as a form of underspecification? Derived words are commonly argued to have a lower frequency of use and a narrower range in meaning compared to their base \citep{Marchand1964, Iacobini2000}. If word class flexibility is a directional process, we should expect that flexible lemmas are subject to more semantic variation in their dominant word class than in their less frequent class. We also test the claim that noun-to-verb flexibility  involves  more  semantic shift  than  verb-to-noun flexibility.  While previous work has explored these questions, it remains challenging to quantify semantic shift and semantic variation, particularly across different languages.

We present a novel probing task that reveals the ability of deep contextualized models to capture semantic information across word classes. Our utilization of deep contextual models predicts human judgment on the spectrum of noun-verb flexible usages including homonymy (unrelated senses), polysemy (different but related senses), and word class flexibility. We find that BERT outperforms ELMo and non-contextual word embeddings, and that the upper layers of BERT capture the most semantic information, which resonates with existing probing studies \citep{bert-rediscovers}. Our source code and data are available at: \url{https://github.com/SPOClab-ca/word-class-flexibility}.

\section{Linguistic background and assumptions}

\subsection{Types of flexibility}

The phenomenon of word class flexibility has been analyzed in different ways. One way is to assume the existence of underspecified word classes. For instance, \citet{Hengeveld2013} claims that basic lexical items in Mundari belong to a single class of \textit{contentives} that can be used to perform all the functions associated with nouns, verbs, adjectives or adverbs in a language like English. Alternatively, word class flexibility can be analyzed as a form of conversion, i.e., as a relation between words that have the same form and closely related senses but different word classes, such as \textit{a fish} and \textit{to fish} in English \citep{Adams1973}. Conversion has been analyzed as a derivational process that relates different lexemes \citep{Jespersen1924, Marchand1969, QuirkEtAl1985}, or as a property of lexemes whose word class is underspecified \citep{Farell2001, BarnerBale2002}. We use word class flexibility as a general term that subsumes these different notions. This allows us to assess whether there is evidence that word class flexibility should be characterized as a directional word formation process, rather than as a form of underspecification.

\subsection{Homonymy and polysemy}

\label{sec:polysemy-homonymy}

Word class flexibility has often been analyzed in terms of homonomy and polysemy \citep{ValeraRuz2020}. Homonymy is a relation between lexemes that share the same word form but are not semantically related \citep[][p.80]{Cruse1986}. Homonyms may differ in word class, such as \textit{ring} `a small circular band' and \textit{ring} `make a clear resonant or vibrating sound.' Polysemy is defined as a relation between different senses of a single lexeme (\textit{ibid.}). Insofar as the nominal and verbal uses of flexible lexical items are semantically related, one may argue that word class flexibility is similar to polysemy, and must be distinguished from homonymy. In practice,  homonymy and polysemy exist on a continuum, so it is difficult to apply a consistent criterion to differentiate them \citep{tuggy-polysemy}. As a consequence, we will not attempt to tease homonymy apart from word class flexibility.

Regarding morphology, word class flexibility excludes pairs of lexical items that are related by overt derivational affixes, such as \textit{to act}/\textit{an actor}. In such cases, word class alternations can be attributed to the presence of a derivational affix, and are therefore part of regular morphology. In contrast, we allow tokens of flexible lexical items to differ in inflectional morphology.

\subsection{Directionality of class conversion}

\label{sec:related-work-directionality}

Word class flexibility can be analyzed either as a static relation between nominal and verbal uses of a single lexeme, or as a word formation process related to derivation. The merits of each analysis have been extensively debated in the literature on conversion \citep[see e.g.,][]{Farell2001, Don2005}. One of the objectives of our study is to show that deep contextualized language models can be used to help resolve this debate. A hallmark of derivational processes is their directionality. Direction of derivation can be established using several synchronic criteria, among which are the principles that a derived form tends to have a lower frequency of use and a smaller range of senses than its base \citep{Marchand1964, Iacobini2000}. In languages where word class flexibility is a derivational process, one should therefore expect greater semantic variation when flexible lemmas are used in their dominant word class---an important issue that we verify  with our methodology.

A related phenomenon is the relationship between frequency and polysemy. Higher frequency words tend to have more senses as they are influenced to a greater extent by phonetic reduction and sense extension processes \citep{zipf-polysemy, fenk-polysemy}. In our work, we compare semantic variation between the noun and verb usages of a word rather than semantic variation across different words; the presence of a similar effect would constitute as evidence of word class flexibility as a derivational process.

Semantic variation has been operationalized in several ways. \citet{kisselew} uses an entropy-based metric, while \citet{balteiro} and \citet{bram-thesis} measure semantic variation by counting the number of different noun and verb senses in a dictionary. The latter study found that the more frequent word class has greater semantic variation at a rate above random chance. Here we propose a novel metric based on contextual word embeddings to compare the amount of semantic variation of flexible lemmas in their dominant and non-dominant grammatical classes. Differing from existing methods, our metric is validated explicitly on human judgements of semantic similarity, and can be applied to many languages without the need for dictionary resources.

\subsection{Asymmetry in semantic shift}

\label{sec:related-work-semantic-shift}

If word class flexibility is a directional process, a natural question is whether derived verbs stand in the same semantic relation to their base as derived nouns. The literature on conversion suggests that there might be significant differences between these two directions of derivation. In English, verbs that are derived from nouns by conversion have been argued to describe events that include the noun's denotation as a participant (e.g. \textit{hammer}, `to hit something with a hammer') or as a spatio-temporal circumstance (\textit{winter} `to spend the winter somewhere'). \citet{clark-clark} argue that the semantic relations between denominal verbs and their base are so varied that they cannot be given a unified description. In comparison, when the base of conversion is a verb, the derived noun most frequently denotes an event of the sort described by the verb (e.g. \textit{throw} `the act of throwing something'), or the result of such an act (e.g. \textit{release} `state of being set free') \citep{Jespersen1942, Marchand1969, Cetnarowska1993}. This has led some authors to suggest that verb to noun conversion in English involves less semantic shift than noun to verb conversion \citep[][p.22]{Bauer2005}. Here we consider a new metric of semantic shift based on contextual embeddings, and we use this metric to test the hypothesis that the expected semantic shift involved in word class flexibility is greater for noun dominant lexical items (as compared to verb dominant lexical items) in our sample of languages. As we will show, this proposal is consistent with the empirical observation that verb-to-noun conversion is statistically more salient than noun-to-verb conversion.

\section{Identification of word class flexibility}

\subsection{Definitions}
\label{sec:definitions}

A lemma is {\em flexible} if it can be used both as a noun and as a verb. To reduce noise, we require each lemma to appear at least 10 times and at least 5\% of the time as the minority class to be considered flexible. The {\em inflectional paradigm} of a lemma is the set of words that have the lemma.

A flexible lemma is {\em noun (verb) dominant} if it occurs more often as a noun (verb) than as a verb (noun). This is merely an empirical property of a lemma: we do not claim that the base POS should be determined by frequency. The {\em noun (verb) flexibility} of a language is the proportion of noun (verb) dominant lemmas that are flexible.

\subsection{Datasets and preprocessing}

Our experiments require corpora containing part-of-speech annotations. For English, we use the British National Corpus (BNC), consisting of 100M words of written and spoken English from a variety of sources \citep{bnc}. Root lemmas and POS tags are provided, and were generated automatically using the CLAWS4 tagger \citep{claws4}. For our experiments, we use BNC-baby, a subset of BNC containing 4M words.

For other languages, we use the Universal Dependencies (UD) treebanks of over 70 languages, annotated with lemmatizations, POS tags, and dependency information \citep{universal-dependencies}. We concatenate the treebanks for each language and use the languages that have at least 100k tokens.

The UD treebanks are too small for our contextualized experiments and are not matched for content and style, so we supplement them with Wikipedia text\footnote{We use Wikiextractor to extract text from Wikimedia dumps: \url{https://github.com/attardi/wikiextractor}.}. For each language, we randomly sample 10M tokens from Wikipedia; we then use UDPipe 1.2 \citep{udpipe} to tokenize the text and generate POS tags for every token. We do not use the lemmas provided by UDPipe, but instead use the lemma merging algorithm to group lemmas.

\subsection{Lemma merging algorithm}

\begin{table}
    \small
    \centering
\begin{tabular}{|l|l|l|l|l|l|}
\hline
\textbf{Language} & \textbf{Nouns} & \textbf{Verbs} & \textbf{\begin{tabular}[c]{@{}l@{}}Noun\\ flexibility\end{tabular}} & \textbf{\begin{tabular}[c]{@{}l@{}}Verb\\ flexibility\end{tabular}} \\ \hline
Arabic & 1517 & 299 & 0.076 & 0.221 \\ \hline
Bulgarian & 786 & 343 & 0.039 & 0.047 \\ \hline
Catalan & 1680 & 590 & 0.039 & 0.147 \\ \hline
Chinese & 1325 & 634 & 0.125 & 0.391 \\ \hline
Croatian & 1031 & 370 & 0.042 & 0.062 \\ \hline
Danish & 324 & 216 & 0.108 & 0.269 \\ \hline
Dutch & 958 & 441 & 0.077 & 0.188 \\ \hline
English & 1700 & 600 & 0.248 & 0.472 \\ \hline
Estonian & 1949 & 592 & 0.032 & 0.115 \\ \hline
Finnish & 1523 & 631 & 0.028 & 0.136 \\ \hline
French & 1844 & 649 & 0.062 & 0.257 \\ \hline
Galician & 802 & 334 & 0.031 & 0.135 \\ \hline
German & 4239 & 1706 & 0.049 & 0.229 \\ \hline
Hebrew & 850 & 315 & 0.111 & 0.321 \\ \hline
Indonesian & 572 & 243 & 0.052 & 0.128 \\ \hline
Italian & 2227 & 770 & 0.067 & 0.256 \\ \hline
Japanese & 1105 & 417 & 0.178 & 0.566 \\ \hline
Korean & 1890 & 1003 & 0.026 & 0.048 \\ \hline
Latin & 1090 & 885 & 0.056 & 0.122 \\ \hline
Norwegian & 1951 & 636 & 0.072 & 0.259 \\ \hline
Old Russian & 527 & 416 & 0.034 & 0.060 \\ \hline
Polish & 2054 & 1084 & 0.069 & 0.427 \\ \hline
Portuguese & 1711 & 638 & 0.037 & 0.185 \\ \hline
Romanian & 1809 & 740 & 0.060 & 0.151 \\ \hline
Slovenian & 746 & 316 & 0.068 & 0.123 \\ \hline
Spanish & 2637 & 873 & 0.046 & 0.202 \\ \hline
Swedish & 784 & 384 & 0.038 & 0.109 \\ \hline \hline
\multicolumn{5}{|l|}{{\bf Excluded Languages}} \\ \hline
Ancient Greek & 1098 & 1022 & 0.015 & 0.026 \\ \hline
Basque & 650 & 247 & 0.020 & 0.105 \\ \hline
Czech & 5468 & 2063 & 0.004 & 0.011 \\ \hline
Hindi & 1364 & 133 & 0.019 & 0.135 \\ \hline
Latvian & 1159 & 603 & 0.022 & 0.061 \\ \hline
Persian & 1125 & 47 & 0.010 & 0.234 \\ \hline
Russian & 3909 & 1760 & 0.005 & 0.024 \\ \hline
Slovak & 488 & 281 & 0.006 & 0.011 \\ \hline
Ukrainian & 659 & 238 & 0.006 & 0.029 \\ \hline
Urdu & 722 & 51 & 0.018 & 0.216 \\ \hline
\end{tabular}
    \caption{Noun and verb flexibility for 37 languages with at least 100k tokens in the UD corpus. We  include the 27 languages with over 2.5\% noun and verb flexibility; 10 languages are excluded from further analysis.}
    \label{tab:ud-frequency}
\end{table}

The UD corpus provides lemma annotations for each word, but these lemmas are insufficient for our purposes because they do not always capture instances of flexibility. In some languages, nouns and verbs are lemmatized to different forms by convention. For example, in French, the word \textit{voyage} can be used as a verb (\textit{il voyage} `he travels') or as a noun (\textit{un voyage} `a trip'). However, verbs are lemmatized to the infinitive \textit{voyager}, whereas nouns are lemmatized to the singular form \textit{voyage}. Since the noun and verb lemmas are different, it is not easy to identify them as having the same stem.

The different lemmatization conventions of French and English reflect a more substantial linguistic difference. French has a stem-based morphology, in which stems tend to occur with an inflectional ending. By contrast, English has a word-based morphology, where stems are commonly used as free forms \citep{Kastovsky2006}. This difference is relevant to the definition of word class flexibility: in stem-based systems, flexible items are stems that may not be attested as free forms \citep[][p.14]{BauerValera2005b}.

We propose a heuristic algorithm to capture stem-based flexibility as well as word-based flexibility. The key observation is that the inflectional paradigms of the noun and verb forms often have some words in common (such is the case for \textit{voyager}). Thus, we merge any two lemmas whose inflectional paradigms have a nonempty intersection. This is implemented with a single pass through the corpus, using the union-find data structure: for every word, we call UNION on the inflected form and the lemmatized form.

Using this heuristic, we can identify cases of flexibility that do not share the same lemma in the UD corpus (Table \ref{tab:ud-frequency}). This method is not perfect, and is unable to identify cases of stem-based flexibility where the inflectional paradigms don't intersect, for example in French, \textit{chant} `song' and \textit{chants} `songs' are not valid inflections of the verb \textit{chanter} `to sing'. There are also false positives that cause two unrelated lemmas to be merged if their inflectional paradigms intersect, for example, \textit{avions} (plural form of \textit{avion} `airplane') happens to have the same form as \textit{avions} (first person plural imperfect form of \textit{avoir} `to have').

\section{Methodology and evaluation}

\subsection{Probing test of contextualized model}

We now utilize contextual word embeddings from language models ELMo, BERT, mBERT, and XLM-R (described in Sections \ref{sec:rnn-lstm} and \ref{sec:transformers}) towards word class flexibility. Contextual embeddings can capture a variety of information other than semantics, which can introduce noise into our results, for example: the lexicographic form of a word, syntactic position, etc. In order to compare different contextual language models on how well they capture semantic information, we perform a probing test of how accurate the models can capture human judgements of word sense similarity.

\begin{table*}[]
    \centering
\begin{tabular}{||l|l|l|l||l|l|l|l||l|l|l|l||}
\hline
\textbf{Word} & \textbf{$|N|$} & \textbf{$|V|$} & \textbf{Sim} & \textbf{Word} & \textbf{$|N|$} & \textbf{$|V|$} & \textbf{Sim} & \textbf{Word} & \textbf{$|N|$} & \textbf{$|V|$} & \textbf{Sim} \\ \hline
aim & 137 & 98 & 2.0 & change & 889 & 858 & 1.6 & force & 470 & 188 & 0.8 \\ \hline
answer & 480 & 335 & 2.0 & claim & 222 & 239 & 1.6 & grant & 108 & 87 & 0.8 \\ \hline
attempt & 302 & 214 & 2.0 & cut & 92 & 488 & 1.6 & note & 287 & 361 & 0.8 \\ \hline
care & 403 & 249 & 2.0 & demand & 169 & 142 & 1.6 & sense & 536 & 88 & 0.8 \\ \hline
control & 519 & 179 & 2.0 & design & 246 & 153 & 1.6 & tear & 124 & 89 & 0.8 \\ \hline
cost & 234 & 192 & 2.0 & experience & 522 & 150 & 1.6 & account & 337 & 122 & 0.6 \\ \hline
count & 143 & 220 & 2.0 & hope & 114 & 571 & 1.6 & act & 644 & 268 & 0.6 \\ \hline
damage & 270 & 82 & 2.0 & increase & 252 & 399 & 1.6 & back & 764 & 88 & 0.6 \\ \hline
dance & 81 & 97 & 2.0 & judge & 80 & 96 & 1.6 & face & 1185 & 281 & 0.6 \\ \hline
doubt & 261 & 132 & 2.0 & limit & 125 & 134 & 1.6 & hold & 130 & 1251 & 0.6 \\ \hline
drink & 456 & 315 & 2.0 & load & 230 & 87 & 1.6 & land & 393 & 123 & 0.6 \\ \hline
end & 1171 & 244 & 2.0 & offer & 93 & 489 & 1.6 & lift & 100 & 165 & 0.6 \\ \hline
escape & 95 & 111 & 2.0 & rise & 164 & 283 & 1.6 & matter & 572 & 294 & 0.6 \\ \hline
estimate & 96 & 118 & 2.0 & smoke & 128 & 100 & 1.6 & order & 841 & 133 & 0.6 \\ \hline
fear & 209 & 99 & 2.0 & start & 159 & 1269 & 1.6 & place & 1643 & 341 & 0.6 \\ \hline
glance & 101 & 161 & 2.0 & step & 401 & 167 & 1.6 & press & 130 & 188 & 0.6 \\ \hline
help & 200 & 897 & 2.0 & study & 1037 & 211 & 1.6 & roll & 135 & 201 & 0.6 \\ \hline
influence & 204 & 150 & 2.0 & support & 290 & 292 & 1.6 & sort & 1613 & 216 & 0.6 \\ \hline
lack & 194 & 107 & 2.0 & trust & 90 & 126 & 1.6 & fire & 444 & 89 & 0.4 \\ \hline
link & 147 & 176 & 2.0 & waste & 103 & 98 & 1.6 & form & 1272 & 354 & 0.4 \\ \hline
love & 495 & 573 & 2.0 & work & 1665 & 1593 & 1.6 & notice & 115 & 387 & 0.4 \\ \hline
move & 131 & 1272 & 2.0 & base & 109 & 378 & 1.4 & play & 185 & 1093 & 0.4 \\ \hline
name & 960 & 112 & 2.0 & cover & 137 & 399 & 1.4 & turn & 226 & 1566 & 0.4 \\ \hline
need & 587 & 2350 & 2.0 & plant & 591 & 82 & 1.4 & wave & 402 & 120 & 0.4 \\ \hline
phone & 382 & 238 & 2.0 & run & 152 & 999 & 1.4 & cross & 102 & 215 & 0.2 \\ \hline
plan & 321 & 161 & 2.0 & stress & 159 & 106 & 1.4 & deal & 191 & 315 & 0.2 \\ \hline
question & 1285 & 96 & 2.0 & approach & 409 & 175 & 1.2 & hand & 1765 & 127 & 0.2 \\ \hline
rain & 182 & 92 & 2.0 & cause & 237 & 530 & 1.2 & present & 219 & 353 & 0.2 \\ \hline
result & 752 & 206 & 2.0 & match & 110 & 123 & 1.2 & set & 387 & 652 & 0.2 \\ \hline
return & 138 & 441 & 2.0 & miss & 320 & 410 & 1.2 & share & 104 & 232 & 0.2 \\ \hline
search & 215 & 163 & 2.0 & process & 720 & 91 & 1.2 & sign & 284 & 121 & 0.2 \\ \hline
sleep & 171 & 291 & 2.0 & shift & 96 & 104 & 1.2 & suit & 162 & 108 & 0.2 \\ \hline
smell & 141 & 149 & 2.0 & show & 132 & 1843 & 1.2 & wind & 189 & 82 & 0.2 \\ \hline
smile & 211 & 422 & 2.0 & sound & 313 & 496 & 1.2 & address & 257 & 148 & 0.0 \\ \hline
talk & 119 & 1302 & 2.0 & dress & 191 & 196 & 1.0 & bear & 110 & 394 & 0.0 \\ \hline
use & 791 & 2801 & 2.0 & lead & 107 & 716 & 1.0 & head & 1355 & 96 & 0.0 \\ \hline
view & 811 & 102 & 2.0 & light & 669 & 124 & 1.0 & mind & 736 & 620 & 0.0 \\ \hline
visit & 136 & 203 & 2.0 & look & 699 & 5893 & 1.0 & park & 179 & 105 & 0.0 \\ \hline
vote & 124 & 93 & 2.0 & mark & 562 & 198 & 1.0 & point & 1534 & 469 & 0.0 \\ \hline
walk & 144 & 914 & 2.0 & measure & 226 & 223 & 1.0 & ring & 185 & 387 & 0.0 \\ \hline
dream & 254 & 107 & 1.8 & rest & 414 & 132 & 1.0 & square & 225 & 82 & 0.0 \\ \hline
record & 1057 & 276 & 1.8 & tie & 82 & 112 & 1.0 & state & 471 & 156 & 0.0 \\ \hline
report & 313 & 331 & 1.8 & break & 117 & 519 & 0.8 & stick & 109 & 294 & 0.0 \\ \hline
test & 273 & 126 & 1.8 & charge & 392 & 115 & 0.8 & store & 95 & 158 & 0.0 \\ \hline
touch & 145 & 271 & 1.8 & drive & 88 & 476 & 0.8 & train & 224 & 94 & 0.0 \\ \hline
call & 209 & 1558 & 1.6 & focus & 92 & 168 & 0.8 & watch & 119 & 940 & 0.0 \\ \hline
\end{tabular}
    \caption{138 flexible words in English (top in BNC corpus) and human similarity scores, average of 5 ratings.}
    \label{tab:english-lemma-ratings}
\end{table*}

We begin with a list of the 138 most frequent flexible words in the BNC corpus. Some of these words are flexible (e.g., \textit{work}), while others are homonyms (e.g., \textit{bear}). For each lemma, we get five human annotators from Mechanical Turk to make a sentence using the word as a noun, then make a sentence using the word as a verb, then rate the similarity of the noun and verb senses on a scale from 0 to 2. The sentences are used for quality assurance, so that ratings are removed if the sentences are nonsensical. We will call the average human rating for each word the {\em human similarity score}; Table \ref{tab:english-lemma-ratings} shows the average ratings for all 138 words.

Next, we evaluate each layer of ELMo, BERT, mBERT, and XLM-R\footnote{We use the models `bert-base-uncased', `bert-base-multilingual-cased', and `xlm-roberta-base' from \citet{huggingface}.} on correlation with the human similarity score. That is, we compute the mean of the contextual vectors for all noun instances of the given word in the BNC corpus, the mean across all verb instances, then compute the cosine distance between the two mean vectors as the model's similarity score. Finally, we evaluate the Spearman correlation of the human and model's similarity scores for 138 words: this score measures the model's ability to gauge the level of semantic similarity between noun and verb senses, compared to human judgements.

For a baseline, we do the same procedure using non-contextual GloVe embeddings \citep{glove}. Note that while all instances of the same word have a static embedding, different words that share the same lemma still have different embeddings (e.g., \textit{work} and \textit{works}), so that the baseline is not trivial.

\begin{figure}
    \centering
    \includegraphics[width=0.5\linewidth]{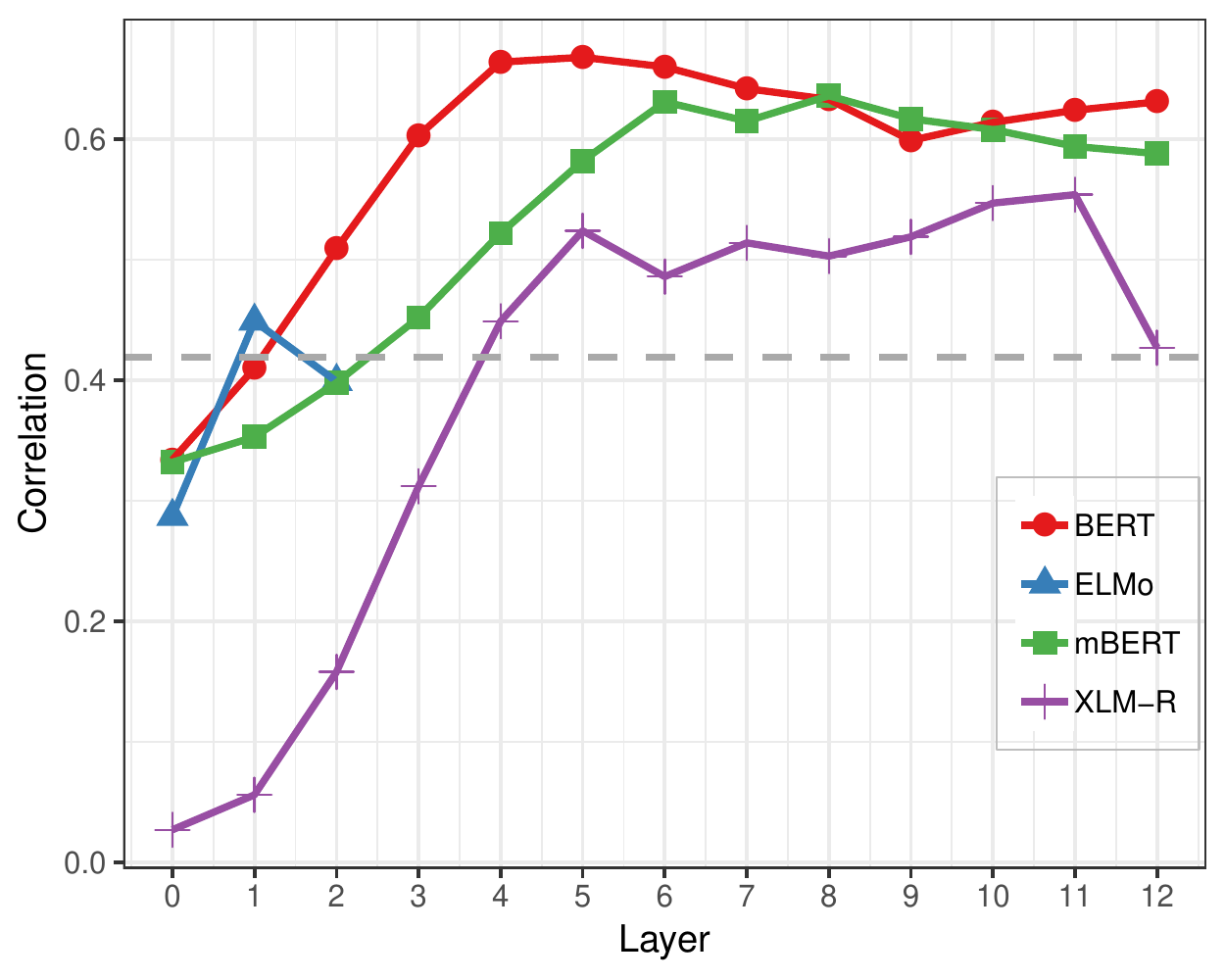}
    \caption{Spearman correlations between human and model similarity scores for ELMo, BERT, mBERT, and XLM-R. The dashed line is the baseline using static GloVe embeddings.}
    \label{fig:mturk-correlations}
\end{figure}

The correlations are shown in Figure \ref{fig:mturk-correlations}. BERT and mBERT are better than ELMo and XLM-R at capturing semantic information, in all transformer models, the correlation increases for each layer up until layer 4 or so, and after this point, the performance neither improves nor degrades in higher layers. Thus, unless otherwise noted, we use the final layers of each model for downstream tasks.

\begin{figure}
    \centering
    \includegraphics[width=0.5\linewidth]{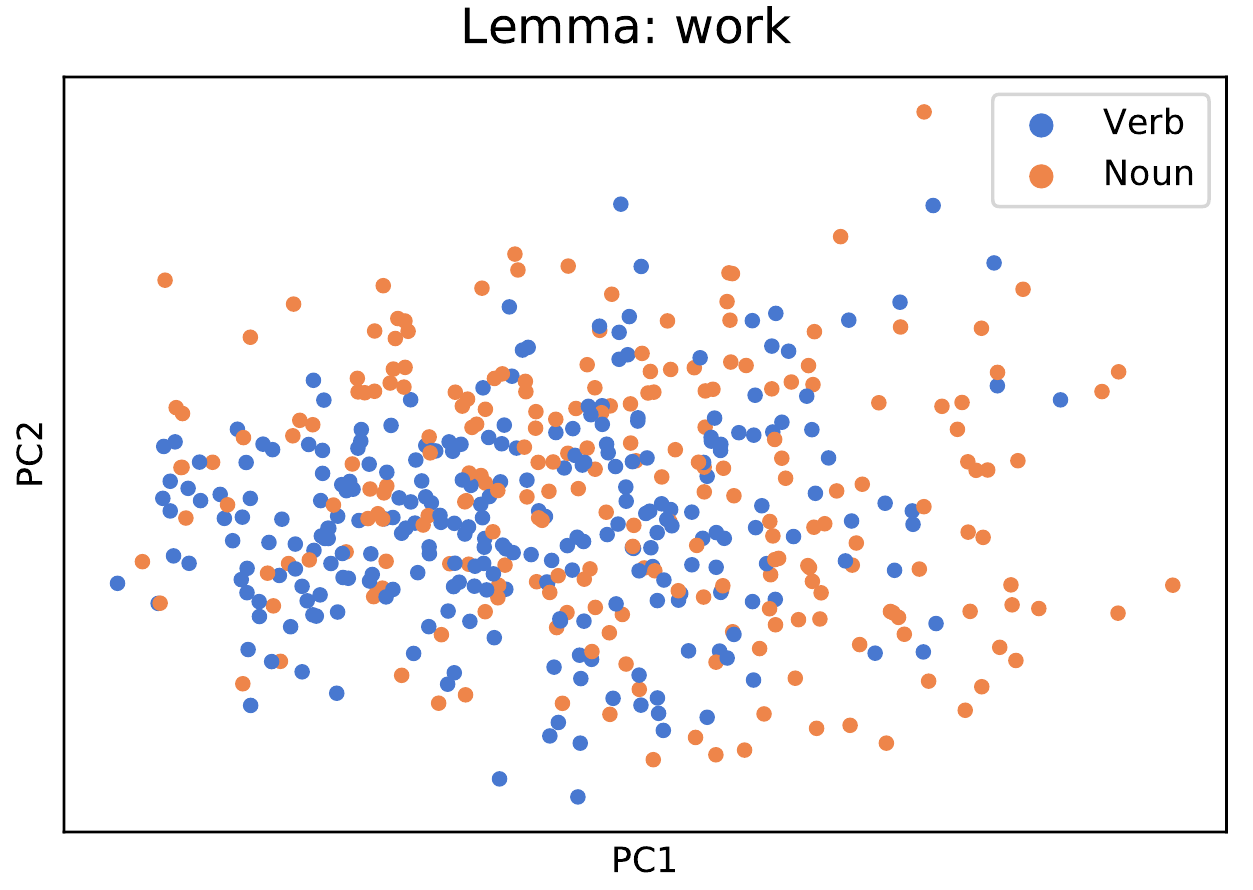}
    \includegraphics[width=0.5\linewidth]{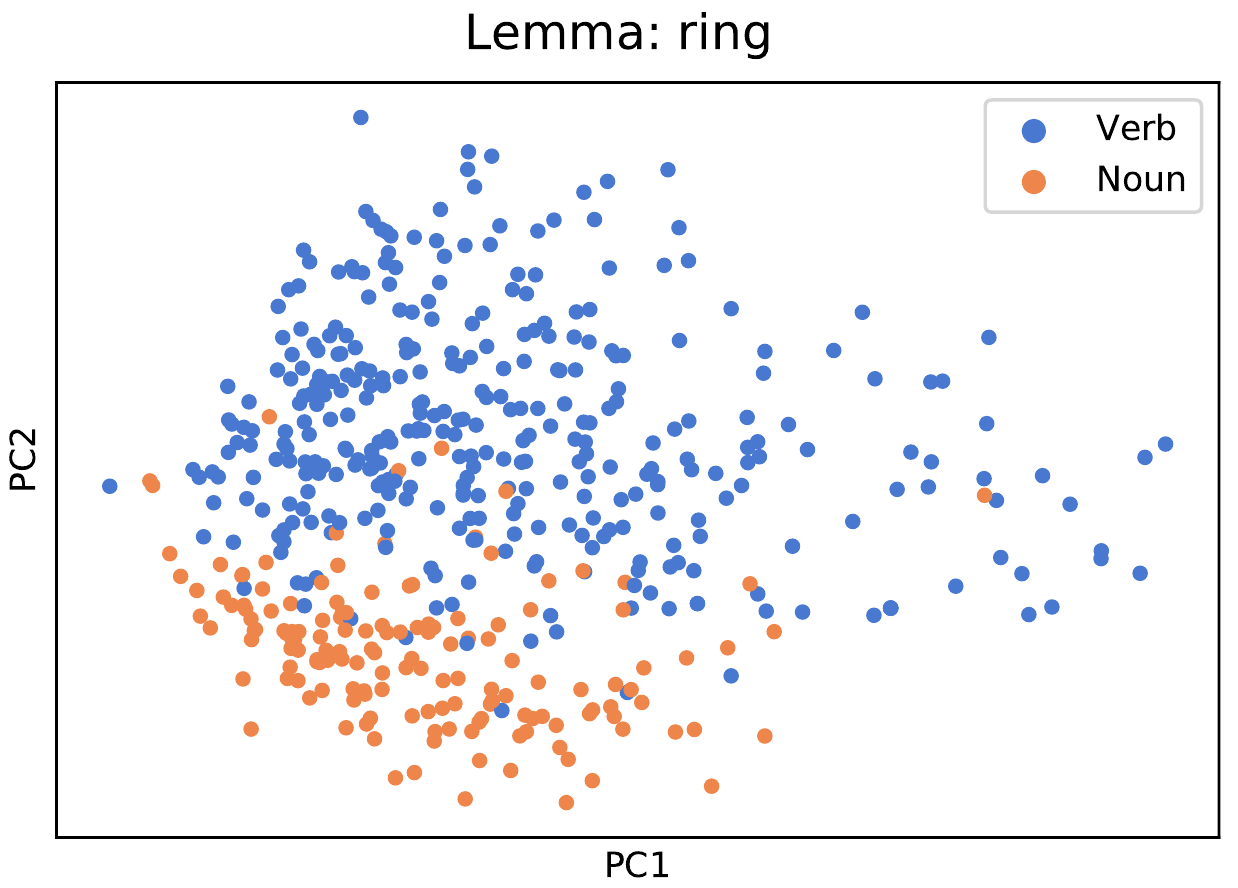}
    \caption{PCA plot of BERT embeddings for the lemmas ``work'' (high similarity between noun and verb senses) and ``ring'' (low similarity).}
    \label{fig:bert-work}
\end{figure}

Figure \ref{fig:bert-work} illustrates the contextual distributions for two lemmas on the opposite ends of the noun-verb similarity spectrum: \textit{work} (human similarity score: 2) and \textit{ring} (human similarity score: 0). We apply PCA to the BERT embeddings of all instances of each lemma in the BNC corpus. For \textit{work}, the noun and verb senses are very similar and the distributions have high overlap. In contrast, for \textit{ring}, the most common noun sense (`a circular object') is etymologically and semantically unrelated to the most common verb sense (`to produce a resonant sound'), and accordingly, their distributions have very little overlap.

\begin{table}
    \centering
    \begin{tabular}{|l||l|l||l|l||l|l|}
    \hline
    \textbf{Language} & \textbf{N$\to$V shift} & \textbf{V$\to$N shift} & \textbf{\begin{tabular}[c]{@{}l@{}}Noun\\ variation\end{tabular}} & \textbf{\begin{tabular}[c]{@{}l@{}}Verb\\ variation\end{tabular}} & \textbf{\begin{tabular}[c]{@{}l@{}}Majority\\ variation\end{tabular}} & \textbf{\begin{tabular}[c]{@{}l@{}}Minority\\ variation\end{tabular}} \\ \hline
    Arabic & 0.098 & 0.109 & 8.268 & 8.672$^{***}$ & 8.762$^{***}$ & 8.178 \\ \hline
    Bulgarian & 0.146 & 0.136 & 8.267 & 8.409 & 8.334 & 8.341 \\ \hline
    Catalan & 0.165 & 0.169 & 8.165 & 8.799$^{***}$ & 8.720$^{***}$ & 8.244 \\ \hline
    Chinese & 0.072 & 0.070 & 7.024 & 7.212$^{***}$ & 7.170$^{***}$ & 7.067 \\ \hline
    Croatian & 0.093 & 0.144$^{**}$ & 8.149 & 8.109 & 8.219$^{**}$ & 8.037 \\ \hline
    Danish & 0.103 & 0.110 & 8.245 & 8.338 & 8.438$^{***}$ & 8.146 \\ \hline
    Dutch & 0.146 & 0.174 & 7.716 & 8.786$^{***}$ & 8.354$^{*}$ & 8.148 \\ \hline
    English & 0.175$^{*}$ & 0.160 & 8.035 & 8.624$^{***}$ & 8.390$^{***}$ & 8.268 \\ \hline
    Estonian & 0.105 & 0.103 & 7.800 & 7.902 & 8.022$^{**}$ & 7.679 \\ \hline
    Finnish & 0.100 & 0.114 & 7.972 & 7.854 & 8.181$^{***}$ & 7.644 \\ \hline
    French & 0.212 & 0.204 & 8.189 & 9.472$^{***}$ & 9.082$^{***}$ & 8.578 \\ \hline
    Galician & 0.111 & 0.117 & 7.922 & 8.340$^{***}$ & 8.137 & 8.127 \\ \hline
    German & 0.382 & 0.355 & 8.078 & 9.758$^{***}$ & 9.096$^{**}$ & 8.740 \\ \hline
    Hebrew & 0.121 & 0.130 & 8.096 & 9.116$^{***}$ & 8.574 & 8.638 \\ \hline
    Indonesian & 0.034 & 0.048 & 7.100 & 7.076 & 7.076 & 7.101 \\ \hline
    Italian & 0.207 & 0.184 & 8.520 & 9.345$^{***}$ & 9.149$^{***}$ & 8.716 \\ \hline
    Japanese & 0.061 & 0.057 & 7.419$^{***}$ & 7.173 & 7.309 & 7.283 \\ \hline
    Latin & 0.092 & 0.139$^{***}$ & 7.920$^{***}$ & 7.710 & 7.905$^{***}$ & 7.724 \\ \hline
    Norwegian & 0.133 & 0.132 & 8.112 & 8.336$^{***}$ & 8.332$^{***}$ & 8.116 \\ \hline
    Polish & 0.090 & 0.080 & 8.318 & 8.751$^{***}$ & 8.670$^{***}$ & 8.399 \\ \hline
    Portuguese & 0.186 & 0.155 & 7.907 & 8.921$^{***}$ & 8.642$^{***}$ & 8.187 \\ \hline
    Romanian & 0.175 & 0.145 & 8.682 & 8.658 & 8.934$^{***}$ & 8.406 \\ \hline
    Slovenian & 0.093 & 0.113 & 8.046 & 7.983 & 8.177$^{***}$ & 7.853 \\ \hline
    Spanish & 0.235 & 0.214 & 7.898 & 8.961$^{***}$ & 8.691$^{***}$ & 8.168 \\ \hline
    Swedish & 0.088 & 0.082 & 8.262$^{*}$ & 8.147 & 8.328$^{***}$ & 8.081 \\ \hline
    {\bf Overall} & {\bf 1 of 3} & {\bf 2 of 3} & {\bf 3 of 17} & {\bf 14 of 17} & {\bf 20 of 20} & {\bf 0 of 20} \\ \hline

    \end{tabular}
    \caption{Semantic metrics for 25 languages, computed using mBERT and 10M tokens of Wikipedia text for each language. Asterisks denote significance at $^{*}p < 0.05$, $^{**}p < 0.01$, $^{***}p<0.001$. For the ``Overall'' row, we count the languages with a significant tendency towards one direction, out of the number of languages with statistical significance towards either direction (with $p < 0.05$ treated as significant).}
    \label{tab:mbert-semantic-stats}
\end{table}

\subsection{Three contextual metrics}

We define three metrics based on contextual embeddings to measure various semantic aspects of word class flexibility. We start by generating contextual embeddings for each occurrence of every flexible lemma. For each lemma $l$, let $E_{n, l}$ and $E_{v, l}$ be the set of contextual embeddings for noun and verb instances of $l$. We define the {\em prototype noun vector} $\vec{p}_{n, l}$ of a lemma $l$ as the mean of embeddings across noun instances, and the {\em noun variation} $V_{n, l}$ as the mean Euclidean distance from each noun instance to the noun vector:
\begin{align}
    & \vec{p}_{n,l} = \frac{1}{|E_{n, l}|} \sum_{\vec{x} \in E_{n,l}} \vec{x} \\
    & V_{n, l} = \frac{1}{|E_{n, l}|} \sum_{\vec{x} \in E_{n, l}} ||\vec{x} - \vec{p}_{n, l}||
\end{align}

The {\em prototype verb vector} $\vec{p}_{v, l}$ and {\em verb variation} $V_{v, l}$ for a lemma $l$ are defined similarly:
\begin{align}
    & \vec{p}_{v,l} = \frac{1}{|E_{v, l}|} \sum_{\vec{x} \in E_{v,l}} \vec{x} \\
    & V_{v, l} = \frac{1}{|E_{v, l}|} \sum_{\vec{x} \in E_{v, l}} ||\vec{x} - \vec{p}_{v, l}||
\end{align}

Lemmas are included if they appear at least 30 times as nouns and 30 times as verbs. To avoid biasing the variation metric towards the majority class, we downsample the majority class to be of equal size as the minority class before computing the variation. The method does not filter out pairs of lemmas that are arguably homonyms rather than flexible (section \ref{sec:polysemy-homonymy}); we choose to include all of these instances rather than set an arbitrary cutoff threshold.

We now define language-level metrics to measure the asymmetries hypothesized in sections \ref{sec:related-work-directionality} and \ref{sec:related-work-semantic-shift}. The {\em noun-to-verb shift (NVS)} is the average cosine distance between the prototype noun and verb vectors for noun dominant lemmas, and the {\em verb-to-noun shift (VNS)} likewise for verb dominant lemmas:
\begin{align}
    & NVS = 1 - \mathbb{E}_{l \textrm{ noun-dominant}} [\cos(\vec{p}_{n, l}, \vec{p}_{v, l})] \\
    & VNS = 1 - \mathbb{E}_{l \textrm{ verb-dominant}} [\cos(\vec{p}_{n, l}, \vec{p}_{v, l})]
\end{align}

We define the {\em noun (verb) variation} of a language as the average of noun (verb) variations across all lemmas. Finally, define the {\em majority variation} of a language as the average of the variation of the dominant POS class, and the {\em minority variation} as the average variation of the smaller POS class, across all lemmas.

\begin{table}
    \centering
    \begin{tabular}{|c|l||l|l||l|l||l|l|}
    \hline
    \textbf{Dataset} & \textbf{Model} & \textbf{N$\to$V shift} & \textbf{V$\to$N shift} & \textbf{\begin{tabular}[c]{@{}l@{}}Noun\\ variation\end{tabular}} & \textbf{\begin{tabular}[c]{@{}l@{}}Verb\\ variation\end{tabular}} & \textbf{\begin{tabular}[c]{@{}l@{}}Majority\\ variation\end{tabular}} & \textbf{\begin{tabular}[c]{@{}l@{}}Minority\\ variation\end{tabular}} \\ \hline
    \multirow{4}{*}{BNC} & ELMo & 0.389$^{*}$  & 0.357 & 20.261  & 20.455 & 20.329 & 20.388  \\ \cline{2-8}
    & BERT & 0.122$^{*}$ & 0.112 & 9.015 & 9.074 & 9.100$^{***}$ & 8.989 \\ \cline{2-8}
    & mBERT & 0.189$^{*}$ & 0.169 & 7.211 & 8.401$^{***}$ & 7.875$^{**}$ & 7.717 \\ \cline{2-8}
    & XLM-R & 0.004 & 0.005 & 2.058 & 2.374$^{***}$ & 2.262 & 2.170 \\ \hline \hline
    \multirow{4}{*}{Wikipedia}& ELMo & 0.339$^{***}$  & 0.330 & 22.556 & 22.521 & 22.463 & 22.614$^{*}$  \\ \cline{2-8}
    & BERT & 0.120$^{***}$ & 0.100 & 9.218$^{***}$ & 8.944 & 9.118$^{**}$ & 9.044 \\ \cline{2-8}
    & mBERT & 0.175$^{*}$ & 0.160 & 8.035 & 8.624$^{***}$ & 8.390$^{***}$ & 8.268 \\ \cline{2-8}
    & XLM-R & 0.004$^{**}$ & 0.003 & 1.966 & 1.954 & 1.946 & 1.974 \\ \hline
    \end{tabular}
    \caption{Comparison of semantic models on BNC and Wikipedia datasets (English), computed using several different language models. Asterisks denote significance at $^{*}p < 0.05$, $^{**}p < 0.01$, $^{***}p<0.001$.}
    \label{tab:english-sem-compare}
\end{table}

\section{Results}

\subsection{Identifying flexible lemmas}

Of the 37 languages in UD with at least 100k tokens; in 27 of them, at least 2.5\% of verb and noun lemmas are flexible, which we take to indicate that word class flexibility exists in the language (Table \ref{tab:ud-frequency}). The lemma merging algorithm is crucial for identifying word class flexibility: only 6 of the 37 languages pass the 2.5\% flexibility threshold using the default lemma annotations provided in UD\footnote{Chinese, Danish, English, Hebrew, Indonesian, and Japanese pass the flexibility threshold without the lemma merging algorithm.}. Languages differ in their prevalence of word class flexibility, but every language in our sample has higher verb flexibility than noun flexibility.

\subsection{Asymmetry in semantic metrics}

Table \ref{tab:mbert-semantic-stats} shows the values of the three metrics, computed using mBERT and Wikipedia data for 25 languages\footnote{We exclude 2 of the 27 languages that we identify word class flexibility. Old Russian was excluded because it is not supported by mBERT; Korean is excluded because the lemma annotations deviate from the standard UD format.}. For testing significance, we use the unpaired Student's t-test to compare N-V versus V-N shift, and the paired Student's t-test for the other two metrics\footnote{We do not apply the Bonferroni correction for multiple comparisons, because we  make claims for trends across all languages, and not for any specific languages.}. The key findings are as follows:

\begin{enumerate}
    \item {\bf Asymmetry in semantic shift.} In English, N-V shift is greater than V-N shift, in agreement with \citet{Bauer2005}. However, this pattern does not hold in general: there is no significant difference in either direction in most languages, and two languages exhibit a difference in the opposite direction as English.
    \item {\bf Asymmetry in semantic variation between noun and verb usages.} Of the 17 languages with a statistically significant difference in noun versus verb variation, 14 of them have greater verb variation than noun variation.
    \item {\bf Asymmetry in semantic variation between majority and minority classes.} All of the 20 languages with a statistically significant difference in majority and minority variation have greater majority variation.
\end{enumerate}

\subsection{Model robustness}

Next, we assess the robustness of our metrics with respect to choices of corpus and language model. Robustness is desirable because it gives confidence that our models capture true linguistic tendencies, rather than artifacts of our datasets or the models themselves. We compute the three semantic metrics on the BNC and Wikipedia datasets, using all 4 contextual language models: ELMo, BERT, mBERT, and XLM-R. Table \ref{tab:english-sem-compare} summarizes the results from this experiment.

We find that in almost every case where there is a statistically significant difference, all models agree on the direction of the difference. One exception is that noun variation is greater when computed using Wikipedia data than when using the BNC corpus. Wikipedia has many instances of nouns used in technical senses (e.g., \textit{ring} is a technical term in mathematics and chemistry), whereas similar nonfiction text is less common in the BNC corpus.

\section{Discussion}

\subsection{Frequency asymmetry}

Every language in our sample has verb flexibility greater than noun flexibility. The reasons for this asymmetry are unclear, but may be due to semantic differences between nouns and verbs. We note that every language in our sample has more noun lemmas than verb lemmas, a pattern that was also attested by \citet{more-nouns-than-verbs}, although this does not provide an explanation of the observed phenomenon. We leave further exploration of the flexibility asymmetry to future work.

\subsection{Implications for theories of flexibility}

There is a strong cross-linguistic tendency for the majority word class of a flexible lemma to exhibit more semantic variation than the minority class. In other words, the frequency and semantic variation criteria of determining the base of a conversion pair agree more than at chance. This supports the analysis of word class flexibility as a directional process of conversion, as opposed to underspecification (section \ref{sec:related-work-directionality})\footnote{Since 18 of the 25 languages for which semantic metrics were calculated are Indo-European, it is unclear whether these results generalize to non-Indo-European languages.}. Within a flexible lemma, verbs exhibit more semantic variation than nouns. It is attested across many languages that nouns are more physically salient, while verbs have more complex event and argument structure, and are harder for children to acquire than nouns \citep{gentner-1982, imai-2008}. Thus, verbs are expected to have greater semantic variation than nouns, which our results confirm. More importantly, for our purposes, this metric serves as a control for the previous metric. Flexible lemmas are more likely to be noun-dominant than verb-dominant, so could the majority and minority variation simply be proxies for noun and verb variation, respectively? In fact, we observe greater verb than noun variation, so this cannot be the case.

Finally, as suggested by \citet{Bauer2005}, we find evidence in English that N-V flexibility involves more semantic shift than V-N flexibility, and the pattern is consistent across multiple models and datasets (Table \ref{tab:english-sem-compare}). However, this pattern is idiosyncratic to English and not a cross-linguistic tendency. It is thus instructive to analyze multiple languages in studying word class flexibility, as one can easily be misled by English-based analyses.

\section{Conclusion}

We used contextual language models to examine shared tendencies in word class flexibility across languages. We found that the majority class often exhibits more semantic variation than the minority class, supporting the view that word class flexibility is a directional process. We also found that in English, noun-to-verb flexibility is associated with more semantic shift than verb-to-noun flexibility, but this is not the case for most languages.

Our probing task revealed that the upper layers of BERT contextual embeddings best reflect human judgment of semantic similarity. We obtained similar results in different datasets and language models in English that support the robustness of our method. In this chapter, we demonstrated the utility of deep contextualized models in linguistic typology, especially for characterizing cross-linguistic semantic phenomena that are otherwise difficult to quantify. The next two chapters will present our work using linguistic theory and experimental data to deepen our understanding of language models.

\chapter{Linguistic anomalies in LMs}

{\em The contents of this chapter are based on my previous publication \citep{layerwise-anomaly}.}

\section{Introduction}

The previous chapter used contextual embeddings to measure the semantic distance between flexible words used in different contexts. One limitation of this method is that different linguistic properties -- morphology, syntax, semantics, and pragmatics -- are conflated into a single metric. From contextual embeddings, we cannot easily determine whether the difference between two words is syntactic (e.g., {\em walk} and {\em walks}) or semantic (e.g., {\em walk} and {\em run}).

\begin{figure}[t]
    \centering
    \includegraphics[width=0.6\linewidth]{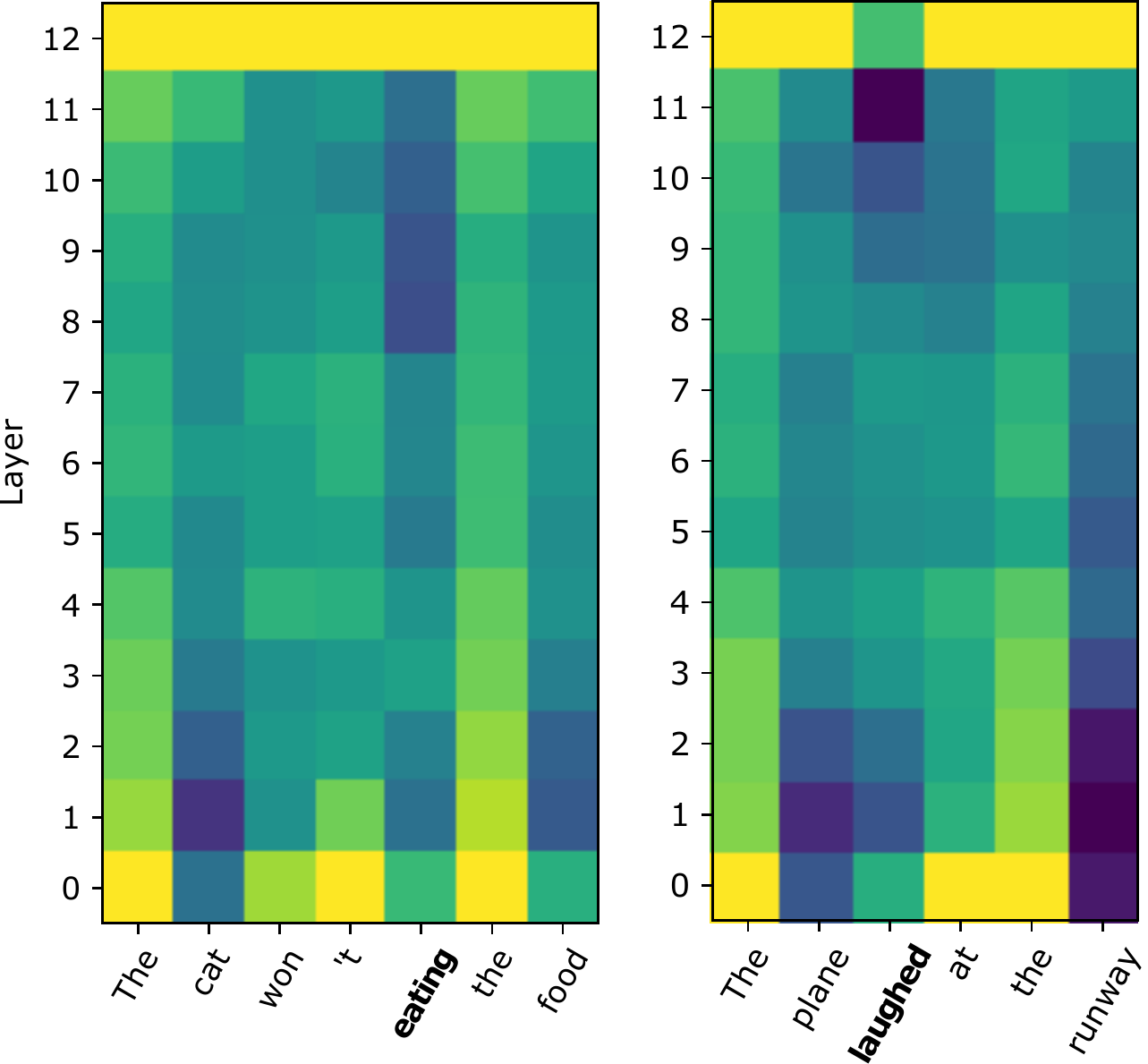}
    \caption{Example sentence with a morphosyntactic anomaly (left) and semantic anomaly (right) (anomalies in bold). Darker colours indicate higher surprisal. We investigate several patterns: first, surprisal at lower layers corresponds to infrequent tokens, but this effect diminishes towards upper layers. Second, morphosyntactic violations begin to trigger high surprisals at an earlier layer than semantic violations.}
    \label{fig:gmm_demo}
\end{figure}

In this chapter, we investigate how Transformer-based language models respond to sentences containing three different types of anomalies: morphosyntactic, semantic, and commonsense. Previous work using behavioural probing found that Transformer LMs have remarkable ability in detecting when a word is anomalous in context, by assigning a higher likelihood to the appropriate word than an inappropriate one \citep{gulordava-colorless, ettinger-psycholinguistic, blimp}. The likelihood score, however, only gives a scalar value of the degree that a word is anomalous in context, and cannot distinguish between different {\em ways} that a word might be anomalous.

It has been proposed that there are different types of linguistic anomalies. \citet{chomsky1957} distinguished semantic anomalies ({\em ``colorless green ideas sleep furiously''}) from ungrammaticality ({\em ``furiously sleep ideas green colorless''}). Psycholinguistic studies initially suggested that different event-related potentials (ERPs) are produced in the brain depending on the type of anomaly; e.g., semantic anomalies produce negative ERPs 400 ms after the stimulus, while syntactic anomalies produce positive ERPs 600 ms after  \citep{psycholinguistics-electrified}. Here, we ask whether Transformer LMs show different surprisals in their intermediate layers depending on the type of anomaly. However, LMs do not compute likelihoods at intermediate layers -- only at the final layer.

We introduce a new tool to probe for surprisal at intermediate layers of BERT \citep{bert}, RoBERTa \citep{roberta}, and XLNet \citep{xlnet}, formulating the problem as density estimation. We train Gaussian models to fit distributions of embeddings at each layer of the LMs. Using BLiMP \citep{blimp} for evaluation, we show that this model is effective at grammaticality judgement, requiring only a small amount of in-domain text for training. Figure \ref{fig:gmm_demo} shows the method using the RoBERTa model on two example sentences.

We apply our model to test sentences drawn from BLiMP and 7 psycholinguistics studies, exhibiting morphosyntactic, semantic, and commonsense anomalies. We find that morphosyntactic anomalies produce out-of-domain embeddings at earlier layers, semantic anomalies at later layers, and commonsense anomalies not at any layer, even though the LM's final accuracy is similar. We show that LMs are internally sensitive to the type of linguistic anomaly, which is not apparent if we only had access to their softmax probability outputs. Our source code and data are available at: \url{https://github.com/SPOClab-ca/layerwise-anomaly}.

\section{Related work}

Our work builds on earlier work on probing LM representations (Section \ref{sec:representational-probes}). Previous work found differences in the linguistic knowledge contained in different layers \citep{bert-rediscovers, kelly-sentence-probe, hewitt-syntax}; we focus on the effects of anomalous inputs on different layers. Behavioural probes (Section \ref{sec:behavioral-probes}) often used anomalous sentences paired with correct sentences to test LMs' sensitivity to linguistic phenomena \citep{linzen2016, gulordava-colorless, blimp, hu-syntax-assessment}; in this work, we extend these tests to probe the sensitivity of internal layer representations to anomalies rather than the model's output.

Most grammaticality studies focused on syntactic phenomena, since they are the easiest to generate using templates, although some studies considered semantic phenomena. Examples of semantic tests include \citet{rabinovich-infelicity}, who tested LMs' sensitivity to semantic infelicities involving indefinite pronouns, and \citet{ettinger-psycholinguistic}, who used data from three psycholinguistic studies to probe BERT's knowledge of commonsense and negation. Another type of linguistic unacceptability is selectional restrictions, defined as a semantic mismatch between a verb and an argument. \citet{sasano} examined the geometry of word classes (e.g., words that can be a direct object of the verb `play') in word vector models; they compared single-class models against discriminative models for learning word class boundaries. \citet{chersoni} tested distributional semantic models on their ability to identify selectional restriction violations using stimuli from two psycholinguistic datasets. Finally, \citet{metheniti-selectional} tested how much BERT relies on selectional restriction information versus other contextual information for making masked word predictions. Our work combines these different types of unacceptability into a single test suite to faciliate comparison.

\section{Model}

We use the transformer language model as a contextual embedding extractor (we write this as BERT for convenience). Let $L$ be the layer index, which ranges from 0 to 12 on all of our models. Using a training corpus $\{w_1, \cdots, w_T\}$, we extract contextual embeddings at layer $L$ for each token:
\begin{equation}
    \vect{x}_1^{(L)}, \cdots, \vect{x}_T^{(L)} = \textrm{BERT}_L(w_1, \cdots, w_T).
\end{equation}
Next, we fit a multivariate Gaussian on the extracted embeddings:
\begin{equation}
    \vect{x}_1^{(L)}, \cdots, \vect{x}_T^{(L)} \sim \mathcal{N} (\widehat{\vect{\mu}}_L, \widehat{\vect{\Sigma}}_L).
\end{equation}

For evaluating the layerwise surprisal of a new sentence $\vect{s} = [t_1, \cdots, t_n]$, we similarly extract contextual embeddings using the language model:
\begin{equation}
    \vect{y}_1, \cdots, \vect{y}_n = \textrm{BERT}_L(t_1, \cdots, t_n).
\end{equation}
The surprisal of each token is the negative log likelihood of the contextual vector according to the multivariate Gaussian:
\begin{equation} \label{eq:token-surprisal}
    G_i = -\log p(\vect{y}_i \mid \widehat{\vect{\mu}}_L, \widehat{\vect{\Sigma}}_L) \quad \textrm{for} \ i=1\ldots n.
\end{equation}
Finally, we define the surprisal of sentence $\vect{s}$ as the sum of surprisals of all of its tokens, which is also the joint log likelihood of all of the embeddings:
\begin{equation} \label{eq:sum-agg}
\begin{split}
    \textrm{surprisal}_L(t_1, \cdots, t_n) &= \sum_{i=1}^n G_i \\
    &= -\log p(\vect{y}_1, \cdots, \vect{y}_n \mid \widehat{\vect{\mu}}_L, \widehat{\vect{\Sigma}}_L).
\end{split}
\end{equation}

\subsection{Connection to Mahalanobis distance}

The theoretical motivation for using the sum of log likelihoods is that when we fit a Gaussian model with full covariance matrix, low likelihood corresponds exactly to high Mahalanobis distance from the in-distribution points. The score given by the Gaussian model is:
\begin{equation}
\begin{split}
    G &= -\log p(\vect{y} \mid \widehat{\vect{\mu}}_L, \widehat{\vect{\Sigma}}_L) \\
    &= -\log \left( \frac{1}{(2 \pi)^{D/2} |\widehat{\vect{\Sigma}}_L|^{1/2}} \exp(-\frac{1}{2} d^2) \right),
\end{split}
\end{equation}
where $D$ is the dimension of the vector space, and $d$ is the Mahalanobis distance:
\begin{equation}
    d = \sqrt{(y - \widehat{\vect{\mu}}_L)^T \widehat{\vect{\Sigma}}_L^{-1} (y - \widehat{\vect{\mu}}_L)}.
\end{equation}
Rearranging, we get:
\begin{equation}
    d^2 = 2G -D \log(2 \pi) - \log |\widehat{\vect{\Sigma}}_L|,
\end{equation}
thus the negative log likelihood is the squared Mahalanobis distance plus a constant.

Various methods based on Mahalanobis distance have been used for anomaly detection in neural networks; for example, \citet{lee-anomaly-mahalanobis} proposed a similar method for out-of-domain detection in neural classification models, and \citet{cao-medical-anomaly} found the Mahalanobis distance method to be competitive with more sophisticated methods on medical out-of-domain detection. In Transformer models, \citet{transformer-mahalanobis} used Mahalanobis distance for out-of-domain detection, outperforming methods based on softmax probability and likelihood ratios.

{\bf Gaussian assumptions.} Our model assumes that the embeddings at every layer follow a multivariate Gaussian distribution. Since the Gaussian distribution is the maximum entropy distribution given a mean and covariance matrix, it makes the fewest assumptions and is therefore a reasonable default. \citet{hennigen2020} found that embeddings sometimes do not follow a Gaussian distribution, but it is unclear what alternative distribution would be a better fit, so we will assume a Gaussian distribution in this work.

\subsection{Training and evaluation}

\begin{figure*}
    \centering
    \begin{subfigure}[]{0.6\linewidth}
        \includegraphics[width=\linewidth]{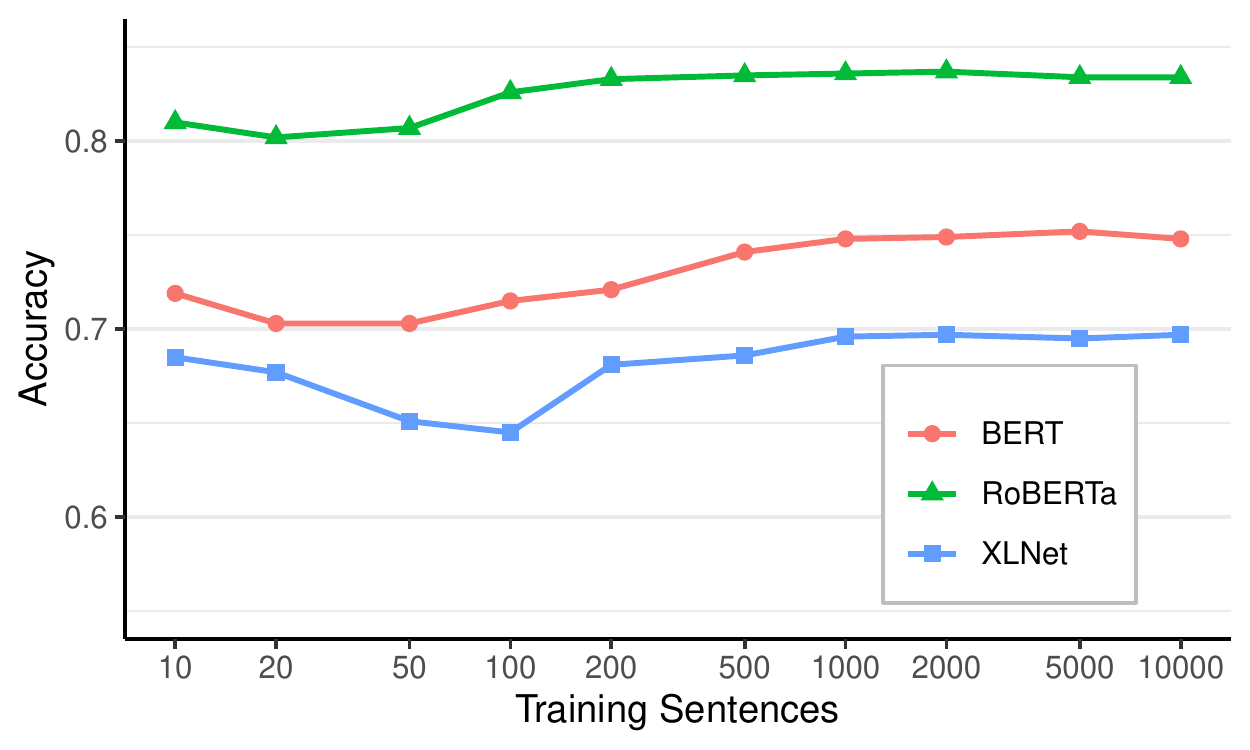}
        \caption{}
        \label{fig:num-sent}
    \end{subfigure}
    \begin{subfigure}[]{0.6\linewidth}
        \includegraphics[width=\linewidth]{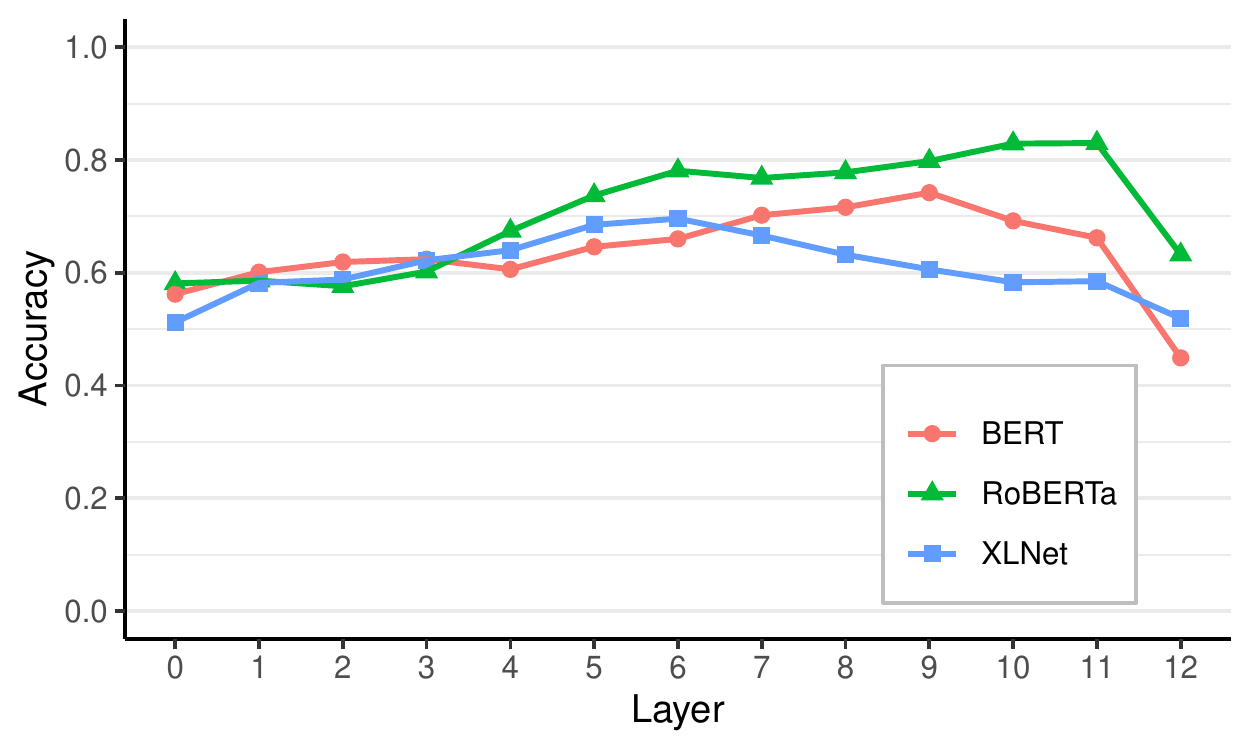}
        \caption{}
        \label{fig:layer-accuracy}
    \end{subfigure}
    \caption{BLiMP accuracy different amounts of training data and across layers, for three LMs. About 1000 sentences are needed before a plateau is reached (mean tokens per sentence = 15.1).}
\end{figure*}

For all of our experiments, we use the `base' versions of pretrained language models BERT \citep{bert}, RoBERTa \citep{roberta}, and XLNet \citep{xlnet}, provided by HuggingFace \citep{huggingface}. Each of these models have 12 contextual layers plus a 0$^\textrm{th}$ static layer, and each layer is 768-dimensional.

We train the Gaussian model on randomly selected sentences from the British National Corpus \citep{bnc}, representative of acceptable English text from various genres. We evaluate on BLiMP \citep{blimp}, a dataset of 67k minimal sentence pairs that test acceptability judgements across a variety of syntactic and semantic phenomena. In our case, a sentence pair is considered correct if the sentence-level surprisal of the unacceptable sentence is higher than that of the acceptable sentence.

{\bf How much training data is needed?} We experiment with training data sizes ranging from 10 to 10,000 sentences (Figure \ref{fig:num-sent}). Compared to the massive amount of data needed for pretraining the LMs, we find that a modest corpus suffices for training the Gaussian anomaly model, and a plateau is reached after 1000 sentences for all three models. Therefore, we use 1000 training sentences (unless otherwise noted) for all subsequent experiments in this chapter.

{\bf Which layers are sensitive to anomaly?} We vary $L$ from 0 to 12 in all three models (Figure \ref{fig:layer-accuracy}). The layer with the highest accuracy differs between models: layer 9 has the highest accuracy for BERT, 11 for RoBERTa, and 6 for XLNet. All models experience a sharp drop in the last layer, likely because the last layer is specialized for the MLM pretraining objective.
\label{sec:best-layer}

{\bf Comparisons to other models.} Our best-performing model is RoBERTa, with an accuracy  of 0.830. This is slightly higher the best model reported in BLiMP (GPT-2, with accuracy 0.801). We do not claim to beat the state-of-the-art on BLiMP: \citet{salazar-mlm-scoring} obtains a higher accuracy of 0.865 using RoBERTa-large. Even though the main goal of this work is not to maximize accuracy on BLiMP, our Gaussian anomaly model is competitive with other transformer-based models on this task.

\subsection{Further ablation studies on Gaussian model}

\begin{table}
\centering
\begin{tabular}{lr}
\hline
\textbf{Covariance} & \multicolumn{1}{l}{\textbf{Accuracy}} \\ \hline
Full              & 0.830                                 \\
Diagonal              & 0.755                                 \\
Spherical         & 0.752                                 \\ \hline
\end{tabular}
\caption{Varying the type of covariance matrix in the Gaussian model.}
\label{tab:cov-acc}
\vspace{3em}

\begin{tabular}{rr}
\hline
\multicolumn{1}{l}{\textbf{Components}} & \multicolumn{1}{l}{\textbf{Accuracy}} \\ \hline
1                                       & 0.830                                 \\
2                                       & 0.841                                 \\
4                                       & 0.836                                 \\
8                                       & 0.849                                 \\
16                                      & 0.827                                 \\ \hline
\end{tabular}
\caption{Using Gaussian mixture models (GMMs) with multiple components.}
\label{tab:gmm-acc}
\vspace{3em}

\begin{tabular}{lr}
\hline
\textbf{Genre} & \multicolumn{1}{l}{\textbf{Accuracy}} \\ \hline
Academic            & 0.797                                 \\
Fiction            & 0.840                                 \\
News           & 0.828                                 \\
Spoken            & 0.795                                 \\
All            & 0.830                                 \\ \hline
\end{tabular}
\caption{Effect of the genre of training data.}
\label{tab:genre-acc}
\vspace{3em}

\begin{tabular}{lr}
\hline
\textbf{Kernel} & \multicolumn{1}{l}{\textbf{Score}} \\ \hline
RBF             & 0.738                              \\
Linear          & 0.726                              \\
Polynomial            & 0.725                              \\ \hline
\end{tabular}
\caption{Using 1-SVM instead of GMM, with various kernels.}
\label{tab:1svm-acc}
\vspace{3em}

\begin{tabular}{lr}
\hline
\textbf{Aggregation} & \multicolumn{1}{l}{\textbf{Accuracy}} \\ \hline
Sum                  & 0.830                              \\
Max                  & 0.773                              \\ \hline
\end{tabular}
\caption{Two sentence-level aggregation strategies}
\label{tab:max-acc}
\end{table}

We explore some variations to our methodology of training the Gaussian model. All of these variations are evaluated on the full BLiMP dataset. In each experiment,  (unless otherwise noted)  the language model is RoBERTa-base, using the second-to-last layer, and the Gaussian model has a full covariance matrix trained with 1000 sentences from the BNC corpus.

{\bf Covariance matrix}. We vary the type of covariance matrix (Table \ref{tab:cov-acc}). Diagonal and spherical covariance matrices perform worse than with the full covariance matrix; this may be expected, as the full matrix has the most trainable parameters.

{\bf Gaussian mixture models}. We try GMMs with up to 16 mixture components (Table \ref{tab:gmm-acc}). We observe a small increase in accuracy compared to a single Gaussian, but the difference is too small to justify the increased training time.

{\bf Genre of training text}. We sample from genres of BNC (each time with 1000 sentences) to train the Gaussian model (Table \ref{tab:genre-acc}). The model performed worse when trained with the academic and spoken genres, and about the same with the fiction and news genres, perhaps because their vocabularies and grammars are more similar to those in the BLiMP sentences.

{\bf One-class SVM}. We try replacing the Gaussian model with a one-class SVM \citep{one-svm}, another popular model for anomaly detection. We use the default settings from scikit-learn with three kernels (Table \ref{tab:1svm-acc}), but it performs worse than the Gaussian model on all settings.

{\bf Sentence aggregation}. Instead of Equation \ref{eq:sum-agg}, we try defining sentence-level surprisal as the maximum surprisal among all tokens (Table \ref{tab:max-acc}):
\begin{equation}
    \textrm{surprisal}(s_1, \cdots, s_n) = \textrm{max}_{i=1}^n G_i; \\
\end{equation}
however, this performs worse than using the sum of token surprisals.

\subsection{Lower layers are sensitive to frequency}

We notice that surprisal scores in the lower layers are sensitive to token frequency: higher frequency tokens produce embeddings close to the center of the Gaussian distribution, while lower frequency tokens are at the periphery. The effect gradually diminishes towards the upper layers.

\begin{figure}[t]
    \centering
    \includegraphics[width=0.6\linewidth]{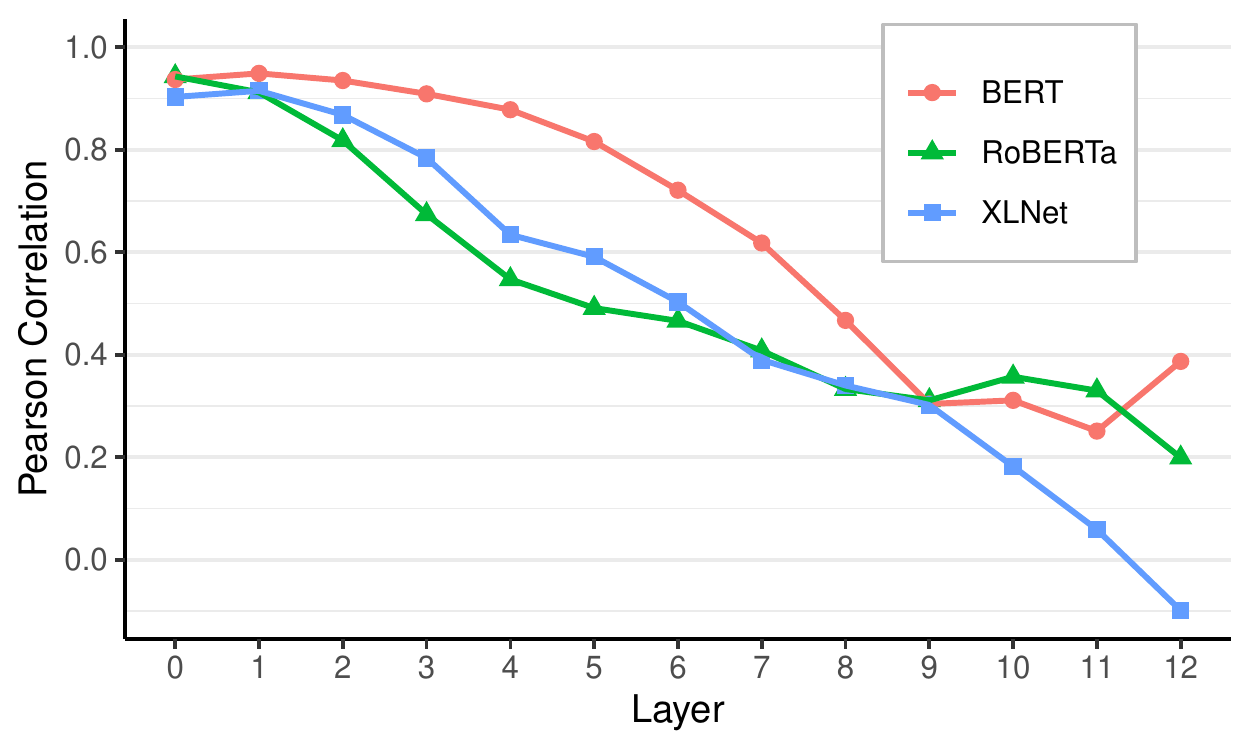}
    \caption{Pearson correlation between token-level surprisal scores (Equation \ref{eq:token-surprisal}) and log frequency. The correlation is highest in the lower layers, and decreases in the upper layers.}
    \label{fig:freq-correlation}
\end{figure}

To quantify the sensitivity to frequency, we compute token-level surprisal scores for 5000 sentences from BNC that were not used in training. We then compute the Pearson correlation between the surprisal score and log frequency for each token (Figure \ref{fig:freq-correlation}). In all three models, there is a high correlation between the surprisal score and log frequency at the lower layers, which diminishes at the upper layers. A small positive correlation persists until the last layer, except for XLNet, in which the correlation eventually disappears.

\begin{figure}[t]
    \centering
    \includegraphics[width=0.7\linewidth]{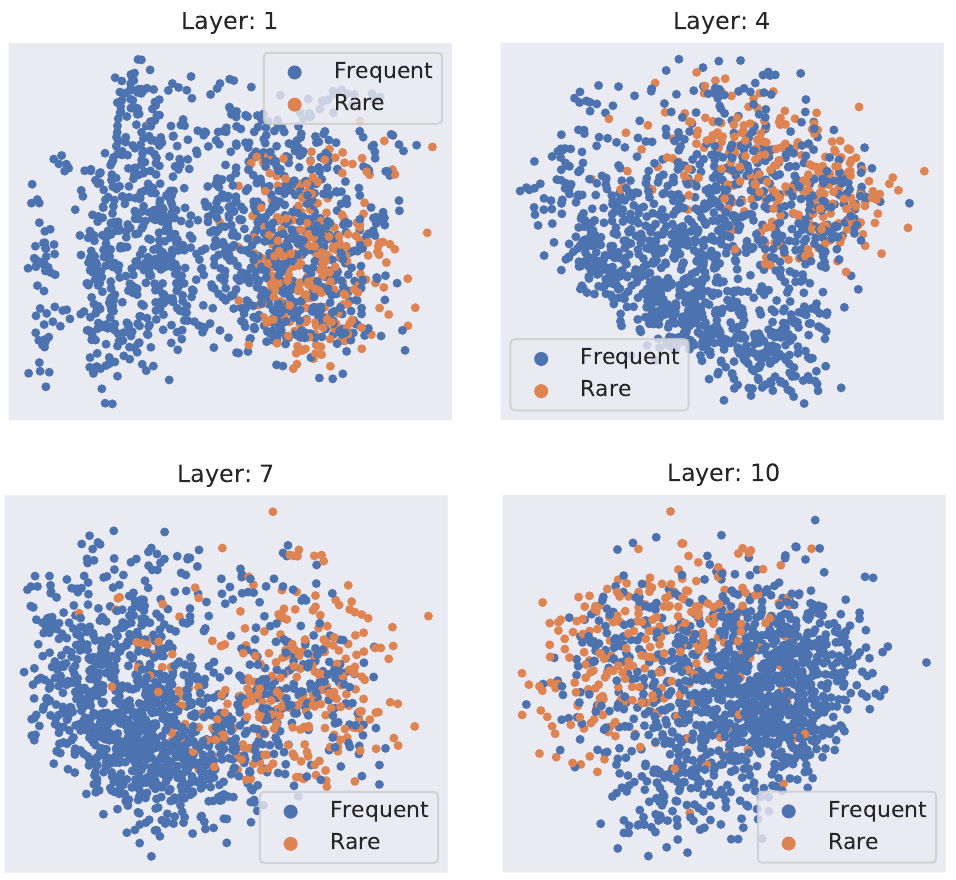}
    \caption{PCA plot of randomly sampled RoBERTa embeddings at layers 1, 4, 7, and 10. Points are colored by token frequency: ``Rare'' means the 20\% least frequent tokens, and ``Frequent'' is the other 80\%.}
    \label{fig:freq-4layer}
\end{figure}

There does not appear to be any reports of this phenomenon in previous work. For static word vectors, \citet{gong-frage} found that embeddings for low-frequency words lie in a different region of the embedding space than high-frequency words. To visualize this phenomenon, we feed a random selection of BNC sentences into RoBERTa and use PCA to visualize the distribution of rare and frequent tokens at different layers (Figure \ref{fig:freq-4layer}). In all cases, we find that infrequent tokens occupy a different region of the embedding space from frequent tokens, similar to what \citet{gong-frage} observed for static word vectors. The Gaussian model fits the high-frequency region and assigns lower likelihoods to the low-frequency region, explaining the positive correlation at all layers, although it is still unclear why the correlation diminishes at upper layers.

\section{Levels of linguistic anomalies}

We turn to the question of whether LMs exhibit different behaviour when given inputs with different types of linguistic anomalies. The task of partitioning linguistic anomalies into several distinct classes can be challenging. Syntax and semantics have a high degree of overlap -- there is no widely accepted criterion for distinguishing between ungrammaticality and semantic anomaly (e.g., \citet{abrusan-grammaticality} gives a survey of current proposals), and \citet{poulsen-grammaticality} challenges this dichotomy entirely. Similarly, \citet{warren} noted that semantic anomalies depend somewhat on world knowledge.

Within a class, the anomalies are also heterogeneous (e.g., ungrammaticality may be due to violations of agreement, {\em wh}-movement, negative polarity item licensing, etc), which might each affect the LMs differently. Thus, we define three classes of anomalies that do not attempt to cover all possible linguistic phenomena, but captures different levels of language processing while retaining internal uniformity:

\begin{enumerate}
    \item {\bf Morphosyntactic anomaly}: an error in the inflected form of a word, for example, subject-verb agreement ({\em \text{*}the boy eat the sandwich}), or incorrect verb tense or aspect inflection ({\em \text{*}the boy eaten the sandwich}). In each case, the sentence can be corrected by changing the inflectional form of one word.
    \item {\bf Semantic anomaly}: a violation of a selectional restriction, such as animacy ({\em \#the house eats the sandwich}). In these cases, the sentence can be corrected by replacing one of the verb's arguments with another one in the same word class that satisfies the verb's selectional restrictions.
    \item {\bf Commonsense anomaly}: sentence describes an situation that is atypical or implausible in the real world but is otherwise acceptable ({\em \#the customer served the waitress}).
\end{enumerate}

\begin{table*}[]
\small
\begin{tabular}{l>{\raggedright}m{0.2\linewidth}ll}
\hline
  \multicolumn{1}{l}{\textbf{Type}} &
  \textbf{Task} &
  \textbf{Correct Example} &
  \textbf{Incorrect Example} \\ \hline
\multirow[b]{3}{*}{\vspace{-2mm}Morphosyntax} &
  BLiMP (Subject-Verb) &
  These casseroles {\bf disgust} Kayla. &
  These casseroles {\bf disgusts} Kayla. \\ [3mm]
&
  BLiMP (Det-Noun) &
  Craig explored that grocery {\bf store}. &
  Craig explored that grocery {\bf stores}. \\ [3mm]
&
  \citet{osterhout-nicol} &
  \begin{tabular}[c]{@{}l@{}}The cats won't {\bf eat} the food that\\ Mary gives them.\end{tabular} &
  \begin{tabular}[c]{@{}l@{}}The cats won't {\bf eating} the food that\\ Mary gives them.\end{tabular} \\ \hline
\multirow[b]{5}{*}{\vspace{-3mm}Semantic} &
  BLiMP (Animacy) &
  \begin{tabular}[c]{@{}l@{}}Amanda was respected by some\\ {\bf waitresses}.\end{tabular} &
  \begin{tabular}[c]{@{}l@{}}Amanda was respected by some\\ {\bf picture}.\end{tabular} \\ [3mm]
&
  \citet{pylkkanen} &
  \begin{tabular}[c]{@{}l@{}}The pilot {\bf flew} the airplane after\\ the intense class.\end{tabular} &
  \begin{tabular}[c]{@{}l@{}}The pilot {\bf amazed} the airplane after\\ the intense class.\end{tabular} \\ [3mm]
&
  \citet{warren} &
  \begin{tabular}[c]{@{}l@{}}Corey's hamster {\bf explored} a nearby\\ backpack and filled it with sawdust.\end{tabular} &
  \begin{tabular}[c]{@{}l@{}}Corey's hamster {\bf entertained} a nearby\\ backpack and filled it with sawdust.\end{tabular} \\ [3mm]
&
  \citet{osterhout-nicol} &
  \begin{tabular}[c]{@{}l@{}}The cats won't {\bf eat} the food that\\ Mary gives them.\end{tabular} &
  \begin{tabular}[c]{@{}l@{}}The cats won't {\bf bake} the food that\\ Mary gives them.\end{tabular} \\ [3mm]
&
  \citet{osterhout-mobley} &
  \begin{tabular}[c]{@{}l@{}}The plane sailed through the air and\\ {\bf landed} on the runway.\end{tabular} &
  \begin{tabular}[c]{@{}l@{}}The plane sailed through the air and\\ {\bf laughed} on the runway.\end{tabular} \\ \hline
\multirow[b]{4}{*}{\vspace{-4mm}Commonsense} &
  \citet{warren} &
  \begin{tabular}[c]{@{}l@{}}Corey's hamster {\bf explored} a nearby\\ backpack and filled it with sawdust.\end{tabular} &
  \begin{tabular}[c]{@{}l@{}}Corey's hamster {\bf lifted} a nearby\\ backpack and filled it with sawdust.\end{tabular} \\ [3mm]
&
  \citet{federmeier-cprag} &
  \begin{tabular}[c]{@{}l@{}}``Checkmate,'' Rosalie announced\\ with glee. She was getting to be\\ really good at {\bf chess}.\end{tabular} &
  \begin{tabular}[c]{@{}l@{}}``Checkmate,'' Rosalie announced\\ with glee. She was getting to be\\ really good at {\bf monopoly}.\end{tabular} \\ [3mm]
&
  \citet{chow-role88} &
  \begin{tabular}[c]{@{}l@{}}The restaurant owner forgot which\\ {\bf customer} the {\bf waitress} had served.\end{tabular} &
  \begin{tabular}[c]{@{}l@{}}The restaurant owner forgot which\\ {\bf waitress} the {\bf customer} had served.\end{tabular} \\ [3mm]
&
  \citet{urbach} &
  \begin{tabular}[c]{@{}l@{}}Prosecutors accuse {\bf defendants} of\\ committing a crime.\end{tabular} &
  \begin{tabular}[c]{@{}l@{}}Prosecutors accuse {\bf sheriffs} of\\ committing a crime.\end{tabular} \\ \hline
\end{tabular}
\caption{Example sentence pair for each of the 12 tasks. The 3 BLiMP tasks are generated from templates; the others are stimuli materials taken from psycholinguistic studies.}
\label{tab:task-example}
\end{table*}

\subsection{Summary of anomaly datasets}

We use two sources of data for experiments on linguistic anomalies: synthetic sentences generated from templates, and materials from psycholinguistic studies. Both have advantages and disadvantages -- synthetic data can be easily generated in large quantities, but the resulting sentences may be odd in unintended ways. Psycholinguistic stimuli are designed to control for confounding factors (e.g., word frequency) and human-validated for acceptability, but are smaller (typically fewer than 100 sentence pairs).

We curate a set of 12 tasks from BLiMP and 7 psycholinguistic studies\footnote{Several of these stimuli have been used in natural language processing research. \citet{chersoni} used the data from \citet{pylkkanen} and \citet{warren} to probe word vectors for knowledge of selectional restrictions. \citet{ettinger-psycholinguistic} used data from \citet{federmeier-cprag} and \citet{chow-role88}, which were referred to as CPRAG-102 and ROLE-88 respectively.}. Each sentence pair consists of a control and an anomalous sentence, so that all sentences within a task differ in a consistent manner. Table \ref{tab:task-example} shows an example sentence pair from each task. We summarize each dataset:

\begin{enumerate}
    \item BLiMP \citep{blimp}: we use subject-verb  and determiner-noun agreement tests as morphosyntactic anomaly tasks. For simplicity, we only use the basic regular sentences, and exclude sentences involving irregular words or distractor items. We also use the two argument structure tests involving animacy as a semantic anomaly task. All three BLiMP tasks therefore have 2000 sentence pairs.
    \item \citet{osterhout-nicol}: contains 90 sentence triplets containing a control, syntactic, and semantic anomaly. Syntactic anomalies involve a modal verb followed by a verb in {\em -ing} form; semantic anomalies have a selectional restriction violation between the subject and verb. There are also double anomalies (simultaneously syntactic and semantic) which we do not use.
    \item \citet{pylkkanen}: contains 70 sentence pairs where the verb is replaced in the anomalous sentence with one that requires an animate object, thus violating the selectional restriction. In half the sentences, the verb is contained in an embedded clause.
    \item \citet{warren}: contains 30 sentence triplets with a possible condition, a selectional restriction violation between the subject and verb, and an impossible condition where the subject cannot carry out the action, i.e., a commonsense anomaly.
    \item \citet{osterhout-mobley}: we use data from experiment 2, containing 90 sentence pairs where the verb in the anomalous sentence is semantically inappropriate. The experiment also tested gender agreement errors, but we do not include these stimuli.
    \item \citet{federmeier-cprag}: contains 34 sentence pairs, where the final noun in each anomalous sentence is an inappropriate completion, but in the same semantic category as the expected completion.
    \item \citet{chow-role88}: contains 44 sentence pairs, where two of the nouns in the anomalous sentence are swapped to reverse their roles. This is the only task in which the sentence pair differs by more than one token.
    \item \citet{urbach}: contains 120 sentence pairs, where the anomalous sentence replaces a patient of the verb with an atypical one.
\end{enumerate}

\subsection{Quantifying layerwise surprisal}

Let $\mathcal{D} = \{(\vect{s}_1, \vect{s}_1'), \cdots, (\vect{s}_n, \vect{s}_n')\}$ be a dataset of sentence pairs, where $\vect{s}_i$ is a control sentence and $\vect{s}_i'$ is an anomalous sentence. For each layer $L$, we define the {\em surprisal gap} as the mean difference of surprisal scores between the control and anomalous sentences, scaled by the standard deviation:
\begin{equation}
    \textrm{surprisal\ gap}_L(\mathcal{D}) = \frac{\mathbb{E}\{\textrm{surprisal}_L(\vect{s}'_i) - \textrm{surprisal}_L(\vect{s}_i)\}_{i=1}^n}
    {\sigma\{\textrm{surprisal}_L(\vect{s}'_i) - \textrm{surprisal}_L(\vect{s}_i)\}_{i=1}^n}
\end{equation}

\begin{figure}
    \centering
    \includegraphics[height=0.9\textheight]{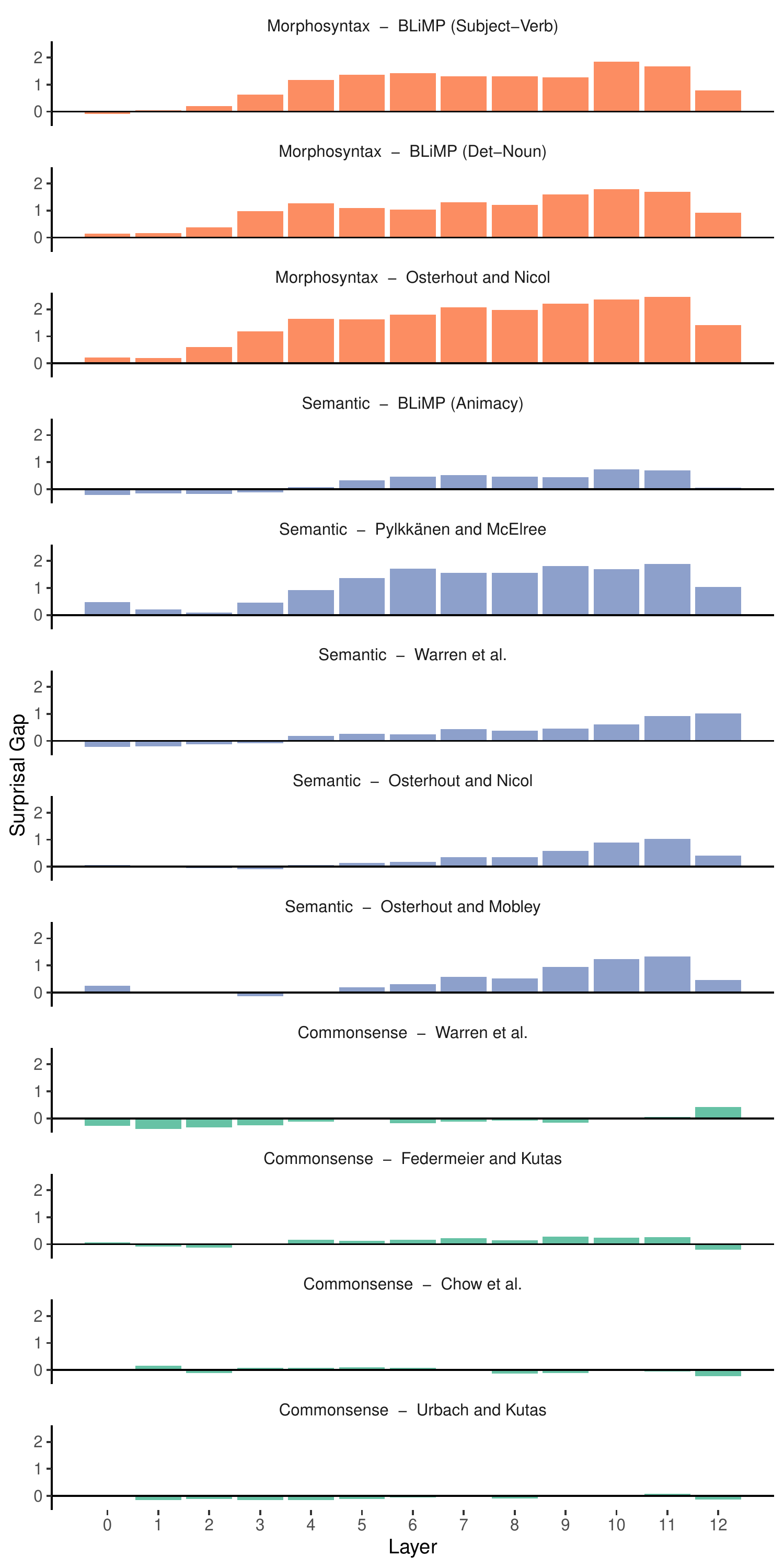}
    \caption{Layerwise surprisal gaps for all tasks using the RoBERTa model. Generally, a positive surprisal gap appears in earlier layers for morphosyntactic tasks than for semantic tasks; no surprisal gap appears at any layer for commonsense tasks.}
    \label{fig:roberta-manual}
\end{figure}

\begin{figure}
    \centering
    \includegraphics[height=0.9\textheight]{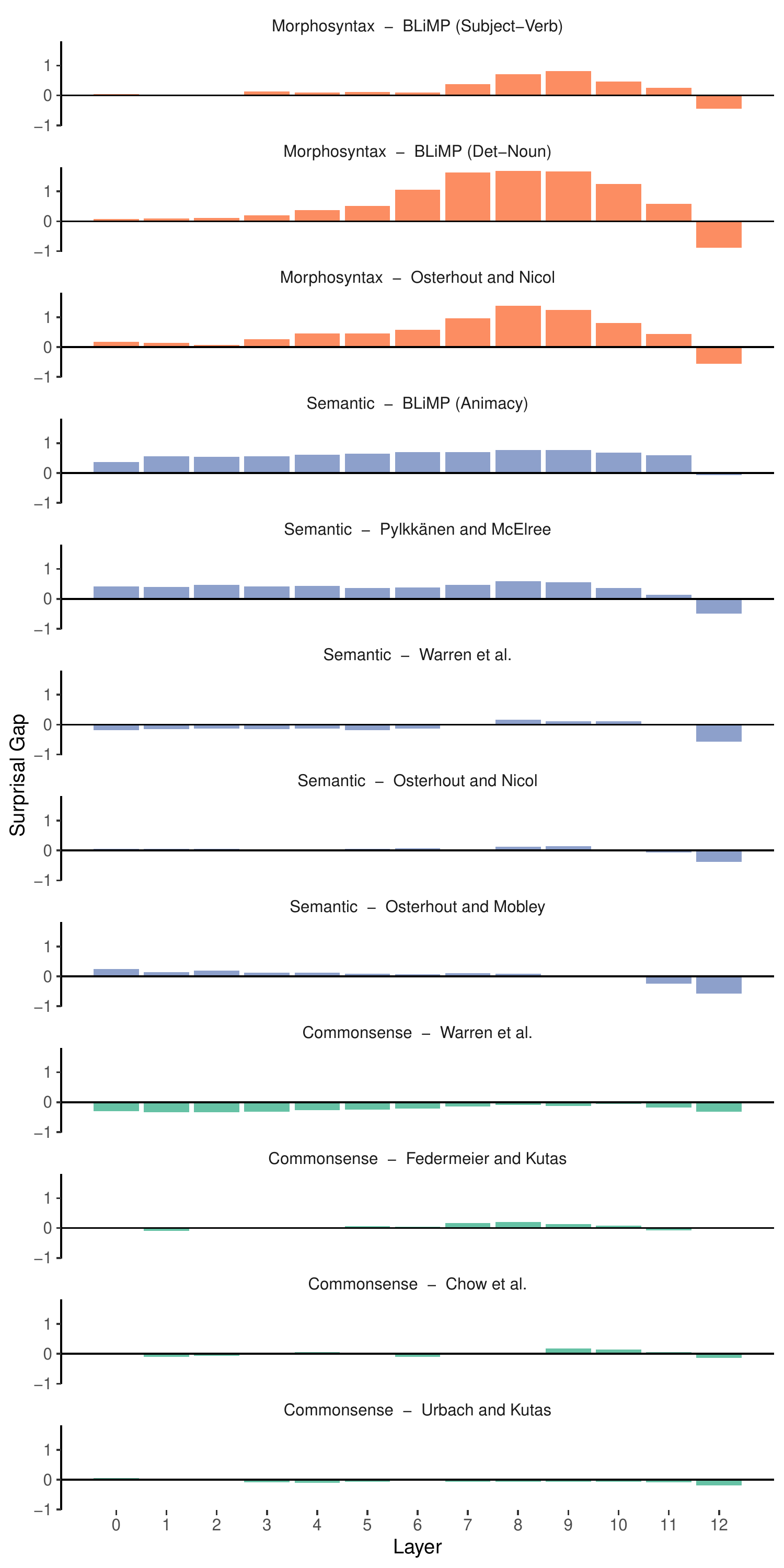}
    \caption{Layerwise surprisal gaps for all tasks using the BERT model.}
    \label{fig:bert-manual}
\end{figure}

\begin{figure}
    \centering
    \includegraphics[height=0.9\textheight]{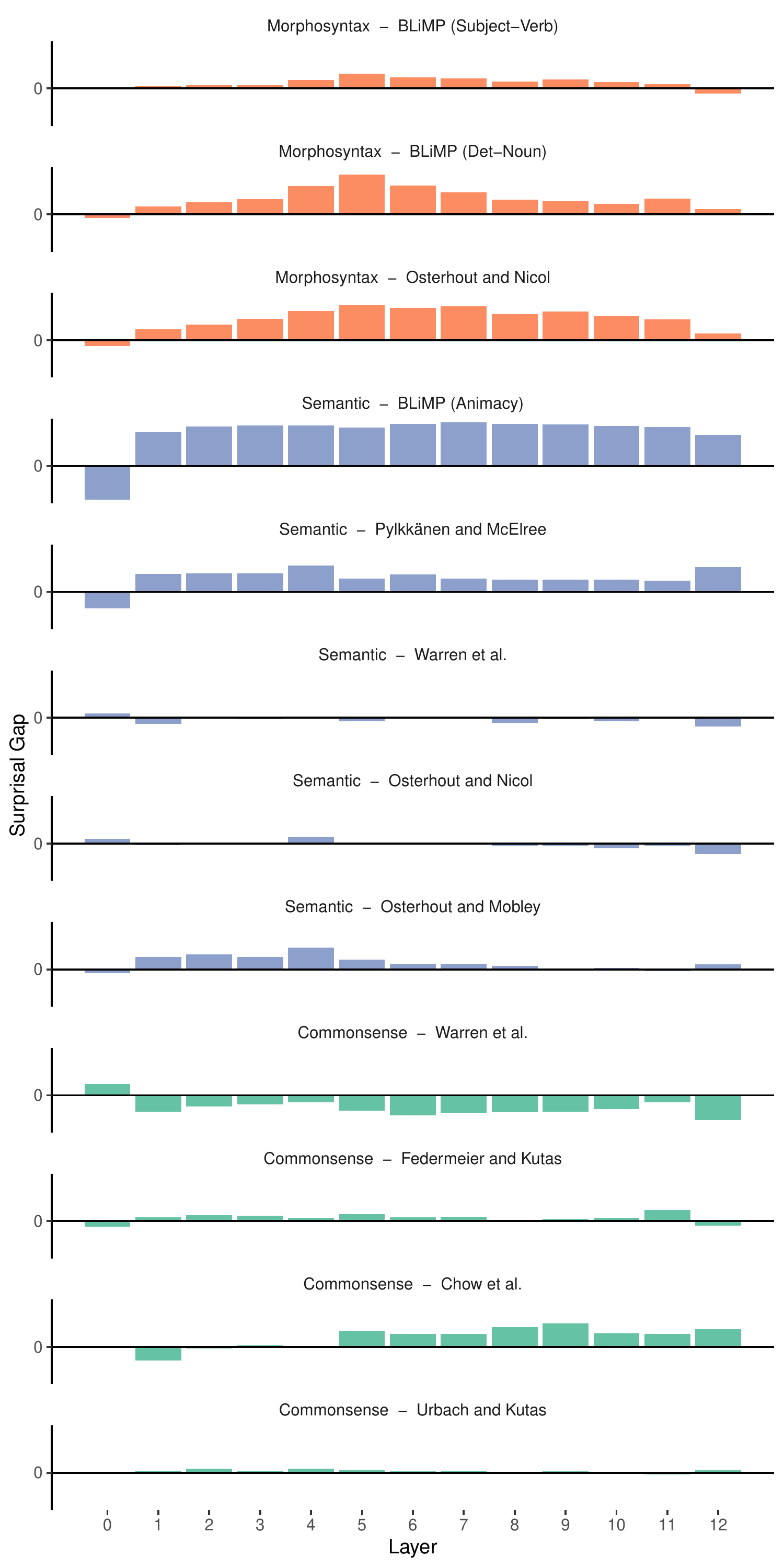}
    \caption{Layerwise surprisal gaps for all tasks using the XLNet model.}
    \label{fig:xlnet-manual}
\end{figure}

The surprisal gap is a scale-invariant measure of sensitivity to anomaly, similar to a signal-to-noise ratio. While surprisal scores are unitless, the surprisal gap may be viewed as the number of standard deviations that anomalous sentences trigger surprisal above control sentences. This is advantageous over accuracy scores, which treats the sentence pair as correct when the anomalous sentence has higher surprisal by any margin; this hard cutoff masks differences in the magnitude of surprisal. The metric also allows for fair comparison of surprisal scores across datasets of vastly different sizes. We plot the surprisal gap for all 12 tasks, using RoBERTa (Figure \ref{fig:roberta-manual}), BERT (Figure \ref{fig:bert-manual}), and XLNet (Figure \ref{fig:xlnet-manual}).

Next, we compare the performance of the Gaussian  model with the masked language model (MLM). We score each instance as correct if the masked probability of the correct word is higher than the anomalous word. One limitation of the MLM approach is that it requires the sentence pair to be identical in all places except for one token, since the LMs do not support modeling joint probabilities over multiple tokens. To ensure fair comparison between GM and MLM, we exclude instances where the differing token is out-of-vocabulary in any of the LMs (this excludes approximately 30\% of instances). For the Gaussian  model, we compute accuracy using the best-performing layer for each model (Section \ref{sec:best-layer}). The results are listed in Table \ref{tab:mlm-results}.

\begin{table*}
\centering \small
\begin{tabular}{llrrrrrrr}
\hline
\multirow{2}{*}{\textbf{Type}} &
  \multirow{2}{*}{\textbf{Task}} &
  \multicolumn{1}{c}{\multirow{2}{*}{\textbf{Size}}} &
  \multicolumn{2}{c}{\textbf{BERT}} &
  \multicolumn{2}{c}{\textbf{RoBERTa}} &
  \multicolumn{2}{c}{\textbf{XLNet}} \\
                              &                          & \multicolumn{1}{c}{} & GM          & MLM         & GM          & MLM   & GM          & MLM         \\ \hline
\multirow{3}{*}{Morphosyntax} & BLiMP (Subject-Verb)     & 2000                 & 0.953       & 0.955       & 0.971       & 0.957 & 0.827       & 0.584       \\
                              & BLiMP (Det-Noun)         & 2000                 & 0.970       & 0.999       & 0.983       & 0.999 & 0.894       & 0.591       \\
                              & \citet{osterhout-nicol}  & 90                   & 1.000       & 1.000       & 1.000       & 1.000 & 0.901       & 0.718       \\ \hline
\multirow{5}{*}{Semantic}     & BLiMP (Animacy)          & 2000                 & 0.644       & 0.787       & 0.767       & 0.754 & 0.675       & 0.657       \\
                              & \citet{pylkkanen}        & 70                   & 0.727       & 0.955       & 0.932       & 0.955 & $^{*}$0.636 & 0.727       \\
                              & \citet{warren}           & 30                   & $^{*}$0.556 & 1.000       & 0.944       & 1.000 & $^{*}$0.667 & $^{*}$0.556 \\
                              & \citet{osterhout-nicol}  & 90                   & 0.681       & 0.957       & 0.841       & 1.000 & $^{*}$0.507 & 0.783       \\
                              & \citet{osterhout-mobley} & 90                   & $^{*}$0.528 & 1.000       & 0.906       & 0.981 & $^{*}$0.302 & 0.774       \\ \hline
\multirow{4}{*}{Commonsense} &
  \citet{warren} &
  30 &
  $^{*}$0.600 &
  $^{*}$0.550 &
  0.750 &
  $^{*}$0.450 &
  $^{*}$0.300 &
  $^{*}$0.600 \\
                              & \citet{federmeier-cprag} & 34                   & $^{*}$0.458 & $^{*}$0.708 & $^{*}$0.583 & 0.875 & $^{*}$0.625 & $^{*}$0.667 \\
                              & \citet{chow-role88}      & 44                   & $^{*}$0.591 & n/a         & $^{*}$0.432 & n/a   & $^{*}$0.568 & n/a         \\
                              & \citet{urbach}           & 120                  & $^{*}$0.470 & 0.924       & $^{*}$0.485 & 0.939 & $^{*}$0.500 & 0.712       \\ \hline
\end{tabular}
\caption{Comparing accuracy scores between Gaussian anomaly model (GM) and masked language model (MLM) for all models and tasks. Asterisks indicate that the accuracy is not better than random (0.5), using a binomial test with threshold of $p < 0.05$ for significance. The MLM results for \citet{chow-role88} are excluded because the control and anomalous sentences differ by more than one token. The best layers for each model (Section \ref{sec:best-layer}) are used for GM, and the last layer is used for MLM. Generally, MLM outperforms GM, and the difference is greater for semantic and commonsense tasks.}
\label{tab:mlm-results}
\end{table*}

\section{Discussion}

\subsection{Anomaly type and surprisal}

We first discuss the results from RoBERTa (Figure \ref{fig:roberta-manual}), where morphosyntactic anomalies generally appear earlier than semantic anomalies. The surprisal gap plot exhibits different patterns depending on the type of linguistic anomaly: morphosyntactic anomalies produce high surprisal relatively early (layers 3-4), while semantic anomalies produce low surprisals until later (layers 9 and above). Commonsense anomalies do not result in surprisals at {\em any} layer: the surprisal gap is near zero for all of the commonsense tasks. The observed difference between morphosyntactic and semantic anomalies is consistent with previous work \citep{bert-rediscovers}, which found that syntactic information appeared earlier in BERT than semantic information.

One should be careful and avoid drawing conclusions from only a few experiments. A similar situation occurred in psycholinguistics research \citep{psycholinguistics-electrified}: early results suggested that the N400 was triggered by semantic anomalies, while syntactic anomalies triggered the P600 -- a different type of ERP. However, subsequent experiments found exceptions to this rule, and now it is believed that the N400 cannot be categorized by any standard dichotomy, like syntax versus semantics \citep{kutas-federmeier}. In our case, \citet{pylkkanen} is an exception: the task is a semantic anomaly, but produces surprisals in early layers, similar to the morphosyntactic tasks. Hence it is possible that the dichotomy is something other than syntax versus semantics; we leave to future work to determine more precisely what conditions trigger high surprisals in lower versus upper layers of LMs.

\subsection{Comparing anomaly model with MLM}

The masked language model (MLM) usually outperforms the Gaussian anomaly model (GM), but the difference is uneven. MLM performs much better than GM on commonsense tasks, slightly better on semantic tasks, and about the same or slightly worse on morphosyntactic tasks. It is not obvious why MLM should perform better than GM, but we note two subtle differences between the MLM and GM setups that may be contributing factors. First, the GM method adds up the surprisal scores for the whole sequence, while MLM only considers the softmax distribution at one token. Second, the input sequence for MLM always contains a {\tt [MASK]} token, whereas GM takes the original unmasked sequences as input, so the representations are never identical between the two setups.

MLM generally outperforms GM, but it does not solve every task: all three LMs fail to perform above chance on the data from \citet{warren}. This set of stimuli was designed so that both the control and impossible completions are not very likely or expected, which may have caused the difficulty for the LMs. We excluded the task of \citet{chow-role88} for MLM because the control and anomalous sentences differed by more than one token\footnote{Sentence pairs with multiple differing tokens are inconvenient for MLM to handle, but this is not a fundamental limitation. For example, \citet{salazar-mlm-scoring} proposed a modification to MLM to handle such cases: they compute a {\em pseudo-log-likelihood} score for a sequence by replacing one token at a time with a {\tt [MASK]} token, applying MLM to each masked sequence, and summing up the log likelihood scores.}.

\subsection{Differences between LMs}

RoBERTa is the best-performing of the three LMs in both the GM and MLM settings: this is expected since it is trained with the most data and performs well on many natural language benchmarks. Surprisingly, XLNet is ill-suited for this task and performs worse than BERT, despite having a similar model capacity and training data.

The surprisal gap plots for BERT (Figure \ref{fig:bert-manual}) and XLNet (Figure \ref{fig:xlnet-manual}) show some differences from RoBERTa: only morphosyntactic tasks produce out-of-domain embeddings in these two models, and not semantic or commonsense tasks. Evidently, how LMs behave when presented with anomalous inputs is dependent on model architecture and training data size; we leave exploration of this phenomenon to future work.

\section{Conclusion}

We used Gaussian models to characterize out-of-domain embeddings at intermediate layers of Transformer language models. The model requires a relatively small amount of in-domain data. Our experiments revealed that out-of-domain points in lower layers correspond to low-frequency tokens, while grammatically anomalous inputs are out-of-domain in higher layers. Furthermore, morphosyntactic anomalies are recognized as out-of-domain starting from lower layers compared to syntactic anomalies. Commonsense anomalies do not generate out-of-domain embeddings at any layer, even when the LM has a preference for the correct cloze completion. These results show that depending on the type of linguistic anomaly, LMs use different mechanisms to produce the output softmax distribution.

\newcommand{\cmark}{\ding{51}}
\newcommand{\xmark}{\ding{55}}

\chapter{Construction grammar in LMs}

{\em The contents of this chapter are based on my previous publication \citep{neural-argument-structure}.}

\section{Introduction}

This chapter continues our theme of probing language models using methods derived from linguistic theory. While the previous chapter focused on linguistic anomalies, here we shift our attention to examining argument structure in construction grammar theories and their representations in language models. Most probing work so far has investigated the linguistic knowledge of LMs on phenomena such as agreement, binding, licensing, and movement \citep{blimp, hu-syntax-assessment} with a particular focus on determining whether a sentence is linguistically acceptable \citep{schutze1996}. Relatively little work has attempted to determine whether the linguistic knowledge induced by LMs is more similar to a formal grammar of the sort postulated by mainstream generative linguistics \citep{chomsky1965, chomsky1981, chomsky1995}, or to a network of form-meaning pairs as advocated by construction grammar \citep{goldberg1995, goldberg2006}.

\begin{figure}
    \centering
    \includegraphics[width=0.5\linewidth]{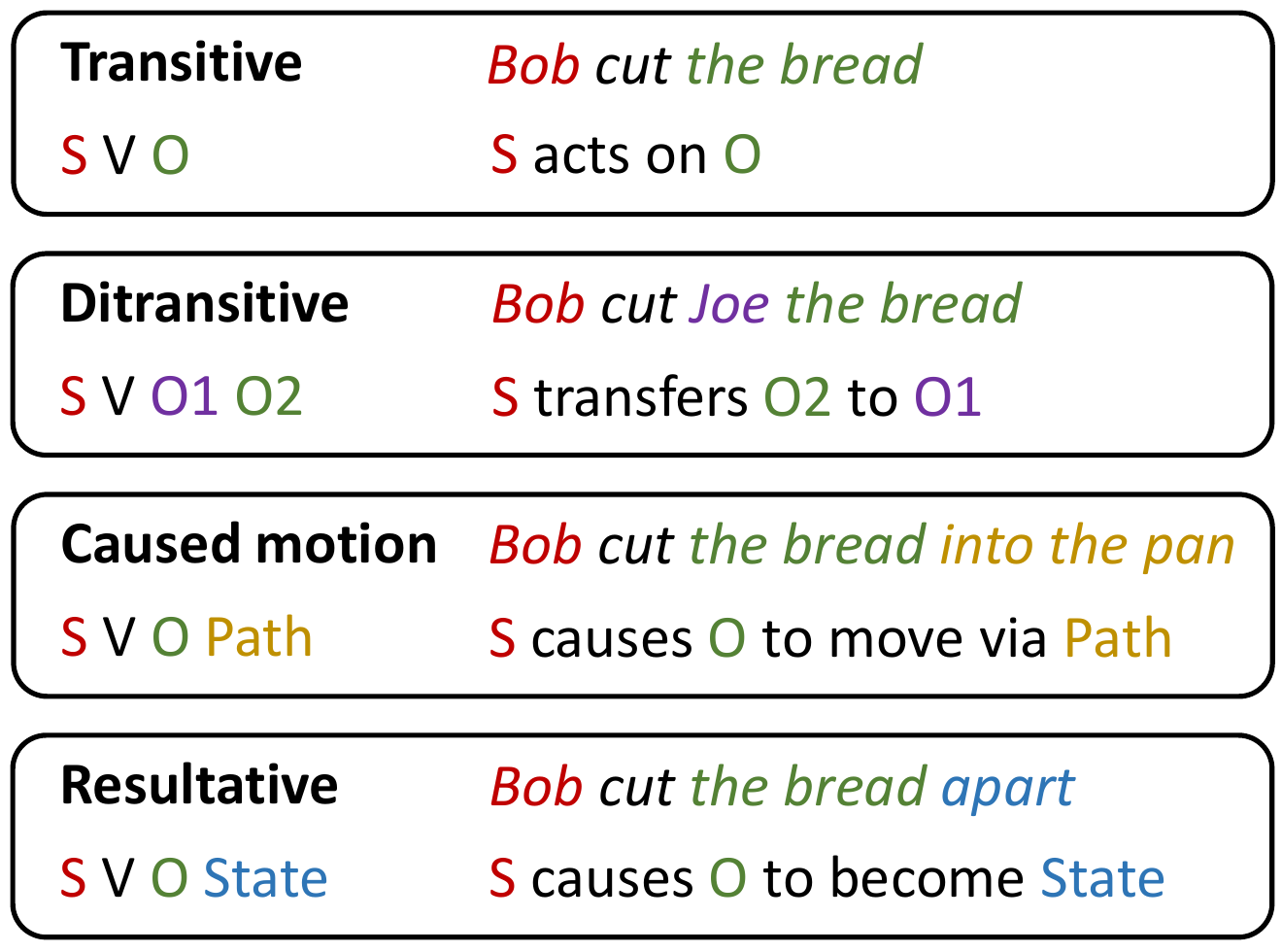}
    \caption{Four argument structure constructions (ASCs) used by \citet{bencini-goldberg}, with example sentences (top right). Constructions are mappings between form (bottom left) and meaning (bottom right).}
    \label{fig:cxn-examples}
\end{figure}

One area where construction grammar disagrees with many generative theories of language is in the analysis of the argument structure of verbs, that is, the specification of the number of arguments that a verb takes, their semantic relation to the verb, and their syntactic form \citep{levin-rappaport2005}. Lexicalist theories were long dominant in generative grammar \citep{chomsky1981, kaplan-bresnan1982, pollard-sag1987}. In lexicalist theories, argument structure is assumed to be encoded in the lexical entry of the verb: for example, the verb {\em visit} is lexically specified as being transitive and as requiring a noun phrase object \citep{chomsky1986}. In contrast, construction grammar suggests that argument structure is encoded in form-meaning pairs known as {\em argument structure constructions} (ASCs, Figure \ref{fig:cxn-examples}), which are distinct from verbs. The argument structure of a verb is determined by pairing it with an ASC \citep{goldberg1995}. To date, a substantial body of psycholinguistic work has provided evidence for the psychological reality of ASCs in sentence sorting \citep{bencini-goldberg, gries-wulff}, priming \citep{ziegler19}, and novel verb experiments \citep{kaschak-glenberg, johnson-goldberg}.

Here we connect basic research in ASCs with neural probing by adapting several psycholinguistic studies to Transformer-based LMs and show evidence for the neural reality of ASCs. Our first case study is based on sentence sorting \citep{bencini-goldberg}; we discover that in English, German, Italian, and Spanish, LMs consider sentences that share the same construction to be more semantically similar than sentences sharing the main verb. Furthermore, this preference for constructional meaning only manifests in larger LMs (trained with more data), whereas smaller LMs rely on the main verb, an easily accessible surface feature. Human experiments with non-native speakers found a similarly increased preference for constructional meaning in more proficient speakers \citep{liang2002, sorting-italian}, suggesting commonalities in language acquisition between LMs and humans.

Our second case study is based on nonsense {``Jabberwocky''} sentences that nevertheless convey meaning when they are arranged in constructional templates \citep{johnson-goldberg}. We adapt the original priming experiment to LMs and show that RoBERTa is able to derive meaning from ASCs, even without any lexical cues. This finding offers counter-evidence to earlier claims that LMs are relatively insensitive to word order when constructing sentence meaning \citep{yu-ettinger, unnatural-inference}. Our source code and data are available at: \url{https://github.com/SPOClab-ca/neural-reality-constructions}.

\section{Linguistic background}

\subsection{Construction grammar and ASCs}

Construction grammar is a family of linguistic theories proposing that all linguistic knowledge consists of {\em constructions}: pairings between form and meaning where some aspects of form or meaning are not predictable from their parts \citep{fillmore-et-all1988, kay-fillmore1999, goldberg1995, goldberg2006}. Common examples include idiomatic expressions such as {\em under the weather} (meaning { ``to feel unwell''}), but many linguistic patterns are constructions, including morphemes (e.g., {\em -ify}), words (e.g., {\em apple}), and abstract patterns like the ditransitive and passive. In contrast to lexicalist theories of argument structure, construction grammar rejects the dichotomy between syntax and lexicon. In contrast to transformational grammar, it rejects any distinction between surface and underlying structure.

We focus on a specific family of constructions for which there is an ample body of psycholinguistic evidence: argument structure constructions (ASCs). ASCs are constructions that specify the argument structure of a verb \citep{goldberg1995}. In the lexicalist, verb-centered view, argument structure is a lexical property of the verb, and the main verb of a sentence determines the form and meaning of the sentence \citep{chomsky1981, kaplan-bresnan1982, pollard-sag1987, levin-rappaport1995}. For example, {\em sneeze} is intransitive (allowing no direct object) and {\em hit} is transitive (requiring one direct object). However, lexicalist theories encounter difficulties with sentences like {\em ``he sneezed the napkin off the table''} since intransitive verbs are not permitted to have object arguments.

Rather than assuming multiple implausible senses for the verb {\em ``sneeze''} with different argument structures, \citet{goldberg1995} proposed that ASCs operate on an arbitrary verb, altering its argument structure while at the same time modifying its meaning. For example, the {\em caused-motion} ASC adds a direct object and a path argument to the verb {\em sneeze}, with the semantics of causing the object to move along the path. Other ASCs include the transitive, ditransitive, and resultative (Figure \ref{fig:cxn-examples}), which specify the argument structure of a verb and interact with its meaning in different ways.

\subsection{Psycholinguistic evidence for ASCs}

\begin{table*}[]
\small \centering
\begin{tabular}{l p{0.19\linewidth}p{0.19\linewidth}p{0.19\linewidth}p{0.19\linewidth}}
\hline
               & \textbf{Transitive}     & \textbf{Ditransitive}         & \textbf{Caused-motion}            & \textbf{Resultative}            \\ \hline
\textbf{Throw} & Anita threw the hammer. & Chris threw Linda the pencil. & Pat threw the keys onto the roof. & Lyn threw the box apart.        \\
\textbf{Get}   & Michelle got the book.  & Beth got Liz an invitation.   & Laura got the ball into the net.  & Dana got the mattress inflated. \\
\textbf{Slice} & Barbara sliced the bread. & Jennifer sliced Terry an apple. & Meg sliced the ham onto the plate. & Nancy sliced the tire open. \\
\textbf{Take}  & Audrey took the watch.  & Paula took Sue a message.     & Kim took the rose into the house. & Rachel took the wall down.      \\ \hline
\end{tabular}
\caption{Stimuli from \citet{bencini-goldberg}, consisting of a 4x4 design, with 4 different verbs and 4 different argument structure constructions.}
\label{table:bencini-goldberg-stimuli}
\end{table*}

{\bf Sentence sorting.} Several psycholinguistic studies have found evidence for argument structure constructions using experimental methods. Among these, \citet{bencini-goldberg} used a sentence sorting task to determine whether the verb or construction in a sentence was the main determinant of sentence meaning. 17 participants were given 16 index cards with sentences containing 4 verbs ({\em throw, get, slice}, and {\em take}) and 4 constructions ({\em transitive, ditransitive, caused-motion}, and {\em resultative}) and were instructed to sort them into 4 piles by overall sentence meaning (Table \ref{table:bencini-goldberg-stimuli}). The experimenters measured the deviation to a purely verb-based or construction-based sort, and found that on average, the piles were closer to a construction sort.

{\bf Non-native sentence sorting.} The same set of experimental stimuli was used with L2 (non-native) English speakers. \citet{gries-wulff} ran the experiment with 22 German native speakers, who preferred the construction-based sort over the verb-based sort, showing that constructional knowledge is not limited to native speakers. \citet{liang2002} ran the experiment on Chinese native speakers of 3 different English levels (46 beginner, 31 intermediate, and 33 advanced), and found that beginners preferred a verb-based sort, while advanced learners produced construction-based sorts similar to native speakers (Figure \ref{fig:human-lm-sorting-deviation}). Likewise, \citet{sorting-italian} found the same result in Italian native speakers with B1 and B2 English proficiency levels. Overall, these studies show evidence for ASCs in the mental representations of native and L2 English speakers alike, and furthermore, preference for constructional over verb sorting increases with increasing English proficiency.

{\bf Multilingual sentence sorting.} Similar sentence sorting experiments have been conducted in other languages, with varying results. \citet{sorting-german} ran a sentence sorting experiment in German with 40 participants and found that they mainly sorted by verb but rarely by construction. \citet{sorting-italian} ran an experiment with non-native learners of Italian (15 participants of B1 level and 10 participants of B2 level): both groups preferred the constructional sort, and similar to \citet{liang2002}, the B2 learners sorted more by construction than the B1 learners. \citet{sorting-spanish} ran an experiment in Spanish with 16 participants, and found approximately equal proportions of constructions and verb sort. In Italian and Spanish, some different constructions were substituted as not all of the English constructions had an equivalent in these languages.

{\bf Priming.} Another line of psycholinguistic evidence comes from priming studies. Priming refers to the condition where exposure to a (prior) stimulus influences the response to a later stimulus \citep{pickering-priming}. \citet{bock90} found that participants were more likely to produce sentences of a given syntactic structure when primed with a sentence of the same structure; \citet{ziegler19} argued that \citet{bock90} did not adequately control for lexical overlap, and instead, they showed that the construction must be shared for the priming effect to occur, not just shared abstract syntax.

{\bf Novel verbs.} Even with unfamiliar words, there is evidence that constructions are associated with meaning. \citet{kaschak-glenberg} constructed sentences with novel denominal verbs and found that participants were more likely to interpret a transfer event when the denominal verb was used in a ditransitive sentence ({\em Tom crutched Lyn an apple}) than a transitive one ({\em Tom crutched an apple}).

\citet{johnson-goldberg} used a ``Jabberwocky'' priming task to show that abstract constructional templates are associated with meaning. Participants were primed with a nonsense sentence of a given construction (e.g., {\em He daxed her the norp} for the ditransitive construction), followed by a lexical decision task of quickly deciding if a string of characters was a real English word or a non-word. The word in the decision task was semantically congruent with the construction ({\em gave}) or incongruent ({\em made}); furthermore, they experimented with target words that were high-frequency ({\em gave}), low-frequency ({\em handed}), or semantically related but not associated with the construction ({\em transferred}). They found priming effects (faster lexical decision times) in all three conditions, with the strongest effect for the high-frequency condition, followed by the low-frequency and the semantically nonassociate conditions.

\subsection{Related work in NLP}

Chapter 3 of this thesis surveyed recent work in language model probing based on linguistic theories, although relatively few of them have approached probing from a construction grammar perspective. \citet{cxgbert} probed for BERT's knowledge of constructions via a sentence pair classification task of predicting whether two sentences share the same construction. Their probe was based on data from \citet{dunn-data}, who used an unsupervised algorithm to extract plausible constructions from corpora based on association strength. However, the linguistic validity of these automatically induced constructions is uncertain, and there is currently no human-labelled wide-coverage construction grammar dataset in any language suitable for probing. Other computational work focused on a few specific constructions, such as identifying caused-motion constructions in corpora \citep{hwang-cm} and annotating constructions related to causal language \citep{dunietz-causal}. \citet{lebani-lenci} is the most similar to our work: they probed distributional vector space models for ASCs based on the Jabberwocky priming experiment by \citet{johnson-goldberg}.

In this work, we adapt several of the previously mentioned psycholinguistic studies to LMs: the sentence sorting experiments in Case study 1, and the Jabberwocky priming experiment in Case study 2. We choose these studies because their designs allow for thousands of stimuli sentences to be generated automatically using templates, avoiding issues caused by small sample sizes from manually constructed sentences.

\section{Case study 1: Sentence sorting}
\label{sec:sentence-sorting}

This section describes our adaptation of the sentence sorting experiments to Transformer LMs.

\subsection{Methodology}

{\bf Models.} To simulate varying non-native English proficiency levels, we use MiniBERTa models \citep{mini-bertas}, trained with 1M, 10M, 100M, and 1B tokens. We also use the base RoBERTa model \citep{roberta}, trained with 30B tokens. In other languages, there are no available pretrained checkpoints with varying amounts of pretraining data, so we use the mBERT model \citep{bert} and a monolingual Transformer LM in each language.\footnote{We use monolingual German and Italian models from \url{https://github.com/dbmdz/berts}, and the monolingual Spanish model from \citet{mono-spanish-lm}.} We obtain sentence embeddings for our models by taking the average of their contextual token embeddings at the second-to-last layer (i.e., layer 11 for base RoBERTa). We use the second-to-last because the last layer is more specialized for the LM pretraining objective and less suitable for sentence embeddings \citep{liu-probing}.

\begin{table*}
\small\centering
\begin{tabular}{l p{0.19\linewidth}p{0.19\linewidth}p{0.19\linewidth}p{0.19\linewidth}}
\hline
               & \textbf{Transitive}     & \textbf{Ditransitive}         & \textbf{Caused-motion}            & \textbf{Resultative}            \\ \hline
\textbf{Slice}  & Harry sliced the bread. & Henry sliced Eric the box. & Sam sliced the ball onto the bed. & John sliced the book apart. \\
\textbf{Kick} & Thomas kicked the box. & Mike kicked Frank the ball. & Michael kicked the wall into the house. & James kicked the door open. \\
\textbf{Cut} & George cut the ball. & Adam cut Paul the tree. & Bill cut the box into the water. & Bob cut the bread apart. \\
\textbf{Get}  & Tom got the book. & Andrew got Steve the door. & Jack got the fridge onto the elevator. & David got the ball stuck. \\ \hline
\end{tabular}
\caption{Example of our 4x4 sentence sorting stimuli, similar to those by \citet{bencini-goldberg} in Table \ref{table:bencini-goldberg-stimuli}, but generated automatically using templates.}
\label{table:templated-bg-stimuli}
\vspace{3em}

\begin{tabular}{l p{0.19\linewidth}p{0.19\linewidth}p{0.19\linewidth}p{0.19\linewidth}}
\hline
               & \textbf{Transitive}     & \textbf{Ditransitive}         & \textbf{Caused-motion}            & \textbf{Resultative}            \\ \hline
\textbf{Werfen}  & Anita warf den Hammer.     & Berta warf Linda den Bleistift.      & Erika warf den Schlüsselbund auf das Dach. & Laura warf die Kisten auseinander.  \\
\textbf{Bringen} & Michelle brachte das Buch. & Simone brachte Lydia eine Einladung. & Emma brachte den Ball ins Netz.            & Leonie brachte die Stühle zusammen. \\
\textbf{Schneiden} & Karolin schnitt das Brot. & Luisa schnitt Paula einen Apfel. & Jennifer schnitt die Wurst auf den Teller. & Doris schnitt den Reifen auf. \\
\textbf{Nehmen}  & Maria nahm die Uhr.        & Sophia nahm Jasmin das Geld.         & Helena nahm die Rosen in das Haus.         & Theresa nahm das Plakat herunter.   \\ \hline
\end{tabular}
\caption{German sentence sorting stimuli, obtained from \citet{sorting-german}.}
\label{table:german-sorting-stimuli}
\vspace{3em}

\begin{tabular}{l p{0.19\linewidth}p{0.19\linewidth}p{0.19\linewidth}p{0.19\linewidth}}
\hline
               & \textbf{Transitive}     & \textbf{Prepositional Dative}         & \textbf{Caused-motion}            & \textbf{Resultative}            \\ \hline
\textbf{Dare} & Lauda dà un esame.	& Carlo dà una mela a Maria. & Luca dà una spinta a Franco.	& Paolo dà una verniciata di verde alla porta. \\
\textbf{Fare} & Mario fa una torta.	& Luigi fa un piacere a Giovanna.	& Fabio fa entrare la macchina in garage.	& Stefano fa bruciare il sugo. \\
\textbf{Mettere} & Annalisa mette la giacca.	& Riccardo mette il cappello al bambino.	& Silvia mette la penna nel cassetto.	& Filippo mette la casa in ordine. \\
\textbf{Portare}  & Linda porta lo zaino.	& Laura porta la pizza a Francesco.	& Michele porta il libro in biblioteca.	& Irene porta l'esercizio a termine. \\ \hline
\end{tabular}
\caption{Italian sentence sorting stimuli, obtained from \citet{sorting-italian}.}
\label{table:italian-sorting-stimuli}
\vspace{3em}

\begin{tabular}{l p{0.19\linewidth}p{0.19\linewidth}p{0.19\linewidth}p{0.19\linewidth}}
\hline
               & \textbf{Transitive}     & \textbf{Ditransitive}         & \textbf{Unplanned Reflexive}            & \textbf{Middle}            \\ \hline
\textbf{Romper}  & Carlos rompió el cristal. &	Alfonso le rompió las gafas a Pepe.&	A Juan se le rompieron los pantalones.&	La porcelana se rompe con facilidad. \\
\textbf{Doblar} & Felipe dobló el periódico.&	Pablo le dobló el brazo a Lucas.&	A Pedro se le dobló el tobillo.&	El aluminio se dobla bien. \\
\textbf{Acabar} & Leonardo acabó su tesis.&	Tomás le acabó la pasta de dientes a Santi.&	A Luis se le acabaron los cigarrillos.&	Las carreras de 10 km se acaban sin problemas. \\
\textbf{Cortar}  & Isidro cortó el pan.&	Jorge le cortó el paso a Yago.&	A Ignacio se le cortó la conexión.&	Esta tela se corta muy bien.  \\ \hline
\end{tabular}
\caption{Spanish sentence sorting stimuli, obtained from \citet{sorting-spanish}.}
\label{table:spanish-sorting-stimuli}
\end{table*}

{\bf Template generation.} We use templates to generate stimuli similar to the 4x4 design in the \citet{bencini-goldberg} experiment. To ensure an adequate sample size, we run multiple empirical trials. In each trial, we sample 4 random distinct verbs from a pool of 10 verbs that are compatible with all 4 constructions ({\em cut, hit, get, kick, pull, punch, push, slice, tear, throw}). We then randomly fill in the slots for proper names, objects, and complements for each sentence according to its verb, such that the sentence is semantically coherent, and there is no lexical overlap among the sentences of any construction. Table \ref{table:templated-bg-stimuli} shows a set of template-generated sentences. In English, we generate 1000 sets of stimuli using this procedure.

For other languages, we use the original stimuli from their respective publications. We present the sentence sorting stimuli for German (Table \ref{table:german-sorting-stimuli}), Italian (Table \ref{table:italian-sorting-stimuli}), and Spanish (Table \ref{table:spanish-sorting-stimuli}). German uses the same four constructions as English. Italian does not have the ditransitive construction but instead uses the prepositional dative construction to express transfer semantics. Spanish has no equivalents for the caused-motion and resultative constructions, so the authors in that experiment instead used the unplanned reflexive (expressing accidental or unplanned events), and the middle construction (expressing states pertaining to the subject).

\begin{figure*}
    \centering
    \includegraphics[width=0.95\linewidth]{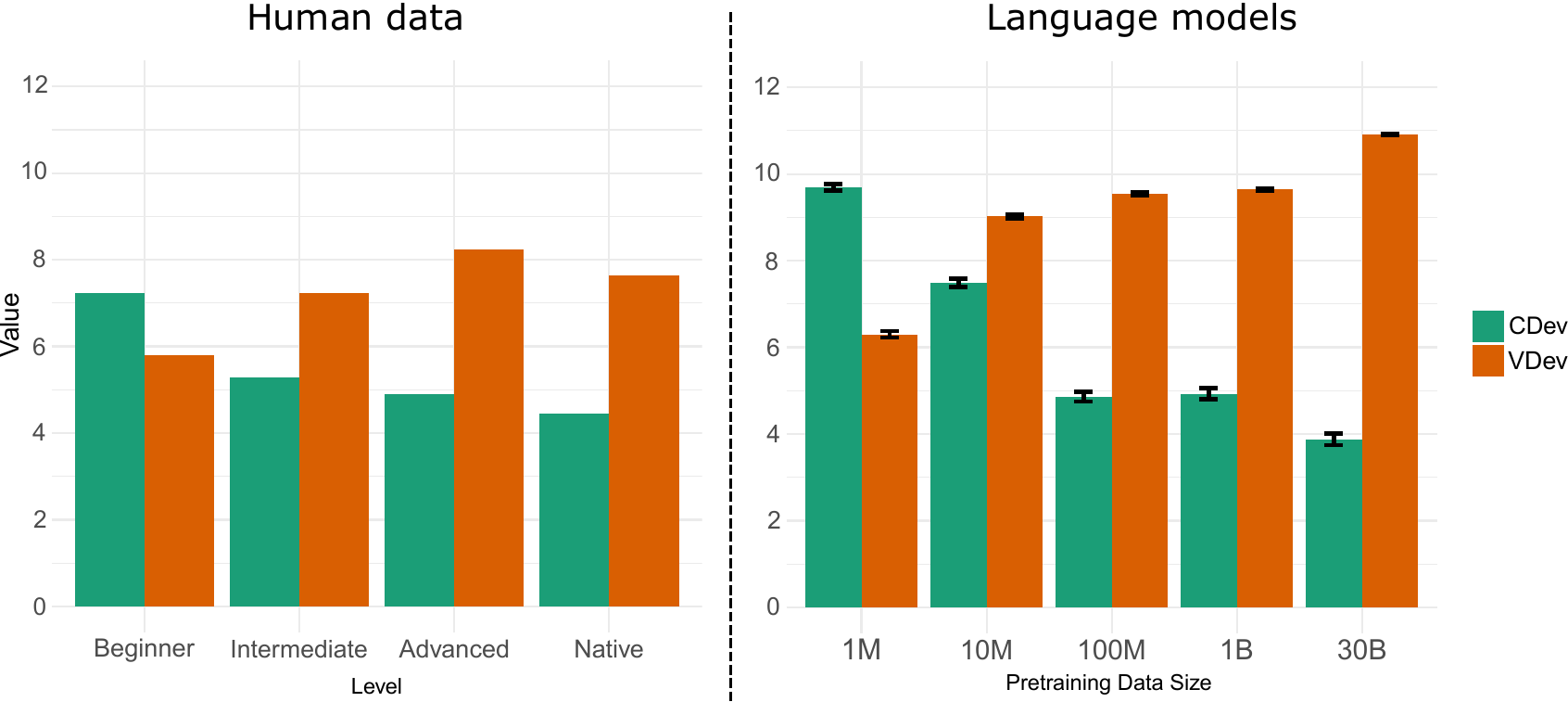}
    \caption{English sentence sorting results for humans and LMs, measured by deviation from pure construction and verb sort (CDev and VDev). Non-native human results are from \citet{liang2002}; native human results from \citet{bencini-goldberg}.\protect\footnotemark LM results are obtained using MiniBERTas \citep{mini-bertas} and RoBERTa \citep{roberta} on templated stimuli. The MiniBERTa models use between 1M to 1B tokens for pretraining, while RoBERTa uses 30B tokens. Error bars indicate 95\% confidence intervals.}
    \label{fig:human-lm-sorting-deviation}
\end{figure*}
\footnotetext{\citet{bencini-goldberg} ran the sentence sorting experiment twice, so we take the average of the two runs.}

\begin{figure}
    \centering
    \includegraphics[width=0.55\linewidth]{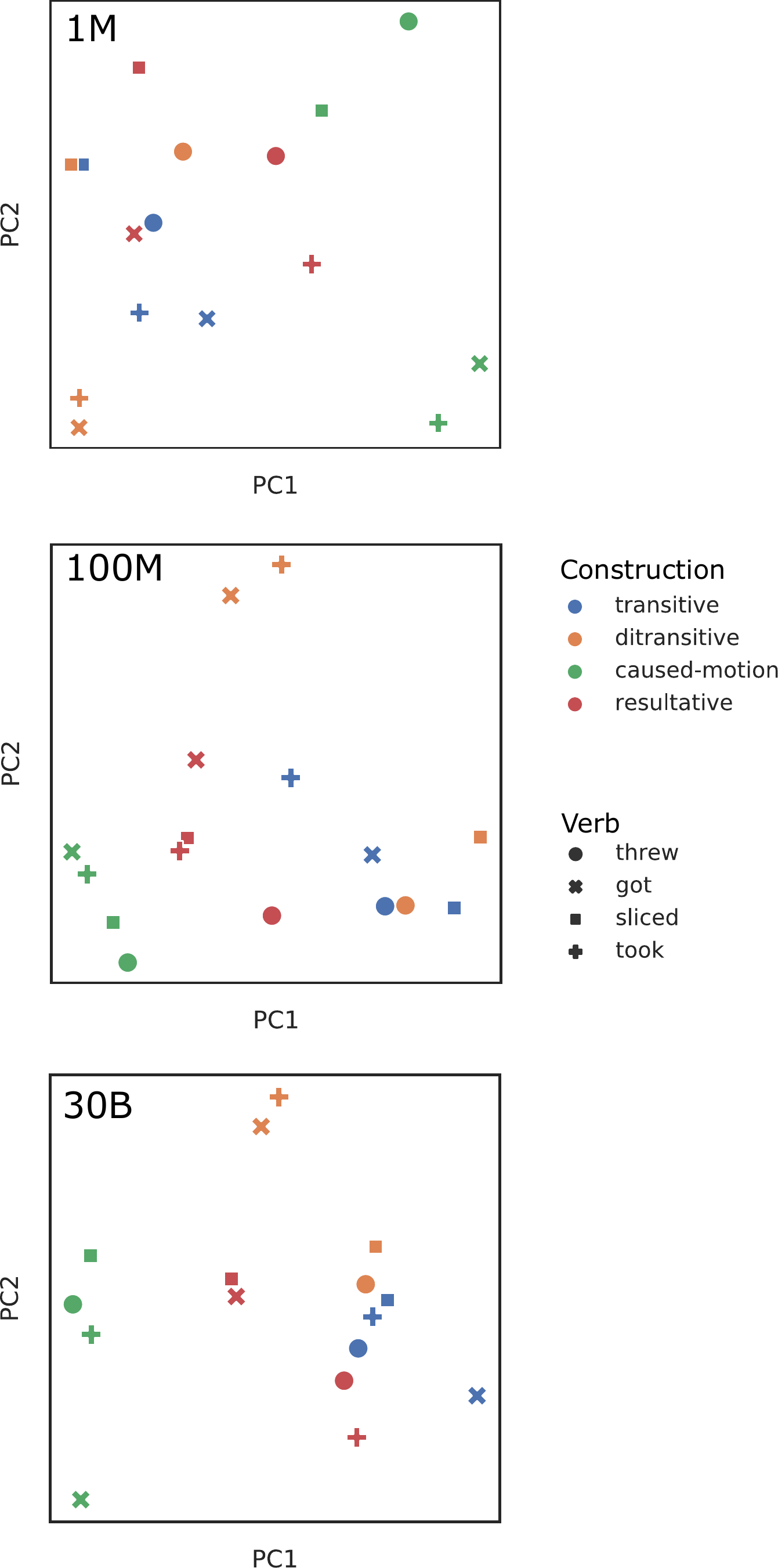}
    \caption{PCA plots of \citet{bencini-goldberg} sentence sorting using the 1M and 100M MiniBERTa models and RoBERTa-base (30B). Figure best viewed in color.}
    \label{fig:pca-plots}
\end{figure}

\begin{figure*}
    \centering
    \includegraphics[width=0.95\linewidth]{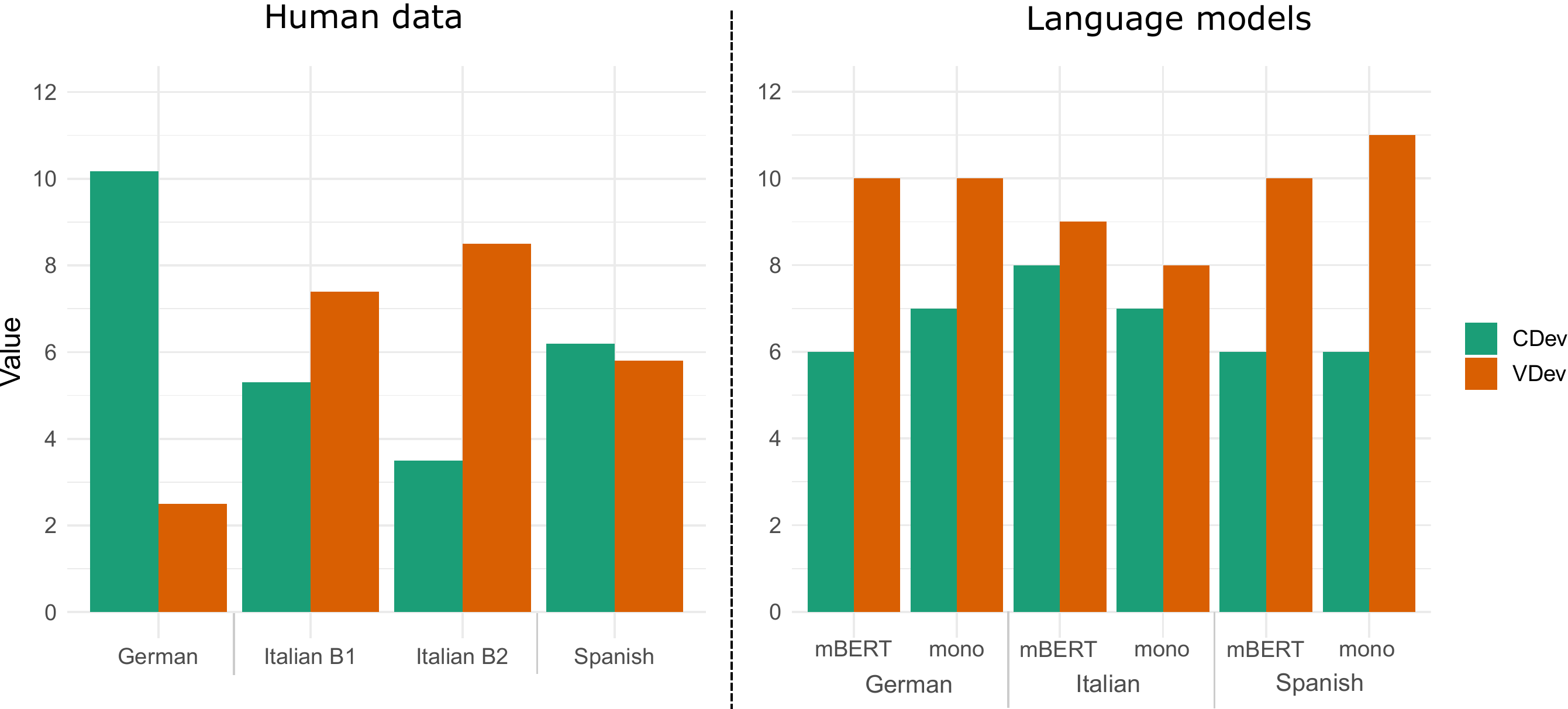}
    \caption{Multilingual sentence sorting results for German \citep{sorting-german}, Italian \citep{sorting-italian}, and Spanish \citep{sorting-spanish}. LM results are obtained using the same stimuli; we use both mBERT and a monolingual LM for each language.}
    \label{fig:multilang-sorting}
\end{figure*}

{\bf Evaluation.} Similar to the human experiments, we group the sentence embeddings into 4 clusters (not necessarily of the same size) using agglomerative clustering by Euclidean distance \citep{sklearn}. We then compute the deviation to a pure construction and pure verb sort using the Hungarian algorithm for optimal bipartite matching. This measures the minimal number of cluster assignment changes necessary to reach a pure construction or verb sort, ranging from 0 to 12. Thus, lower construction deviation indicates that constructional information is more salient in the LM's embeddings.

\subsection{Results and interpretation}
 
Figure \ref{fig:human-lm-sorting-deviation} shows the LM sentence sorting results for English. All differences are statistically significant ($p < .001$). The smallest 1M MiniBERTa model is the only LM to prefer verb over construction sorting, and as the amount of pretraining data grows, the LMs increasingly prefer sorting by construction instead of by verb. This closely mirrors the trend observed in the human experiments. To visualize this effect, we apply principal components analysis (PCA) on sentence embeddings for the 1M and 100M token MiniBERTa models and RoBERTa-base (Figure \ref{fig:pca-plots}). In RoBERTa, there is strong evidence of clustering based on constructions; the effect is unclear in the 100M model and nonexistent in the 1M model, visually confirming our quantitative evaluation based on the construction and verb deviation metrics.

The results for multilingual sorting are shown in Figure \ref{fig:multilang-sorting}. Both mBERT and the monolingual LMs consistently prefer constructional sorting over verb sorting in all three languages, whereas the results from the human experiments are less consistent.

Our results show that RoBERTa can generalize meaning from abstract constructions without lexical overlap. Only larger LMs and English speakers of more advanced proficiency are able to make this generalization, while smaller LMs and less proficient speakers derive meaning more from surface features like lexical content. This finding agrees with \citet{mini-bertas}, who found that larger LMs have an inductive bias towards linguistic generalizations, while smaller LMs have an inductive bias towards surface generalizations; this may explain the success of large LMs on downstream tasks. A small quantity of data (10M tokens) is sufficient for LMs to prefer the constructional sort, indicating that ASCs are relatively easy to learn: roughly on par with other types of linguistic knowledge, and requiring less data than commonsense knowledge \citep{zhang-billions, probing-across-time}.

We note some limitations in these results, and reasons to avoid drawing unreasonably strong conclusions from them. Human sentence sorting experiments can be influenced by minor differences in the experimental setup: \citet{bencini-goldberg} obtained significantly different results in two runs that only differed on the precise wording of instructions. In the German experiment \citep{sorting-german}, the author hypothesized that the participants were influenced by a different experiment that they had completed before the sentence sorting one. Given this experimental variation, we cannot attribute differences across languages to differences in their linguistic typology. Although LMs do not suffer from the same experimental variation, we cannot conclude statistical significance from the multilingual experiments, where only one set of stimuli is available in each language.

\section{Case study 2: Jabberwocky constructions}
\label{sec:jabberwocky-priming}

We next adapt the ``Jabberwocky'' priming experiment from \citet{johnson-goldberg} to LMs, and make several changes to the original setup to better assess the capabilities of LMs. Priming is a standard experimental paradigm in psycholinguistic research, but it is not directly applicable to LMs: existing methods simulate priming either by applying additional fine-tuning \citep{prasad19}, or by concatenating sentences that typically do not co-occur in natural text \citep{misra-priming}. Therefore, we instead propose a method to probe LMs for the same linguistic information using only distance measurements on their contextual embeddings.

\subsection{Methodology}

{\bf Template generation.} We generate sentences for the four constructions randomly using the templates in Table \ref{tab:jg-examples}. Instead of filling nonce words like {\em norp} into the templates as in the original study, we take an approach similar to \citet{gulordava-colorless} and generate 5000 sentences for each construction by randomly filling real words of the appropriate part-of-speech into construction templates (Table \ref{tab:jg-examples}). This gives nonsense sentences like {\em ``She traded her the epicenter''}; we refer to these random words as {\em Jabberwocky words}. By using real words, we avoid any potential instability from feeding tokens into the model that it has never seen during pretraining. We obtain a set of singular nouns, past tense verbs, and adjectives from the Penn Treebank \citep{ptb}, excluding words with fewer than 10 occurrences.

\begin{table}
\centering
\begin{tabular}{ll}
\hline
\textbf{Construction}          & \textbf{Template / Examples}             \\ \hline
\multirow{3}{*}{Ditransitive}  & S/he V-ed him/her the N.                  \\
                               & \textit{She traded her the epicenter.}    \\
                               & \textit{He flew her the donut.}           \\ \hline
\multirow{3}{*}{Resultative}   & S/he V-ed it Adj.                         \\
                               & \textit{He cut it seasonal.}              \\
                               & \textit{She surged it civil.}             \\ \hline
\multirow{3}{*}{Caused-motion} & S/he V-ed it on the N.                    \\
                               & \textit{He registered it on the diamond.} \\
                               & \textit{She awarded it on the corn.}    \\ \hline
\multirow{3}{*}{Removal}       & S/he V-ed it from him/her.                \\
                               & \textit{He declined it from her.}         \\
                               & \textit{She drove it from him.}           \\ \hline
\end{tabular}
\caption{Templates and example sentences for the Jabberwocky construction experiments. The templates are identical to the ones used in \citet{johnson-goldberg}, except that we use random real words instead of nonce words.}
\label{tab:jg-examples}
\end{table}

\begin{figure}
    \centering
    \includegraphics[width=0.35\linewidth]{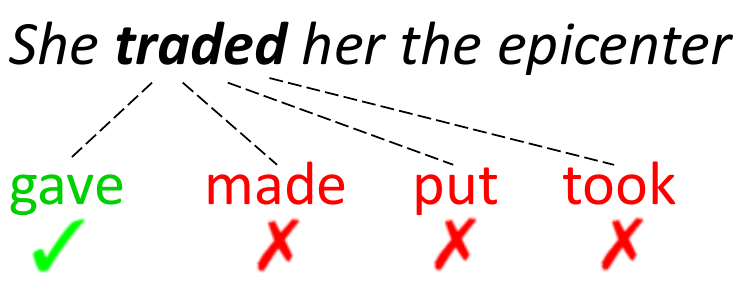}
    \caption{In our adapted Jabberwocky experiment, we measure the Euclidean distance from the Jabberwocky verb ({\em traded}) to the 4 prototype verbs, of which 1 is congruent ({\color{green}\cmark}) with the construction of the sentence, and 3 are incongruent ({\color{red}\xmark}).}
    \label{fig:jabberwocky-schema}
\end{figure}

{\bf Verb embeddings.} Our probing strategy is based on the assumption that the contextual embedding for a verb captures its meaning in context. Therefore, if LMs associate ASCs with meaning, we should expect the contextual embedding for the Jabberwocky verb to contain the meaning of the construction. Specifically, we measure the Euclidean distance to a {\em prototype} verb for each construction (Figure \ref{fig:jabberwocky-schema}). These are verbs that \citet{johnson-goldberg} selected whose meaning closely resembles the construction's meaning: {\em gave}, {\em made}, {\em put}, and {\em took} for the ditransitive, resultative, caused-motion, and removal constructions, respectively.\footnote{The reader may notice that the four constructions here are slightly different from \citet{bencini-goldberg}: the transitive construction is replaced with the removal construction in \citet{johnson-goldberg}.} We also run the same setup using lower frequency prototype verbs from the same study: {\em handed}, {\em turned}, {\em placed}, and {\em removed}.\footnote{\citet{johnson-goldberg} also included a third experimental condition using four verbs that are semantically related but not associated with the construction, but one of the verbs is very low-frequency ({\em ousted}), so we exclude this condition in our experiment.} As a control, we measure the Euclidean distance to the prototype verbs of the other three unrelated constructions.

The prototype verb embeddings are generated by taking the average across their contextual embeddings across a 4M-word subset of the British National Corpus (BNC; \citet{bnc}). We use the second-to-last layer of RoBERTa-base, and in cases where a verb is split into multiple subwords, we take the embedding of the first subword token as the verb embedding.

\subsection{Results and interpretation}

\begin{figure}
    \centering
    \includegraphics[width=0.6\linewidth]{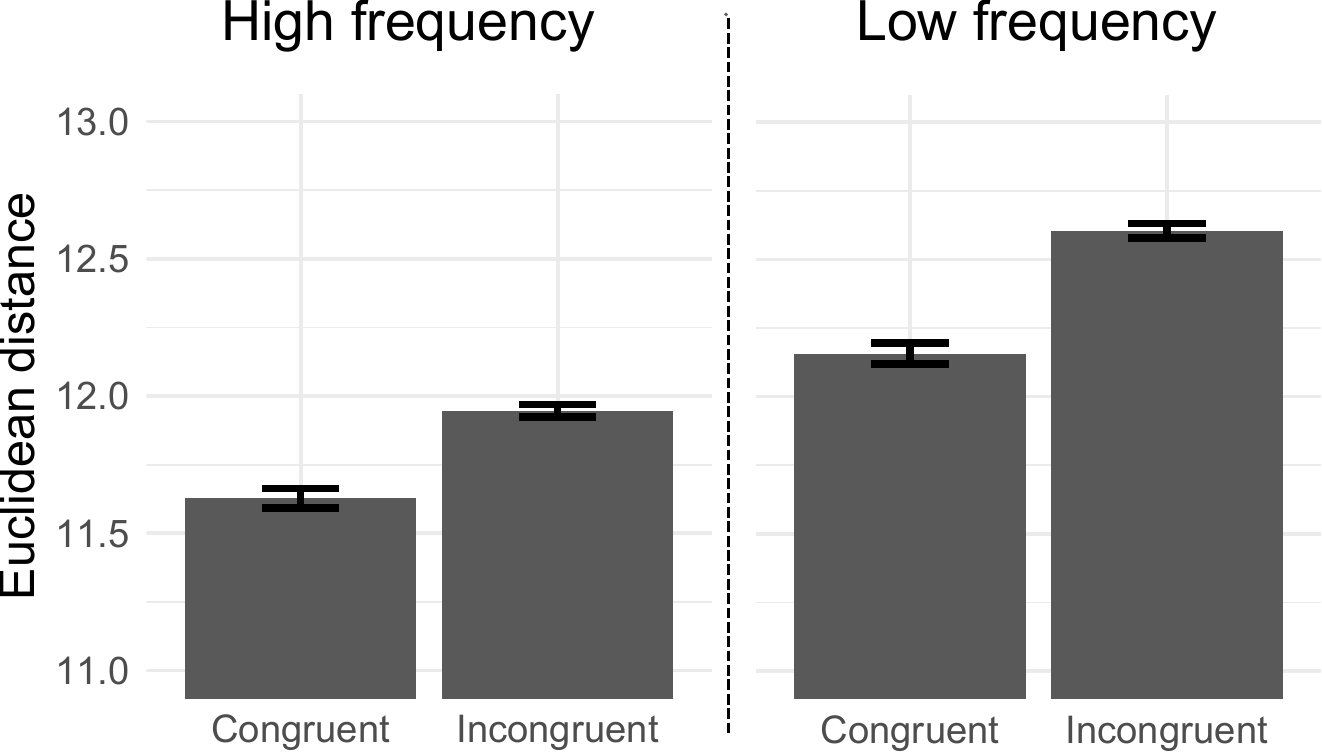}
    \caption{Euclidean distance between Jabberwocky and prototype verbs for congruent and incongruent conditions. Error bars indicate 95\% confidence intervals.}
    \label{fig:jg-barplot}
\end{figure}

\begin{figure*}
    \centering
    \includegraphics[width=1.0\linewidth]{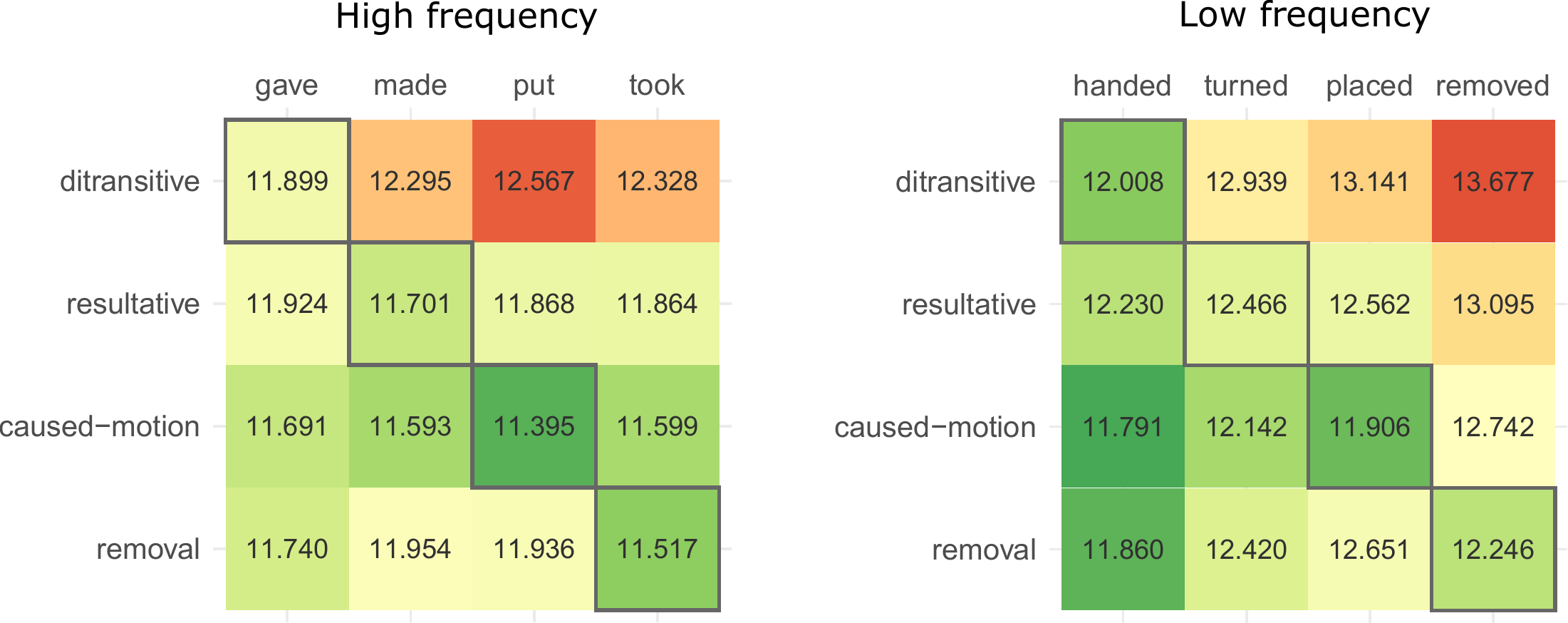}
    \caption{Mean Euclidean distance between Jabberwocky and prototype verbs in each verb-construction pair. Diagonal entries (gray border) are the congruent conditions; off-diagonal entries are incongruent.}
    \label{fig:jg-grid-4x4}
\end{figure*}

We find that the Euclidean distance between the prototype and Jabberwocky verb embeddings is significantly lower ($p < .001$) when the verb is congruent with the construction than when they are incongruent, and this is observed for both high and low-frequency prototype verbs (Figure \ref{fig:jg-barplot}). Examining the individual constructions and verbs (Figure \ref{fig:jg-grid-4x4}), we note that in the high-frequency scenario, the lowest distance prototype verb is always the congruent one, for all four constructions. In the low-frequency scenario, the result is less consistent: the congruent verb is not always the lowest distance one, although it is always still at most the second-lowest distance out of the four.

The main result holds for both high and low-frequency scenarios, but the correct prototype verb is associated more consistently in the high-frequency case. This agrees with \citet{wei-frequency}, who found that LMs have greater difficulty learning the linguistic properties of less frequent words. We also note that the Euclidean distances are higher overall in the low-frequency scenario, which is consistent with previous work that found lower frequency words to occupy a peripheral region of the embedding space \citep{layerwise-anomaly}.

\subsection{Potential confounds}

In any experiment, one must be careful to ensure that the observed patterns are due to the phenomenon under investigation rather than confounding factors. We discuss potential confounds arising from lexical overlap, anisotropy of contextual embeddings, and neighboring words.

{\bf Lexical overlap}. The randomized experiment design ensures that the Jabberwocky words cannot be lexically biased towards any construction, since each verb is equally likely to occur in every construction. Technically, the lexical content in the four constructions are not identical: i.e., words like {\em ``from''} (occurring only in the removal construction) or {\em ``on''} (in the caused-motion construction) may provide hints to the sentence meaning. However, the ditransitive and resultative constructions do not contain any such informative words, yet RoBERTa still associates the correct prototype verb for these constructions, so we consider it unlikely to be relying solely on lexical overlap. There is substantial evidence that RoBERTa is able to associate abstract constructional templates with their meaning without lexical cues. This result is perhaps surprising, given that previous work found that LMs are relatively insensitive to word order in compositional phrases \citep{yu-ettinger} and downstream inference tasks \citep{unnatural-inference, pham-out-of-order}, where their performance can be largely attributed to lexical overlap.

{\bf Anisotropy}. Recent probing work have found that contextual embeddings suffer from anisotropy, where embeddings lie in a narrow cone and have much higher cosine similarity than expected if they were directionally uniform \citep{how-contextual}. Furthermore, a small number of dimensions dominate geometric measures such as Euclidean and cosine distance, resulting in a degradation of representation quality \citep{kovaleva21, timkey21}. Since our experiments rely heavily on Euclidean distance, anisotropy is a significant concern. Following \citet{timkey21}, we perform standardization by subtracting the mean vector and dividing each dimension by its standard deviation, where the mean and standard deviation for each dimension is computed from a sample of the BNC. We observe little difference after standardization: in both the high and low frequency scenarios, the Euclidean distances are lower for the congruent than the incongruent conditions, by a similar margin compared to the original experiment without standardization. We also run standardization on the first case study, and find that the results remain essentially unchanged: smaller LMs still prefer verb sorting while larger LMs prefer construction sorting. Thus, neither of our experiments appear to be affected by anisotropy.

{\bf Neighboring words.} A final confounding factor is our assumption that RoBERTa's contextual embeddings represent word meaning, when in reality, they contain a mixture of syntactic and semantic information. Contextual embeddings are known to contain syntax trees \citep{hewitt-syntax} and linguistic information about neighboring words in a sentence \citep{klafka-ettinger}; although previous work did not consider ASCs, it is plausible that our verb embeddings leak information about the sentence's construction in a similar manner. If this were the case, the prototype verb embedding for {\em gave} would contain not only the semantics of transfer that we intended, but also information about its usual syntactic form\footnote{\citet{bresnan-gradience} estimated that 87\% of usages of the word {\em ``give''} occur in the ditransitive construction.} of {\em ``S gave NP1 NP2''}, and both would be captured by our Euclidean distance measurement. Controlling for this syntactic confound is difficult -- one could alternatively probe for transfer semantics without syntactic confounds using a natural language inference setup (e.g., whether the sentence entails the statement {\em ``NP1 received NP2''}), but we leave further exploration of this idea to future work.

\section{Conclusion}

We found evidence for argument structure constructions in Transformer language models from two separate angles: sentence sorting and Jabberwocky construction experiments. Our work extended the existing body of literature on LM probing by taking a constructionist instead of generative approach to linguistic probing. Our sentence sorting experiments identified a striking resemblance between humans' and LMs' internal language representations as LMs are exposed to increasing quantities of data, despite the differences between neural language models and the human brain. Our two studies suggest that LMs are able to derive meaning from abstract constructional templates with minimal lexical overlap. Both sets of experiments were inspired by psycholinguistic studies, which we adapted to fit the capabilities of LMs -- this illustrates the potential for future work on grounding LM probing methodologies in psycholinguistic research.
\chapter{Conclusion}

\section{Synopsis}

In this dissertation, I explored ways in which Transformer-based language models can provide evidence to support theories in linguistics, and how linguistic theory can provide probing frameworks for interpreting language models. My research has built connections between natural language processing and linguistics (drawing on research from both the theoretical and experimental psycholinguistic sides of linguistics). The two fields have much to contribute to each other, so it is worthwhile for researchers of both disciplines to be familiar with the tools and theories of the other, and look for opportunities to apply cross-disciplinary ideas in their own work.

Chapter 1 introduced the problem of interpreting neural language models and the motivations for probing over other evaluation methods. In Chapter 2, I surveyed the models to be probed, beginning with word vector models from the onset of the deep learning revolution, and culminating with the highly engineered Transformer-based models that rank at the top of leaderboards today. In Chapter 3, I reviewed recent linguistic probing research that tested the outputs of language models via behavioural probes and the internals of models via representational probes. Here, a wide range of linguistic phenomena combined with a diverse assortment of probing methods led to many novel results about what linguistic knowledge our models are capable of, and in which areas they remain deficient.

The next three chapters of my thesis contained my own contributions to the field. In Chapter 4, I tackled word class flexibility, a problem that is controversial in linguistic typology because linguists disagree about how it should be analyzed and how it should be compared across languages. My approach used contextual embeddings to argue that word class flexibility should be treated as a directional phenomenon, based on semantic evidence automatically computed from corpora across multiple languages. Chapter 5 explored how different types of linguistic anomalies are represented differently in language models. Inspired by human language processing studies of event-related potentials that trigger depending on the type of anomaly, I devised a method to probe for similar patterns in language models. The experiments revealed a notable difference in how various types of anomalies are represented. Chapter 6 draws on the psycholinguistic literature more directly, adapting several influential experiments to probe language models. The original studies presented evidence for the psychological reality of argument structure constructions in humans, while my results demonstrated their existence in language models, via a similar and parallel methodology.

Sadly, my thesis has come to an end, yet my discoveries leave many questions unanswered and opportunities for further exploration. I will next discuss some promising avenues for future work in the linguistic probing direction. I hope that my work will inspire future collaboration between natural language processing researchers and linguists.

\section{Future directions}

\subsection{Which models to probe?}

When engaging in probing research, a decision must be made at some point about which models to probe. BERT is the most popular choice, but many newer models have surpassed it in performance so that it is no longer state-of-the-art; it can be misleading to present its deficiencies as representative of language models in general, given that newer models may have improved in these aspects \citep{bowman-hype}. In my work, I used BERT, ELMo, RoBERTa, and XLNet in English experiments, and mBERT, XLM-R, and various monolingual models for non-English experiments; these are more or less the most popular models in the community at this time.

One may wonder how relevant this body of work will be in the future, when BERT and RoBERTa are surpassed by newer models. Indeed, many architectures such as ELECTRA \citep{electra} and DeBERTa \citep{deberta} claim improvements over BERT and RoBERTa, but these newer models are rarely the subject of probing research. When probing sentence representations, models dedicated to the task such as Sentence-BERT \citep{sentence-bert} are rarely used, despite their superior performance over average pooling over token vectors or taking the \texttt{[CLS]} vector, methods commonly used in probing setups.

Inevitably, newer models will exhibit similar linguistic patterns as current models in some cases, while differing in other cases. In my view, probing work will remain relevant despite newer models that behave differently, because the primary contributions are the novel methodologies to probe for various linguistic phenomena in continuous representations, and not the results of the probing experiments themselves. As long as newer models continue to use similar layers of continuous vectors, it is straightforward to adapt existing linguistic probing tests and obtain an assessment of the capabilities of the new model, using far less effort than inventing probing procedures from scratch.

The trouble is that when a probing procedure gives different results when applied to different models, it is often not possible to explain why, in a satisfactory manner. This limitation applies to Chapter 4 of this dissertation (where XLM-R performed worse than mBERT on judging similarity between noun-verb pairs), as well as Chapter 5 (where XLNet did not exhibit the same difference between anomaly types as RoBERTa). Explaining these differences is problematic because architectural differences are generally far removed from concepts in linguistic theory. For example, Sentence-BERT uses a siamese architecture with triplet loss to learn sentence embeddings; ELECTRA uses a pretraining task of predicting corrupted tokens instead of masked language modelling. Any attempts to find connections to linguistic theory would likely only be speculative.

As language model pretraining becomes more accessible, exploring these differences in a systematic manner will become more feasible. Some recent work investigated the effects of structural versus sequential model architecture \citep{hu-syntax-assessment}, and genre of training data \citep{babyberta} on probing performance. These experiments require training many variants of models to isolate the effects of each architectural parameter, and should become easier to perform in future work as language model tools and frameworks continue to improve.

\subsection{Evidence from learnability}

A common criticism of neural network probing research is its lack of relevance to linguistic theory \citep{baroni-proper-role}. Even as we analyze the linguistic abilities of BERT and other models in increasing detail, this type of work does not lead to an improved understanding of human language processing, so its impact outside of natural language processing will likely be limited. One promising direction is using language models to study learnability (e.g., \citet{wilcox-island}), an approach that is currently underexplored. This idea is that when neural networks are able to learn some linguistic feature from corpora alone, that constitutes evidence that no other mechanisms (such as innate grammar or interactions grounded in the real world) are necessary to learn the feature.

Learnability has been featured in studies of argument structure constructions as well. \citet{goldberg-casenhiser} proposed that learning of ASCs is facilitated when the distribution of verbs in a construction is skewed towards a frequent prototypical verb (for example, ``give'' for the ditransitive construction), compared to a balanced distribution of several verbs. Their evidence came from studies of a child language corpus and an artificial language learning experiment in which subjects learned novel verbs and constructions. We are limited to indirect studies because it is impractical to manipulate the language input to children over their lifetime for an experiment. Unlike humans, language models can be trained from scratch on artificial data to serve as tools to test learnability hypotheses. In this example, we may train one model using a balanced distribution of verbs and train another model using a skewed distribution of verbs, and compare which one is more successful at learning ASCs by probing them. If such an experiment finds that skewed verb distributions are helpful for language models learning ASCs, the theory of construction learning in humans would be strengthened.

\subsection{Psycholinguistic-based probing}

Language model probing has a lot in common with psycholinguistics: the goal of both fields is to probe the internals of a language processing entity through indirect experimental methods. Psycholinguistics benefits from being a more mature field and closer alignment with linguistic theory, since many psycholinguistic studies are designed to support or refute theories of how language is processed cognitively.

Currently, only a tiny fraction of the numerous psycholinguistic publications in the last few decades have been considered for adaptation to neural network probing. Given this choice, one can either select studies that examine a specific linguistic phenomenon (using different methodologies), or select studies that employ the same methodology (studying different phenomena). In this thesis, I have mostly used the former strategy: Chapter 5 took data from multiple sources that contain linguistic anomalies, and Chapter 6 adapted studies on argument structure constructions. Other authors aggregated psycholinguistic studies using the same methodology, such as \citet{michaelov-n400}, who focused on the N400 effect, and \citet{prasad19}, who examined syntactic priming. In either case, adapting psycholinguistic work to neural network probing is an effective way of bridging the gap between theoretical linguistics and natural language processing, thereby improving our understanding of language models through the lens of linguistic theory.

\addcontentsline{toc}{chapter}{Bibliography}
\bibliographystyle{acl_natbib}
\bibliography{thesis}


\end{document}